\documentclass[10pt,letterpaper]{article}

\usepackage{graphicx}
\usepackage{subcaption}
\usepackage{url}
\usepackage{hyperref}  
\usepackage{comment}
\usepackage{caption}
\usepackage{float}
\usepackage{adjustbox}
\usepackage{array, caption, tabularx, ragged2e, booktabs}
\usepackage{listings}
\usepackage{longtable}
\newcolumntype{P}[1]{>{\RaggedRight\arraybackslash}p{#1}}
\usepackage{pdflscape}
\usepackage{xcolor}
\usepackage{amsmath, amsfonts}
\usepackage{algorithmic}
\usepackage{algorithm}
\usepackage{textcomp}
\usepackage{stfloats}
\usepackage{multirow}
\usepackage{verbatim}

\usepackage{cleveref}
\usepackage{soul}
\usepackage[top=0.85in,left=2.75in,footskip=0.75in]{geometry}

\usepackage{amsmath,amssymb}

\usepackage{changepage}

\usepackage{textcomp,marvosym}

\usepackage{cite}

\usepackage{nameref,hyperref}

\usepackage[right]{lineno}

\usepackage[nopatch=eqnum]{microtype}
\DisableLigatures[f]{encoding = *, family = * }

\usepackage[table]{xcolor}

\usepackage{array}

\newcolumntype{+}{!{\vrule width 2pt}}

\newlength\savedwidth

\raggedright
\usepackage[aboveskip=1pt,labelfont=bf,labelsep=period,justification=raggedright,singlelinecheck=off]{caption}

\makeatletter
\renewcommand{\@biblabel}[1]{\quad#1.}
\makeatother

\usepackage{lastpage,fancyhdr,graphicx}
\usepackage{epstopdf}
\fancyheadoffset[L]{2.25in}
\fancyfootoffset[L]{2.25in}
\begin{document}
\vspace*{0.2in}

\begin{flushleft}
{\Large
\textbf\newline{Reassessing feature-based Android malware detection in a contemporary context} 
}
\newline
\\
Ali Muzaffar*\textsuperscript{1},
Hani Ragab Hassen\textsuperscript{1},
Hind Zantout\textsuperscript{1},
Michael A Lones\textsuperscript{2},

\bigskip
\textbf{1} Heriot-Watt University, Dubai, UAE
\\
\textbf{2} Heriot-Watt University, Edinburgh EH14 4AS, United Kingdom
\\

\bigskip

%
%





* ali.muzaffar@hw.ac.uk

\end{flushleft}
\section{Abstract}
\textcolor{black}{We report the findings of a reimplementation of 18 foundational studies in feature-based machine learning for Android malware detection, published during the period 2013--2023. These studies are reevaluated on a level playing field using a contemporary Android environment and a balanced dataset of 124,000 applications. Our findings show that feature-based approaches can still achieve detection accuracies beyond 98\%, despite a considerable increase in the size of the underlying Android feature sets. We observe that features derived through dynamic analysis yield only a small benefit over those derived from static analysis, and that simpler models often out-perform more complex models.  
We also find that API calls and opcodes are the most productive static features within our evaluation context, network traffic is the most predictive dynamic feature, and that ensemble models provide an efficient means of combining models trained on static and dynamic features. Together, these findings suggest that simple, fast machine learning approaches can still be an effective basis for malware detection, despite the increasing focus on slower, more expensive machine learning models in the literature.}



\section{Introduction}

Smartphone usage has increased exponentially over recent years. \textcolor{black}{According to Ericsson's mobility report \cite{ericsson2023}, there was a year-on-year increase of 6 per cent in the number of smartphones in the first quarter of 2024, with 71.67\% of them using an Android smartphone \cite{androidData}.} It is common practice to augment smartphones with applications, which in Android can be downloaded from the official Google Play Store \cite{playstore} or third-party application stores. The availability and easy access to applications through third party stores, in particular, provides attackers with a means of distributing malware. According to the latest report published by the computer security company G Data, a new piece of Android malware appears on the Internet every 12 seconds \cite{malwareStat}.

Various methods have been proposed to secure mobile operating systems (OS), including the application sandbox approach used by Android \cite{Elenkov2015}. In order to grant access to device services outside of the sandbox, Android uses a permissions system; however, this has its own shortcomings \cite{Almomani2020}. A number of security solutions, including malware detectors, vulnerability detection, user and developer reviews have been proposed \cite{Sufatrio2015}. Of these solutions, Android malware detection, or anti-malware, is one of the most widely used. Anti-malware can initially be used to stop an application from being released into application stores or later at the user level, preventing the user from installing the application. Traditional approaches to malware detection use signatures to detect malicious files. However, any slight variation to the file might cause this approach to fail. In particular, this makes it difficult to detect zero-day attacks or existing malware variations. With the rapid growth of Android malware, there may be occasions when signature-based anti-malware will not be able to detect thousands, and potentially more, existing malware. \textcolor{black}{Machine learning (ML) based approaches, on the other hand, attempt to get around these problems by learning models of malware that generalise beyond single applications. These models can then be used to detect malware samples that were not seen during training.}

\textcolor{black}{ML is a fast-changing field. One recent change is the move from feature-based approaches to end-to-end learning. In the context of Android malware detection, feature-based approaches involve analysing applications in order to extract features that can then be used to train models.} This analysis might be static or dynamic. Static analysis uses domain knowledge to extract relevant features from the source code of the application, whereas dynamic analysis extracts features observed whilst monitoring the behaviour of the application in a running state \cite{Sikorski2012}. \textcolor{black}{End-to-end approaches, by comparison, do not delineate feature extraction from modelling. Instead, features are implicitly learnt as part of the model training process. In theory, this can lead to more sensitive modelling of patterns in data, but it comes with greater computational expense, lower interpretability, and an increased likelihood of overfitting.}

\textcolor{black}{In 2022, we surveyed approaches that use ML to detect Android malware \cite{Muzaffar2022}, and observed that most feature-based approaches report detection accuracies well above 95\%, in some cases approaching 100\%. Since then, the focus of new studies has moved on to more complex end-to-end modelling approaches such as transformers. However, given that published accuracies were already very high, this begs the question of whether there is a need for more complex, more expensive, approaches.}

\textcolor{black}{As discussed by us in \cite{Muzaffar2022}, and more recently by Guerra-Manzanares \cite{guerra2023android}, published accuracy figures can be unreliable. There are several reasons for this. One is the historical context in which they were evaluated; Android malware continues to evolve, and approaches that worked in the past may not work today. Further to this are weaknesses in the original evaluation methodologies; many were evaluated with obsolete versions of Android running historic, and often quite small datasets. Third, ML as a modelling approach has considerable problems in terms of replicability \cite{lones2024avoiding,kapoor2024reforms}. It is common for the results of reimplemented studies to be significantly different to the originally published results \cite{KAPOOR2023100804, gibney2022ai}. This is often due to errors made by practitioners, and it has been reported that such errors are commonplace in the security domain \cite{arp2022and}.}

\textcolor{black}{Within this context, the aim of this study is to determine whether feature-based approaches to Android malware detection are still sufficient to detect malware within a contemporary Android environment.} We primarily do this by reimplementing foundational works from the period 2013-2023 and reevaluating them using a large dataset that comprises recent Android applications collected during the period 2019--2021.
It is not feasible, nor perhaps desirable, to reimplement all the previous work in this area. Instead we focus on a group of studies which are representative of the diversity of modelling approaches used in this field, favouring those which have reported high levels of malware discrimination, guided by \cite{Muzaffar2022}. \textcolor{black}{Our emphasis is on the comparative analysis of feature representations, model selection and feature selection strategies rather than on the chronological progression of research.} We also conduct a number of new experiments to fill in knowledge gaps in the existing literature, and consider the benefits of using ensemble models that combine existing approaches.


These are the main contributions of this work:
    \begin{itemize}
        \item \textcolor{black}{We reimplement and reevaluate 18 foundational studies in feature-based Android anti-malware, and find that the majority still achieve accuracies beyond 95\%, with the best models reaching above 98\%. This questions the increasing focus on more complex and more expensive models in malware detection.}
        \item \textcolor{black}{We assess these approaches on a level playing field using a balanced and up-to-date dataset of Android malware and benign applications.}
        We share both this dataset and the tools used to create it to make it easier for the community to develop and evaluate future anti-malware.
        \item
        \textcolor{black}{We find that ML models with static features often perform well against those with more expensive dynamic and hybrid features.
        We also find that more traditional ML models often out-perform deep learning models, which again questions the increasing focus on more complex approaches.}
        \item We present an ensemble model that leverages the best performing static and dynamic models, achieving the highest accuracy and true positive rate on our contemporary dataset. Notably this does not require the use of brittle network features, which are used by the best performing individual models.
        \item We study the role of feature selection, and show that significantly better models can be produced through aggressive feature selection. This also provides a route to handle challenges caused by the rapidly increasing complexity of Android, which has considerably boosted feature counts.
    \end{itemize}

The paper is organized as follows. Section \ref{sec:methods} describes our methodology, including dataset collection, feature extraction, and our evaluation framework. This methodology is then applied to static analysis, dynamic analysis, and hybrid analysis --- the three main branches of feature analysis and model building in malware detection --- in Sections \ref{sec:static}, \ref{sec:dynamic} and \ref{sec:hybrid}, respectively. \textcolor{black}{Section \ref{sec:bestmodels} then presents the best individual models from the previous sections and uses ensemble methods to combine these.} Section \ref{sec:discussion} discusses the main findings of the study, \textcolor{black}{Section \ref{sec:limitations} discusses limitations and future work,} and Section \ref{sec:conclusions} concludes.

\section{\textcolor{black}{Background}}

\textcolor{black}{There is an extensive literature on the use of ML in Android malware detection, and many of these past studies have reported a strong ability for ML models to discriminate malware from benign applications. In general, ML models are not trained directly on raw application files. Rather, static analysis and dynamic analysis are used to extract features that are then used to train and evaluate models. To place our study in context, this section provides a brief review of past works on Android malware detection based on ML, focusing on the broad directions of travel. For a more in depth review of the literature in this area, we point readers to \cite{Muzaffar2022}.} 

\subsection{\textcolor{black}{Static approaches}}

\textcolor{black}{Android applications are distributed in APK (Android Package Kit) format, which can be unpacked into metadata and source code files. Static analysis is the process of extracting relevant information from these files. It does not involve any information about the application's runtime behaviour.}

\textcolor{black}{Permissions are a core part of the Android security model and have shown strong performance when used as the primary feature in ML models. Permissions represent a relatively concise feature set, with up to 166 permissions depending on the Android OS version. Peiravian and Zhu \cite{Peiravian2013} were one of the earliest to use them to train ML models, and reported promising results. Using various feature selection approaches, later studies then narrowed the range of permissions used. These reported accuracies of up to 99.6\% \cite{Wang2014,Rathore2021,Sahin2023,Ojo2023,mahindru2024permdroid}, though it is surprising that such high accuracies can be achieved using features which do not directly relate to an application's source code.} 

\textcolor{black}{Another important source of features in previous studies has been the use of API calls. Unlike the use of permissions, this directly captures properties of an application's source code. The simplest approach captures whether each API call is used, or how often it is used. More complex approaches involve analysing the sequence of API calls or the structure of the API call graph. ML models trained on API call features seem to perform as well as, or in some cases better than, those that use permissions \cite{Ma2019,muzaffar2023android,wang2023android}, especially when feature selection is used \cite{Jung2018,Muzaffar2021}. However, API call information can be challenging to work with in practice because of the large number of different API calls that may occur in source code. For instance, there are currently over 100,000 distinct calls available in the Android API.}


\textcolor{black}{Another reportedly productive approach has been the use of features that represent Android opcodes, the low-level virtual machine instructions into which source code is compiled. Past works have represented opcodes as n-grams \cite{Kang2016}, or used natural language processing representations such as word2vec \cite{yeboah2022}, reporting F1 scores of up to 98\%. Opcode sequences have also been translated into psuedoimages for use with convolutional neural networks (CNNs) \cite{Xiao2019a,daoudi2021dexray}.}

\textcolor{black}{Another influential feature set is the one first used within Arp et al's highly-cited 2014 tool Drebin \cite{Arp2014}, where accuracies of up to 94\% were reported. This comprised a more diverse range of static features, including certain permissions and API calls in addition to other application components. Despite its age, more recent studies have continued to use either the Drebin feature set or its accompanying application data set \cite{Muzaffar2022}.}

\textcolor{black}{The last few years have seen the use of increasingly more complex deep learning models to identify malware, reflecting broader trends towards increasing complexity within ML \cite{zhao2025android,sun2025malicious}. However, it is notable that the performance metrics reported in these studies have not increased significantly above those reported in earlier studies, and this places some questions on the need for more complex approaches.}

\subsection{\textcolor{black}{Dynamic and hybrid approaches}}

\textcolor{black}{Dynamic analysis involves the extraction of information at runtime, i.e.\ whilst an Android application is executing. This involves running the application on a virtual or real device and monitoring its behaviour. Dynamic analysis has the potential to provide more relevant features for malware detection. For example, API call features can also be extracted during dynamic analysis, and when done at this stage, give an indication of actual call usage rather than potential API calls \cite{Afonso2015}. However, this comes at the expense of a more complex and resource-intensive extraction process.}

\textcolor{black}{The added complexity of dynamic analysis means that there are fewer studies that involve dynamic features. However, one feature which has been widely used in these studies is system calls, which are low-level requests made to the operating system kernel. System calls, like API calls, can be represented as sequences \cite{Malik2019} or using numeric features that capture frequency or usage \cite{Ananya2020}. Accuracies of up to 99.4\% have been reported for ML models that use system call features.}

\textcolor{black}{Network traffic generated while an application is running is another dynamic feature which appears to be effective for identifying malware, with accuracies of over 98\% consistently reported in past works \cite{Zulkifli2018,Wang2016}. However, in practice, there are challenges around collecting reliable network traces, especially when dynamic analysis is done in a virtual environment.} 

\textcolor{black}{Some past works have used a combination of static and dynamic features to build ML models. Kandukuru and Sharma \cite{Kandukuru2017} and Shyong et al.\ \cite{Shyong2020} both combined permissions with network traffic features, again reporting high accuracies. Other combinations, including permissions and system calls \cite{Kapratwar2017}, have also shown promising results.} 

\subsection{\textcolor{black}{End-to-end approaches}}

\textcolor{black}{End-to-end approaches learn the entire process without intervention, from input data to classifying malware or benign samples. End-to-end approaches typically use transformers, process low-level data (e.g.~sequences of API calls or raw code), and do not explicitly carry out feature extraction or feature selection. The benefit of this is the need for less manual feature engineering and more streamlined pipelines that can lead to easier deployments. However, the downside is that the models may learn meaningless features, and they are typically expensive and slow to use in comparison to feature-based ML.}

\textcolor{black}{The published results of studies suggest that end-to-end approaches do not currently outperform feature-based ML-based approaches. Chimezie Obidiagha et al. \cite{Obidiagha2024} use payloads directly as input and feed them into transformers, achieving accuracy rates of up to 96\%. Souani et al.\cite{Souani2022} utilised a BERT model on Android permissions and achieved 97\% accuracy in the binary classification of apps. Bourebaa and Benmohammed \cite{Bourebaa2025} utilised permissions and API calls with a range of more efficient BERT transformer models, reporting accuracies of up to 91.6\%. Alshomrani et al.~\cite{alshomrani2024survey} survey other recent approaches.}

\textcolor{black}{These results suggest that, despite the ease and appeal of end-to-end learning, such approaches may not offer significant advantages in terms of accuracy over traditional ML methods, making them potentially less beneficial for practical deployments. However, it should be borne in mind that the performance metrics published in feature-based studies may not accurately reflect their contemporary utility, and this further justifies our aim of reimplementing these studies, in order to give a more realistic baseline against which to compare more recent approaches.}

\subsection{\textcolor{black}{Challenges}}

\textcolor{black}{The performance metrics reported in the literature might suggest that Android malware detection is largely a solved problem, yet there are a number of reasons to doubt whether these metrics are reliable indicators of how well these approaches would work in practice \cite{Muzaffar2022, guerra2024machine}.}

\textcolor{black}{Most notably, many of the studies cited above used small, dated or inbalanced collections of Android malware and benign applications \cite{Muzaffar2022} This includes the 2014 Drebin dataset \cite{Arp2014} and the 2017 AMD dataset \cite{AMD}. In the years (and decades) since these datasets were released, malware has changed significantly, so the conclusions drawn by these studies may be unreliable.}

\textcolor{black}{In addition, most studies used historic versions of the Android operating system; often these were considered historic at the time of the study, with a lack of up-to-date feature extraction software preventing the use of a more contemporary environment \cite{Muzaffar2022}. This means that the feature sets used are not representative of current Android versions; for example, the number of API calls has grown from a little over 1000 to well over 100,000 since the earliest studies cited above were published.}

\textcolor{black}{There are also issues with reporting standards within the Android malware detection literature. Imbalanced datasets are commonly used, yet some studies only report accuracy values, which can be very misleading when derived from imbalanced data \cite{lones2024avoiding}. Other studies report a limited range of metrics which do not enable a full understanding of their ML model's behaviour. Despite the widespread use of stochastic models, many studies only report the performance of a single model, or provide averages without measures of spread. Furthermore, statistical tests are rarely used to compare models, which further limits the reliability of findings.}

\textcolor{black}{These issues, combined with the general high error rates in ML-based studies \cite{KAPOOR2023100804}, leaves the community in a position where it is challenging to draw robust conclusions from the published literature. For instance, many studies have found that relatively simple features in combination with relatively simple ML models can be used to consistently identify malware. It is unclear whether this remains the case today and it is important to establish the current state of play given the recent proliferation of AI genrated malware and to understand whether there is any need for more complex approaches.}

\textcolor{black}{The primary purpose of this study is to cast new light on the existing literature, by reavaluating existing approaches within a contemporary Android environment, using a large balanced up-to-date dataset, and following rigourous reporting standards. We also extend the existing literature in a number of ways by filling in notable gaps and by applying ensemble methods that combine existing modelling approaches in novel ways.}

\section{Methodology} \label{sec:methods}

\begin{figure}
    \centering
    \includegraphics[width=0.95\columnwidth]{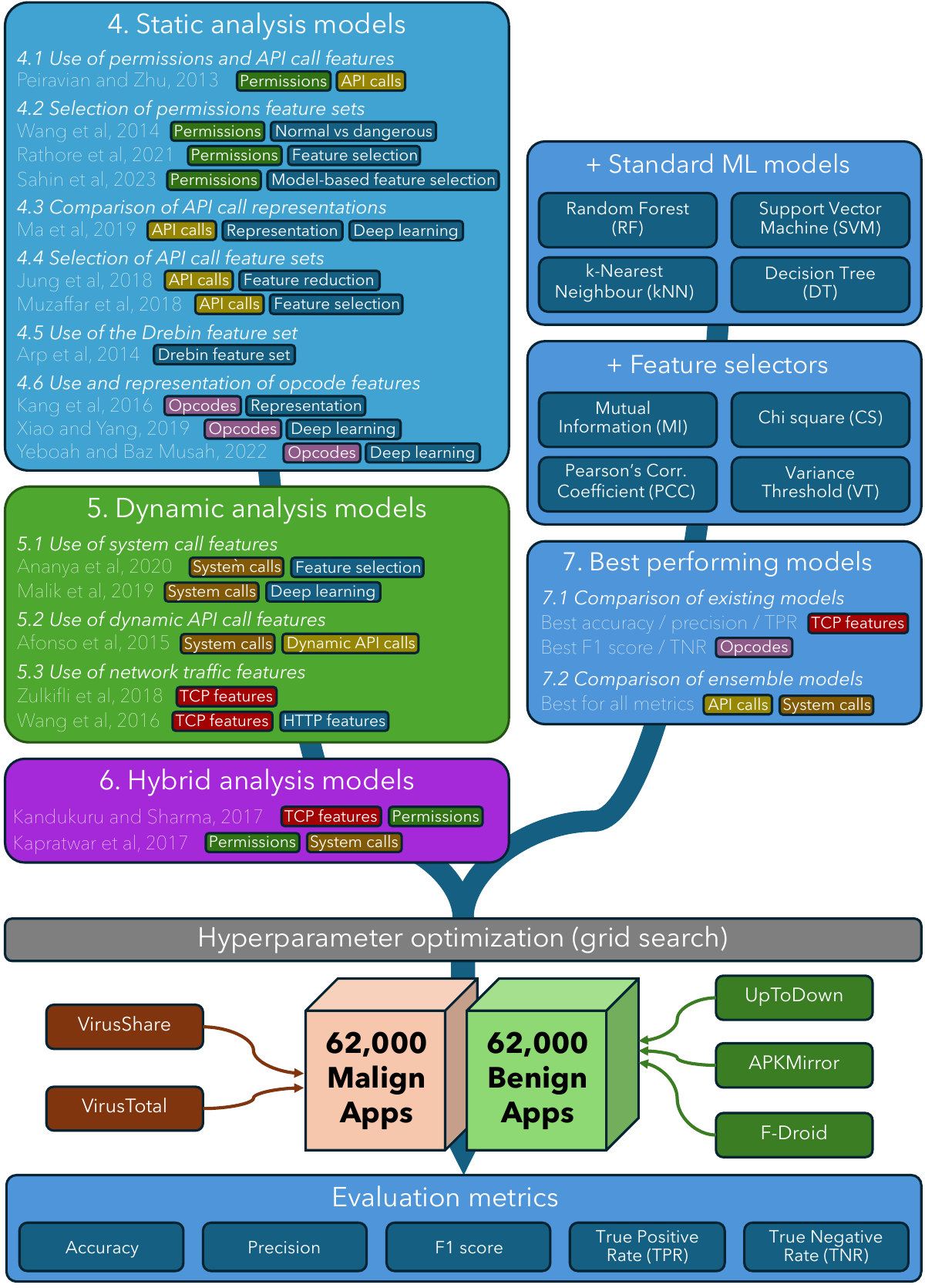}
    \caption{\textcolor{black}{Overview of the methodology, showing section numbers containing results}}
    \label{fig:summary}
\end{figure}

In this section, we present our experimental framework for assessing Android anti-malware approaches. \textcolor{black}{Section \ref{sec:methodology:studies} discusses the basis for selecting studies to include. Section \ref{sec:methodology:core} presents the core set of ML models and feature selection methods that we use to extend and standardise the original studies. Section \ref{sec:methodology:dataset} describes the evaluation dataset, and how it was collected. Section \ref{sec:methodology:droiddissector} summarises how we extract features from Android applications through static and dynamic analysis. Section \ref{sec:methodology:metrics} outlines the metrics and procedures we use to assess ML models. Fig. \ref{fig:summary} provides an overview.}
 
 

\subsection{\textcolor{black}{Choice of studies to reimplement}}\label{sec:methodology:studies}

\textcolor{black}{There have been a large number of studies which have applied feature-based ML approaches to Android malware detection.}
It is not feasible, or desirable, to reimplement all of these.
Instead, guided by existing reviews of the literature \cite{Naway2018,Alqahtani2019a,Bayazit2020,Wang2020, Muzaffar2022}, we have chosen to reimplement (a) seminal studies that introduced specific approaches, and (b) studies which built on these approaches in notable ways. The latter includes those which further refined the approach, considered alternative representations, or applied more complex models. These are categorised and listed in Fig.~\ref{fig:summary}, in each case highlighting the features which were used to build ML models, alongside other aspects such as the use of feature selection methods and the use of deep learning models. A primary guiding principle was to select studies which explored a diversity of underlying features, in order to objectively compare their utility. A secondary principle was to include studies which used different ML models, in order to objectively gain insight into the influence of model choice and complexity.

We focus on the binary classification task of discriminating malware from benign applications. However, some studies (e.g. \cite{son2021risk,son2022risk}) consider multi-class problems, typically with the aim of discriminating between different malware families. We chose to exclude these for several reasons. First, a key aim of this study is to understand the role of different models and features in the malware detection process, and binary classification provides the most generalisable insights into this. Second, there is no universally-accepted method for labeling malware families, with the ever-changing and diverse nature of malware leading to varying taxonomies and classifications across the cybersecurity community. Again, this poses difficulties when trying to reach generalisable insights. Third, there is a significant imbalance within the natural distribution of malware types. For instance, in the dataset collected for this study, adware comprises around 70\% of the samples. This introduces a challenge of handling imbalanced data, but also the challenge of separating the effects of mechanisms used to address imbalance from the effects of models and features.

Apart from correcting any clear errors of practice, during the process of reimplementation we have preserved the modelling processes as described in the original studies. All the studies are then compared within the same experimental scenario, so that we have a level playing field from which we are able to make robust insights. This includes standardising the data used to train and evaluate models (see Section \ref{sec:methodology:dataset}), carrying out consistent hyperparameter tuning to avoid unfairness (Section \ref{sec:methodology:core}), and (where appropriate) including relevant modelling combinations that were not considered in the original study in order to provide a more complete picture of its potential.

\subsection{\textcolor{black}{Inclusion of} core ML models and feature selection algorithms} \label{sec:methodology:core}

Previous studies have used a diverse range of ML models. In general, we reimplement the model(s) used in the original study wherever possible. However, to provide better consistency and comparability across studies, we also augment the original studies with the following core set of ML models, if they are not already included: support vector machine (SVM), random forest (RF), decision tree (DT) and $k$-nearest neighbour (kNN) \cite{hastie2009elements}. All of these are widely used both within the Android malware detection literature (as surveyed in \cite{Muzaffar2022}) and the broader ML literature. We use grid search to tune the hyperparameters of all the models implemented. This is to ensure fairness when making comparisons between models \textcolor{black}{\cite{Lones2021}}.

Many studies also use feature selection algorithms to reduce feature counts to a manageable level. Again, whenever possible, we reimplement the approach(es) used in the original study. However, we also augment these with the following core set of feature selection algorithms: mutual information (MI), chi-square, Pearson's correlation coefficient (PCC), and variance threshold (VT). All of these are widely used in the literature.

\subsection{Dataset collection} \label{sec:methodology:dataset}

The quality of the data used is a key factor in evaluating the quality of any ML model. In the context of assessing Android anti-malware approaches, we require a dataset that is both up-to-date and representative of the Android software ecosystem. We focus on the binary classification task of discriminating malware from benign software, and hence the dataset needs to contain samples of both malware and benign software. Ideally these two classes should be equal in number, since this mitigates against dealing with unbalanced data during the course of building and evaluating ML models. 

We collected\footnote{\textcolor{black}{Dataset available at}  \url{https://researchportal.hw.ac.uk/en/datasets/android-dataset-for-malware-detection}} Android applications released from 2019-2021 from the stores and repositories which are typically used by end users. The lack of availability of scripts to build datasets prompted us to build crawlers to download these applications. Through periodic use, these platform-independent Python scripts can be used to maintain an up-to-date dataset. The stores we targeted were UpToDown \cite{uptodown}, APKMirror \cite{apkmirror} and F-Droid \cite{fdroidsite}. The scripts crawl the websites for all applications irrespective of the categories the applications are filed under. We downloaded a total of 62,000 benign applications. To the best of our knowledge, this is the most realistic and up to date benign dataset available right now.   
 
To collect malware, we used the most recent Android malware dataset from VirusShare \cite{virusshare} for malware released in 2020. We also identified malware while downloading applications from the application stores. We used VirusTotal \cite{virustotal} reports to label all the applications. To prevent false positives in the benign dataset, we only included applications with zero positive tags from anti-malware in VirusTotal reports. The final malware dataset consisted of 62,000 applications, matching the size of the benign dataset.

\subsection{\textcolor{black}{Feature extraction process}}\label{sec:methodology:droiddissector}

\begin{figure}
    \centering
    \includegraphics[width=\columnwidth]{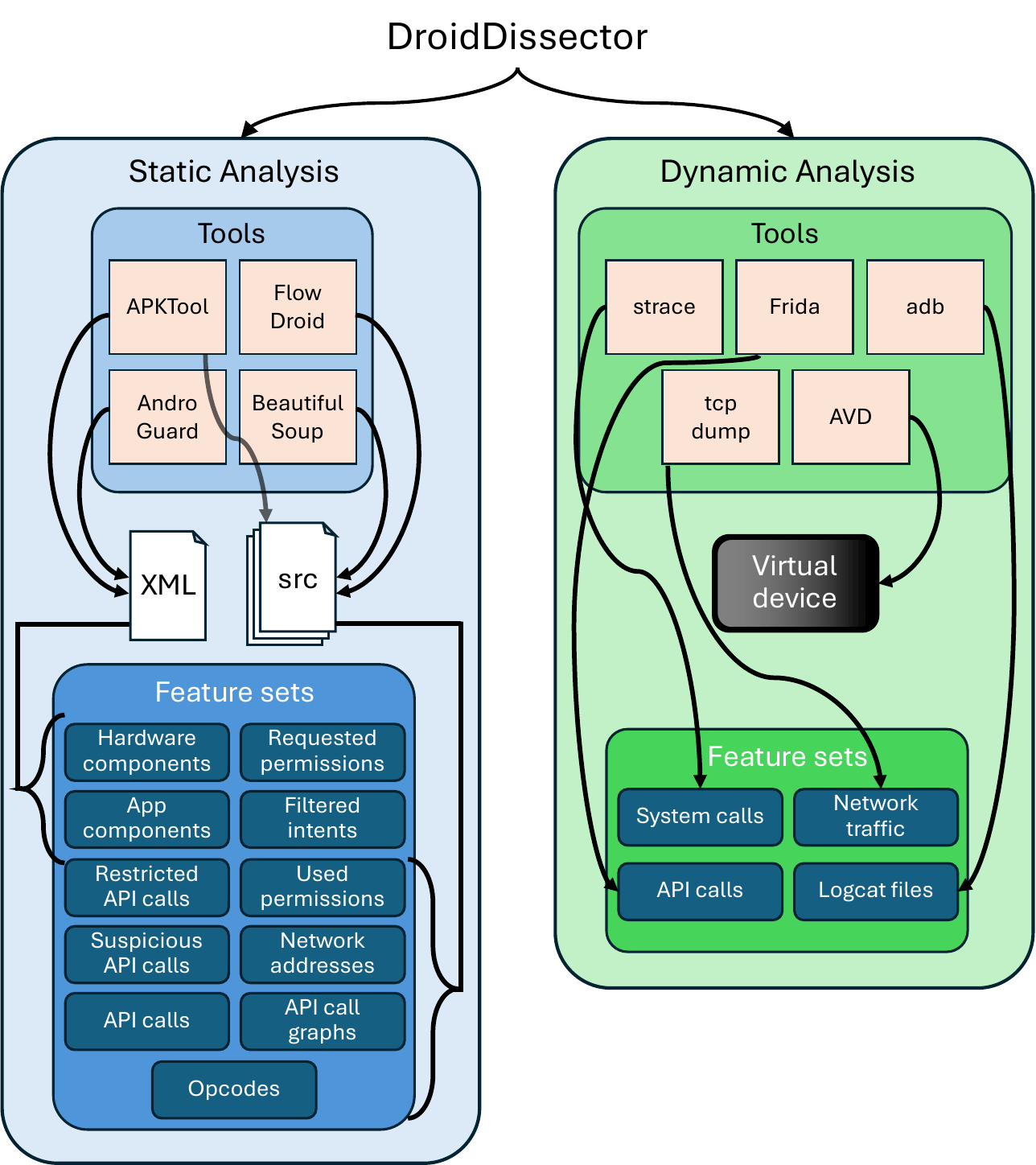}
    \caption{Overview of the DroidDissector feature extraction tool used in this study}
    \label{fig:toolarchitecture}
\end{figure}

We developed an automated analysis tool, DroidDissector\footnote{\textcolor{black}{Code and scripts available at} \url{https://github.com/alisolehria/DroidDissector-A-Static-and-Dynamic-Analysis-Tool-for-Android-Malware-Detection}}~\cite{muzaffar2023droiddissector}, to extract the static and dynamic features needed for reimplementing studies. The tool's architecture is depicted in Figure \ref{fig:toolarchitecture}. It comprises two sub-systems: one for extracting static features, and another for extracting dynamic features. \textcolor{black}{Full details of how these are implemented together with the features they extract are provided in \ref{appendix:droiddissector}. For static analysis, the tool extracts features that capture permissions, API calls, opcodes, hardware and app components, intents and network addresses. It also extracts the API call graph, and provides features summarising use of restricted and suspicious API calls, as used in the Drebin feature set. For dynamic analysis, the tool extracts features that capture system and API call usage, alongside network traffic and system logs. To provide scalability of large datasets, dynamic analysis is done whilst the application is running in an emulator. Unlike other commonly-used tools for dynamic analysis, DroidDissector runs on current versions of Android.}


\subsection{\textcolor{black}{Model evaluation procedure}} \label{sec:methodology:metrics}
The following metrics are used to present results for the ML models trained in this study:
 
\begin{description}
    \item [Confusion matrix] \textcolor{black}{The predictions of a malware detection model can be divided into four classes:}
    \begin{enumerate}
    \item True Positive (TP): The application is correctly predicted as malicious.
    \item False Positive (FP): The application is falsely predicted as malicious.
    \item True Negative (TN): The application is correctly predicted as benign.
    \item False Negative (FN): The application is falsely predicted as benign.
\end{enumerate}

    \textcolor{black}{These four numeric values can be presented as a confusion matrix:}

\begin{table}[h]
\centering
\label{confmatrix}
\begin{tabular}{lll}
\toprule
\textbf{Class} & \textbf{Positive} & \textbf{Negative} \\ 
\midrule
Malware        & TP                & FN                \\ 
Benign         & FP                & \textcolor{black}{TN} \\ 
\bottomrule
\end{tabular}
\end{table}

\item [Accuracy] is the percentage of correct predictions. 
\begin{equation}
Accuracy = \frac{TP+TN}{TP+TN+FP+FN}
\end{equation}

\item [Precision] is the percentage of correctly classified malware from all predictions of malware. 
\begin{equation}
Precision = \frac{TP}{TP+FP}
\end{equation}

\item [True Positive Rate (TPR)] is the proportion of actual malware that are correctly identified. Also called recall. 
\begin{equation}
TPR = \frac{TP}{TP+FN}
\end{equation}

\item [F1-Score] is the harmonic mean of precision and recall. 
\begin{equation}
    F_1 =  \frac{2*Precision*Recall}{Precision+Recall} = \frac{2*TP}{2*TP+FP+FN}
\end{equation}

\item [True Negative Rate (TNR)] is the proportion of actual benign applications that are correctly identified. 
\begin{equation}
TNR = \frac{TN}{FP+TN}
\end{equation}

\end{description}

In general, it is necessary to use a range of metrics in order to give a full picture of the performance of different ML models \cite{Lones2021}. With this in mind, we report the accuracy, precision, F1-score, TPR and TNR for each model. TPR and TNR, in particular, help in understanding a model's error rates. In the case of malware detection, it is particularly important to consider TPR, since a benign application labelled as malware is less likely to cause as many problems as malware labelled as benign.
 
The most common approach to comparing ML models in the Android malware literature is to report the average of metrics after using a resampling method like $k$-fold cross validation. Following this approach, we use standard 10-fold cross-validation (CV), and report the averages for all metrics. \textcolor{black}{The choice of 10 folds gives a good balance between bias and variance. It is computationally efficient, especially when dealing with complex feature sets, and is widely adopted in similar malware detection studies, ensuring reliable and reproducible results.}

However, the average of a distribution can be misleading, particularly when there is a high variance between folds. To account for this, we report standard deviations and also carry out statistical tests when comparing averages of different approaches for malware detection. In a statistical test, a null hypothesis is set up \textbf{(H\textsubscript{0})}: the average is equal for all sample population groups. A significant difference between the samples is found if the results from the test were to reject the null hypothesis and accept the alternative hypothesis \textbf{(H\textsubscript{a})}: the average is \textbf{not} equal for all sample population groups.

In this work, we use the Kruskal–Wallis \cite{Kruskal1952} omnibus test followed by pairwise Dunn’s \cite{Dunn1964} tests. Kruskal–Wallis is used to determine if there are significant differences between the medians of two or more distributions. In this case, the distributions are the set of $k$ metric scores produced by each ML model across $k$ folds. \textcolor{black}{Kruskal–Wallis is non-parametric, so does not assume that the data fits a normal distribution. This is important because the distribution of accuracies from k-fold CV may not be normally distributed, and there is no reliable means of measuring normality in small samples.} If the test statistic p-value is less than a threshold level, the null hypothesis is rejected, indicating that there is a significant difference between the groups. Kruskal-Wallis is an omnibus test and so does not indicate which sample group is statistically different from another. If Kruskal–Wallis finds a significant difference \textcolor{black}{at the group level}, Dunn’s test is then used to identify the samples that are statistically different from the others. \textcolor{black}{Dunn's is a pairwise test, i.e.~for the CV distributions of two models, it determines whether there is a significant difference in their medians. If there are more than two models in a group, a pairwise tournament between all ML models is used to those which exhibit significant statistical differences from the others.} Dunn's test automatically adjusts for multiple comparisons when used with Kruskal-Wallis. We use a threshold of p=0.05 for all statistical tests, i.e. a confidence of 95\%. \textcolor{black}{The previous studies we review did not, in general, carry out statistical tests, but these are important in order to reach reliable conclusions.}

\section{\textcolor{black}{Reimplementation and extension of static modelling approaches}} \label{sec:static}
 
In this section, we focus on the use of static features in building Android malware detection models. Previous work in this area has considered various types of static features, including permissions, API calls, API call graphs and opcodes. We reimplement \textcolor{black}{eleven} of these previous studies, focusing on those that were notable either in terms of the methodologies they used or the performance of their models on the original datasets. We also report the results of new experiments that are intended to fill in some of the knowledge gaps found in previous studies.
 
\subsection{\textcolor{black}{Use of permissions and API call features}}
 
To begin with, we revisit one of the earliest studies that looked at permissions and API calls, carried out by \textbf{Peiravian and Zhu \cite{Peiravian2013}} in 2013. The authors compared models built using permissions alone, API calls alone, and a combination of both. In each case, they trained an SVM, a decision tree and a bagging classifier. They did not mention the kernel used for \textcolor{black}{the} SVM, nor the base classifier used for bagging, so in our reimplementation we used the default choices of a linear kernel for SVM and decision trees as the base classifier. At the time of their study, the total number of permissions in Android was 130 and the total number of API calls was only 1,326. As shown in Table \ref{tab:PZUS}, the numbers of both have grown since then, and by two orders of magnitude in the case of API calls.
 
\begin{table}[t!]
    \centering
    \resizebox{\columnwidth}{!}{%
    \begin{tabular}{lrrrr}
        \toprule
        \bfseries & \bfseries Benign & \bfseries Malware & \bfseries Permissions & \bfseries API Calls \\ 
        \midrule
        \bfseries Peiravian and Zhu & 1,250 & 1,260 & 130 & 1,326 \\ 
        \bfseries Our dataset & 62,000 & 62,000 & 166 & 134,207 \\ 
        \bottomrule
    \end{tabular}%
    }
    \caption{Comparison of samples and feature counts in Peiravian and Zhu's and our dataset, showing differences in data set size and the increase in feature set sizes}
    \label{tab:PZUS}
\end{table}

\begin{table}[t!]

\centering
\resizebox{\textwidth}{!}{%
\begin{tabular}{@{}lllllllllll@{}}
\toprule
\multicolumn{6}{@{}c@{}}{\textbf{Peiravian and Zhu. Results}}                                                                                                                                                                         & \multicolumn{5}{@{}c@{}}{\textbf{Our Results}}                                                                                                                                              \\ \cmidrule(lr){1-6} \cmidrule(lr){7-11}
\multicolumn{1}{l}{\textbf{Features (Classifier)}} & \multicolumn{1}{l}{\textbf{Accuracy}} & \multicolumn{1}{l}{\textbf{Precision}} & \multicolumn{1}{l}{\textbf{F1-Score}} & \multicolumn{1}{l}{\textbf{TPR}} & \textbf{TNR} & \multicolumn{1}{l}{\textbf{Accuracy}} & \multicolumn{1}{l}{\textbf{Precision}} & \multicolumn{1}{l}{\textbf{F1-Score}} & \multicolumn{1}{l}{\textbf{TPR}}      & \textbf{TNR}      \\  \midrule
\multicolumn{1}{l}{Perm (SVM)}                   & \multicolumn{1}{l}{0.935}             & \multicolumn{1}{l}{0.924}              & \multicolumn{1}{l}{-}                 & \multicolumn{1}{l}{0.875}        & -            & \multicolumn{1}{l}{0.935$ \pm 0.002$} & \multicolumn{1}{l}{0.940$ \pm 0.002$}  & \multicolumn{1}{l}{0.933$ \pm 0.002$} & \multicolumn{1}{l}{0.931$ \pm 0.002$} & 0.941$ \pm 0.001$ \\  
\multicolumn{1}{l}{API (SVM)}                    & \multicolumn{1}{l}{0.958}             & \multicolumn{1}{l}{0.917}              & \multicolumn{1}{l}{-}                 & \multicolumn{1}{l}{\textbf{0.957}}        & -            & \multicolumn{1}{l}{0.953$ \pm 0.004$} & \multicolumn{1}{l}{0.957$ \pm 0.002$}  & \multicolumn{1}{l}{0.954$ \pm 0.001$} & \multicolumn{1}{l}{0.956$ \pm 0.003$} & 0.958$ \pm 0.001$ \\  
\multicolumn{1}{l}{Combined (SVM)}                   & \multicolumn{1}{l}{\textbf{0.969}}             & \multicolumn{1}{l}{0.957}              & \multicolumn{1}{l}{-}                 & \multicolumn{1}{l}{0.948}        & -            & \multicolumn{1}{l}{0.956$ \pm 0.002$} & \multicolumn{1}{l}{0.952$ \pm 0.002$}  & \multicolumn{1}{l}{0.954$ \pm 0.002$} & \multicolumn{1}{l}{0.960$ \pm 0.001$} & 0.952$ \pm 0.001$ \\  
\multicolumn{1}{l}{Perm (DT)}                    & \multicolumn{1}{l}{0.924}             & \multicolumn{1}{l}{0.898}              & \multicolumn{1}{l}{-}                 & \multicolumn{1}{l}{0.866}        & -            & \multicolumn{1}{l}{0.901$ \pm 0.003$} & \multicolumn{1}{l}{0.921$ \pm 0.003$}  & \multicolumn{1}{l}{0.918$ \pm 0.002$} & \multicolumn{1}{l}{0.917$ \pm 0.004$} & 0.921$ \pm 0.001$ \\  
\multicolumn{1}{l}{API (DT)}                     & \multicolumn{1}{l}{0.933}             & \multicolumn{1}{l}{0.894}              & \multicolumn{1}{l}{-}                 & \multicolumn{1}{l}{0.903}        & -            & \multicolumn{1}{l}{0.945$ \pm 0.002$} & \multicolumn{1}{l}{0.950$ \pm 0.002$}  & \multicolumn{1}{l}{0.954$ \pm 0.001$} & \multicolumn{1}{l}{0.947$ \pm 0.002$} & 0.950$ \pm 0.002$ \\  
\multicolumn{1}{l}{Combined (DT)}                    & \multicolumn{1}{l}{0.945}             & \multicolumn{1}{l}{0.906}              & \multicolumn{1}{l}{-}                 & \multicolumn{1}{l}{0.928}        & -            & \multicolumn{1}{l}{0.949$ \pm 0.002$} & \multicolumn{1}{l}{0.949$ \pm 0.003$}  & \multicolumn{1}{l}{0.949$ \pm 0.001$} & \multicolumn{1}{l}{0.950$ \pm 0.001$} & 0.949$ \pm 0.002$ \\  
\multicolumn{1}{l}{Perm (Bagging)}               & \multicolumn{1}{l}{0.936}             & \multicolumn{1}{l}{0.920}              & \multicolumn{1}{l}{-}                 & \multicolumn{1}{l}{0.882}        & -            & \multicolumn{1}{l}{0.930$ \pm 0.002$} & \multicolumn{1}{l}{0.932$ \pm 0.001$}  & \multicolumn{1}{l}{0.933$ \pm 0.003$} & \multicolumn{1}{l}{0.931$ \pm 0.001$} & 0.932$ \pm 0.002$ \\  
\multicolumn{1}{l}{API (Bagging)}                & \multicolumn{1}{l}{0.949}             & \multicolumn{1}{l}{0.936}              & \multicolumn{1}{l}{-}                 & \multicolumn{1}{l}{0.907}        & -            & \multicolumn{1}{l}{\textbf{0.961}$ \pm 0.002$} & \multicolumn{1}{l}{\textbf{0.960}$ \pm 0.003$}  & \multicolumn{1}{l}{\textbf{0.959}$ \pm 0.002$} & \multicolumn{1}{l}{\textbf{0.961}$ \pm 0.002$} & \textbf{0.960}$ \pm 0.002$ \\  
\multicolumn{1}{l}{Combined (Bagging)}               & \multicolumn{1}{l}{0.964}             & \multicolumn{1}{l}{\textbf{0.949}}              & \multicolumn{1}{l}{-}                 & \multicolumn{1}{l}{0.941}        & -            & \multicolumn{1}{l}{0.948$ \pm 0.001$} & \multicolumn{1}{l}{0.949$ \pm 0.002$}  & \multicolumn{1}{l}{0.945$ \pm 0.002$} & \multicolumn{1}{l}{0.946$ \pm 0.001$} & 0.948$ \pm 0.002$ \\  \bottomrule
\end{tabular}%
}
 \caption{Result comparison between Peiravian and Zhu \cite{Peiravian2013} and our reimplementation, showing the effect of feature type (Perm: permissions alone, API: API calls alone, Combined: permissions and API calls) and ML model choice on performance metrics. \textcolor{black}{For this and subsequent results tables, the figures on the left are reproduced from the original publication, and those on the right are from our reimplementation. Where values on the left are missing (-), this indicates that these metrics were not published in the original study.}}
 \label{tab:per}
\end{table}
 
Table \ref{tab:per} reports the results both for Peiravian and Zhu's original study and for our reimplementation using our much larger contemporary dataset. First of all, it is interesting to see that, despite the much larger feature and data set sizes, our results cover a similar range of accuracies. Peiravian and Zhu provided figures for accuracy, precision and recall in their evaluation. For completeness, we report the full set of metrics for our reimplementation.

Peiravian and Zhu found that models based only on API calls outperformed models based only on permissions. The figures in Table \ref{tab:per} show that this also appears to be the case for our reimplementation, with an even larger margin than in the original study.

We analysed the differences between models further by applying statistical tests. First, Kruskal-Wallis showed that there are significant differences in the accuracies reported in Table \ref{tab:per} (H = 50.04, p $<$ 0.05). Post hoc Dunn's tests then showed that (i) the bagging and DT models using API calls perform better, on average, than all models that use permissions, and (ii) SVM models using API calls perform better, on average, than SVM models that use permissions. This supports the conclusion that API calls are more useful than permissions for building Android malware detection models.

All other group-wise differences were not significant at the 95\% confidence level, suggesting that the exact choice of ML model is less important. However, we note that, for API calls, the bagging model had the best performance across all metrics, achieving an average accuracy of 96\%.

\subsection{\textcolor{black}{Selection of permissions feature sets}} 

Nevertheless, permissions are easier to extract than other static features. For this reason, they are widely used in practice. To provide more insight into the best way to use permissions for Android malware detection, we reimplemented a study by \textbf{Wang et al.\ \cite{Wang2014}} that aimed to provide a more fine-grained view of the risk associated with permissions. Their study used two groups of Android permissions to build ML models: normal permissions, which are automatically granted by the system, and dangerous permissions, which provide access to private data or allow control over the device. For more information about these categories, see \cite{android_permission}. \textcolor{black}{Table \ref{tab:WUS} shows the number of normal and dangerous permissions as tagged by Android compared to the full permission set as shown in Table \ref{tab:PZUS}.}

Their study used four feature selection algorithms to identify the permissions that are most useful for building ML models: MI, PCC, sequential forward selection (SFS), and T-tests. They also used principal component analysis (PCA) as an alternative means of dimensionality reduction. In our reimplementation, we trained our four base ML models using feature sets derived both from the feature reduction methods used by Wang et al.\ and also our standard set of feature selection methods. We generated feature sets of sizes 10 to 70 features using each method.

 
\begin{table}[t]
 
    \centering
  
    \begin{tabular}{lrrr}
   \toprule
    \bf & \bf Benign  & \bf Malware & \bf Permissions \\  
 \midrule
        \bf Wang et al.\ & 310,926 & 4,868 & 88  \\ 
        \bf Our dataset &  62,000 & 62,000 & 82  \\ 
 \bottomrule
    \end{tabular}
       \caption{\textcolor{black}{Comparison of samples and total number of normal and dangerous permissions in Wang et al's and our dataset}}
    \label{tab:WUS}
 
\end{table}

\begin{table}[t]
 
\centering
\resizebox{\textwidth}{!}{%
\begin{tabular}{llllllllllll}
\toprule
\multicolumn{6}{c}{\textbf{Wang et al.\ Results}}                                                                                                                                                                 & \multicolumn{6}{c}{\textbf{Our Results}}                                                                                                                                                                                     \\ \cmidrule(lr){1-6} \cmidrule(lr){7-12}
\multicolumn{1}{l}{\textbf{Method}} & \multicolumn{1}{l}{\textbf{Accuracy}} & \multicolumn{1}{l}{\textbf{Precision}} & \multicolumn{1}{l}{\textbf{F1-Score}} & \multicolumn{1}{l}{\textbf{TPR}} & \textbf{TNR} & \multicolumn{1}{l}{\textbf{Method}} & \multicolumn{1}{l}{\textbf{Accuracy}} & \multicolumn{1}{l}{\textbf{Precision}} & \multicolumn{1}{l}{\textbf{F1-Score}} & \multicolumn{1}{l}{\textbf{TPR}}      & \textbf{TNR}      \\  \midrule
\multicolumn{1}{l}{MI}         & \multicolumn{1}{l}{0.996}                 & \multicolumn{1}{l}{-}                  & \multicolumn{1}{l}{0.895}                 & \multicolumn{1}{l}{0.923}        & -            & \multicolumn{1}{l}{MI (RF)}    & \multicolumn{1}{l}{\textbf{0.930}$ \pm 0.001$} & \multicolumn{1}{l}{\textbf{0.932}$ \pm 0.001$}  & \multicolumn{1}{l}{0.929$ \pm 0.002$} & \multicolumn{1}{l}{0.931$ \pm 0.002$} &0.931$ \pm 0.003$ \\ 
\multicolumn{1}{l}{PCC}        & \multicolumn{1}{l}{0.996}                 & \multicolumn{1}{l}{-}                  & \multicolumn{1}{l}{0.895}                 & \multicolumn{1}{l}{0.923}        & -            & \multicolumn{1}{l}{PCC (RF)}   & \multicolumn{1}{l}{0.907$ \pm 0.002$} & \multicolumn{1}{l}{0.910$ \pm 0.003$}  & \multicolumn{1}{l}{0.905$ \pm 0.002$} & \multicolumn{1}{l}{0.902$ \pm 0.003$} & 0.911$ \pm 0.001$ \\ 
\multicolumn{1}{l}{T-Test}          & \multicolumn{1}{l}{0.996}                 & \multicolumn{1}{l}{-}                  & \multicolumn{1}{l}{0.895}                 & \multicolumn{1}{l}{0.923}        & -            & \multicolumn{1}{l}{T-Test  (RF)}    & \multicolumn{1}{l}{0.919$ \pm 0.001$} & \multicolumn{1}{l}{0.920$ \pm 0.002$}  & \multicolumn{1}{l}{0.921$ \pm 0.002$} & \multicolumn{1}{l}{0.919$ \pm 0.002$} & 0.921$ \pm 0.001$ \\ 
\multicolumn{1}{l}{SFS}             & \multicolumn{1}{l}{0.996}                 & \multicolumn{1}{l}{-}                  & \multicolumn{1}{l}{0.895}                 & \multicolumn{1}{l}{0.923}        & -            & \multicolumn{1}{l}{SFS (SVM)}       & \multicolumn{1}{l}{0.902$ \pm 0.002$} & \multicolumn{1}{l}{0.902$ \pm 0.003$}  & \multicolumn{1}{l}{0.901$ \pm 0.002$} & \multicolumn{1}{l}{0.901$ \pm 0.001$} & 0.902$ \pm 0.002$ \\ 
\multicolumn{1}{l}{PCA}             & \multicolumn{1}{l}{0.996}                 & \multicolumn{1}{l}{-}                  & \multicolumn{1}{l}{0.895}                 & \multicolumn{1}{l}{\textbf{0.926}}        & -            & \multicolumn{1}{l}{PCA (SVM)}       & \multicolumn{1}{l}{0.928$ \pm 0.001$} & \multicolumn{1}{l}{0.926$ \pm 0.002$}  & \multicolumn{1}{l}{0.927$ \pm 0.003$} & \multicolumn{1}{l}{\textbf{0.932}$ \pm 0.001$} & 0.925$ \pm 0.002$ \\ 

\multicolumn{1}{l}{Chi-square} & \multicolumn{1}{l}{-}        & \multicolumn{1}{l}{-}         & \multicolumn{1}{l}{-}        & \multicolumn{1}{l}{-}     & -                             & \multicolumn{1}{l}{Chi-square(RF)} & \multicolumn{1}{l}{0.927$ \pm 0.002$} & \multicolumn{1}{l}{0.923$ \pm 0.003$} & \multicolumn{1}{l}{\textbf{0.946}$ \pm 0.002$} & \multicolumn{1}{l}{0.901$ \pm 0.003$} & \textbf{0.949}$ \pm 0.003$                           \\ \bottomrule
\end{tabular}%
}
 \caption{Result comparison between Wang et al.\ \cite{Wang2014} and our reimplementation, \textcolor{black}{showing the effect of feature selection algorithm choice. The original study used SVM; for the results of our reimplementation, we also show the best-performing model in each case. Accuracy and F1-scores for the original study are approximate values as they were read from plots provided by the authors.}}
 \label{tab:WRES}
\end{table}

Table \ref{tab:WRES} shows the best results for each of Wang et al's feature reduction methods, and for chi-square, which was our overall best performing feature selection method in terms of F1-score. The results from Wang et al.\ are for SVM models. For our results, we also indicate the best performing ML model for each feature reduction method. Our models performed significantly better than Wang et al's in terms of F1-score. Although Wang et al.\ reported very high accuracies in their study, this is likely to be a reflection of their very imbalanced data set, and possibly an indication that their models overfit the majority class. This would also explain why their F1-scores were relatively low.

For our models, a positive Kruskal-Wallis test (H = 30.62, p $<$ 0.05) followed by pairwise Dunn's tests showed that the only significant difference in mean accuracies was between the SVM trained on the SFS feature set and the other models. This does not allow us to draw any firm conclusions about which model or feature selection method is best.
 
\begin{figure}[t!]
\includegraphics[width=\columnwidth]{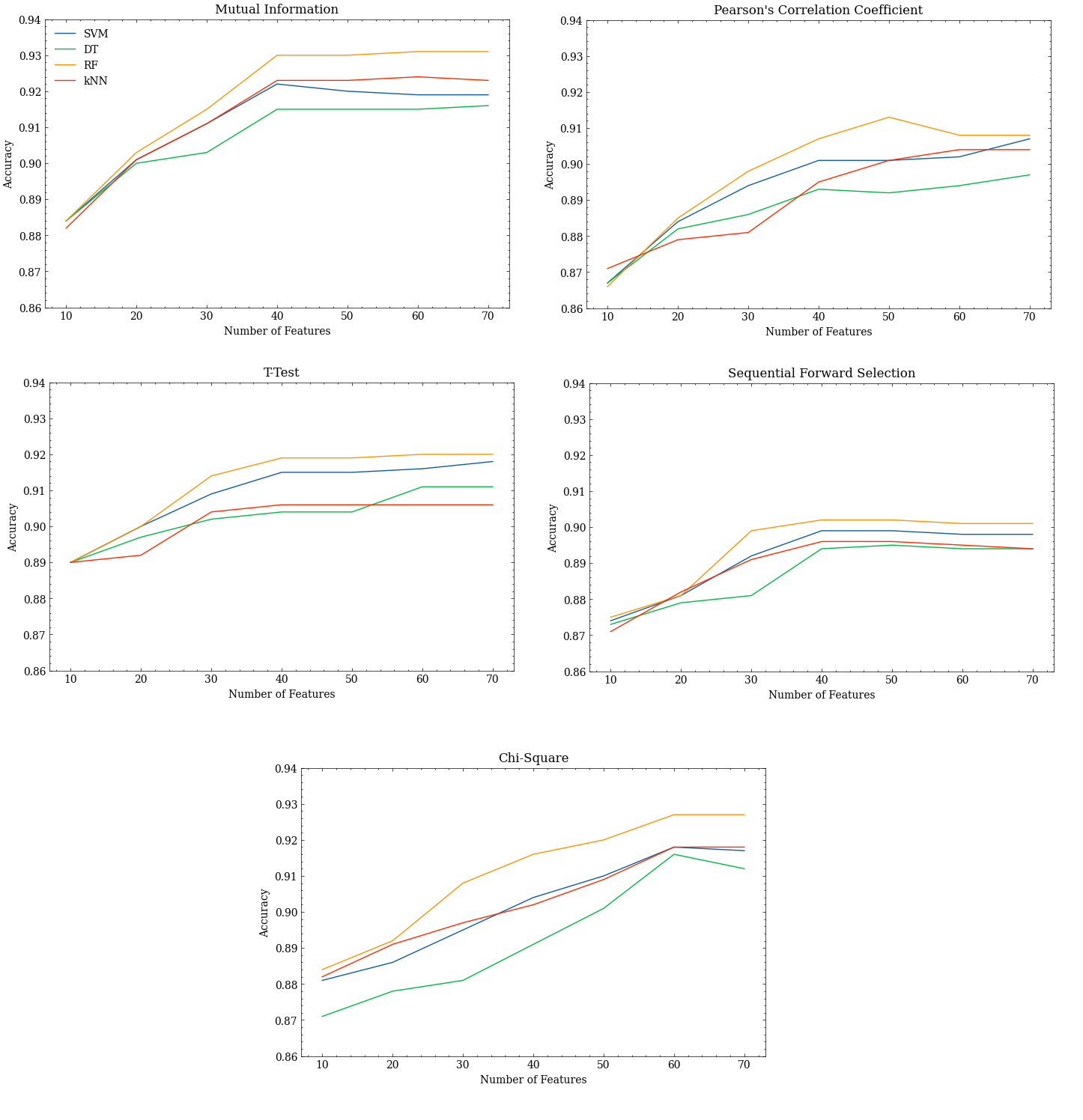}
\centering
\caption{\textcolor{black}{Relationship between number of permissions used and model accuracy when permissions are ranked using different methods}}
\label{fig:WangGraph}
\end{figure}

Fig. \ref{fig:WangGraph} shows how the accuracy of each model changes based on the size of the feature set. It can be seen that, for all models and feature selection methods, accuracy initially improves as the number of features is increased, but then reaches a plateau point. The best overall accuracy was achieved by RF and SVM models when using 40 permissions, and using more than 40 permissions did not lead to improvement in most cases.
A notable difference from the findings of Wang et al.\ is that this plateau point comes later, which may indicate that more permissions are now required to build reliable malware detection models. Interestingly, the lists of the top 40 permissions produced by MI, PCC and T-test had 38 permissions in common across all the CV test folds. For chi-square, the best performing feature set, of 60 permissions, included all the permissions chosen by these other feature selection algorithms. Moreover, the models trained using chi-square permissions plateau later than the other feature selection models and so require more permissions to achieve high accuracy. The best chi-square model also achieved the highest F1-score.  


\textbf{Rathore et al.\ \cite{Rathore2021}} further experimented with different feature selection and feature dimensionality reduction methods, with the aim of building an efficient model that used a minimal set of permissions. Apart from the original dataset with full permissions (OD), they used VT, PCA, and autoencoders (with one and three layers: AE-1L,AE-3L) to construct reduced feature sets. To this, we added our core set of feature selection algorithms. Unlike Wang et al., they carried out feature selection on the entire set of Android permissions, not just the normal and dangerous permissions. They trained six ML models for each of the feature selection/reduction methods: DT, kNN, SVM, RF, AdaBoost and DNNs (with two, four and seven layers). These were compared against models trained on the full set of 166 permissions.

\begin{table}[t]

\centering
\resizebox{\textwidth}{!}{%
 
\begin{tabular}{lllllllllllll}
\toprule
\textbf{}           & \multicolumn{6}{c}{\textbf{Rathore et al.\ Results}}                                                                                                                                                                & \multicolumn{6}{c}{\textbf{Our Results}}                                                                                                                                                                                     \\ \cmidrule(lr){2-7} \cmidrule(lr){8-13}
\textbf{Classifier} & \multicolumn{1}{l}{\textbf{Method}} & \multicolumn{1}{l}{\textbf{Accuracy}} & \multicolumn{1}{l}{\textbf{Precision}} & \multicolumn{1}{l}{\textbf{F1-Score}} & \multicolumn{1}{l}{\textbf{TPR}} & \textbf{TNR} & \multicolumn{1}{l}{\textbf{Method}} & \multicolumn{1}{l}{\textbf{Accuracy}} & \multicolumn{1}{l}{\textbf{Precision}} & \multicolumn{1}{l}{\textbf{F1-Score}} & \multicolumn{1}{l}{\textbf{TPR}}      & \textbf{TNR}      \\ \midrule
DT      & \multicolumn{1}{l}{OD}              & \multicolumn{1}{l}{0.926}             & \multicolumn{1}{l}{-}                  & \multicolumn{1}{l}{-}                 & \multicolumn{1}{l}{0.919}        & -            & \multicolumn{1}{l}{OD}              & \multicolumn{1}{l}{0.901$ \pm 0.003$} & \multicolumn{1}{l}{0.921$ \pm 0.003$}  & \multicolumn{1}{l}{0.918$ \pm 0.002$} & \multicolumn{1}{l}{0.917$ \pm 0.004$} & 0.921$ \pm 0.001$ \\ 
kNN                 & \multicolumn{1}{l}{PCA}             & \multicolumn{1}{l}{0.914}             & \multicolumn{1}{l}{-}                  & \multicolumn{1}{l}{-}                 & \multicolumn{1}{l}{0.913}        & -            & \multicolumn{1}{l}{MI}              & \multicolumn{1}{l}{0.932$ \pm 0.002$} & \multicolumn{1}{l}{0.928$ \pm 0.001$}  & \multicolumn{1}{l}{0.931$ \pm 0.002$} & \multicolumn{1}{l}{0.935$ \pm 0.001$} & 0.928$ \pm 0.002$ \\ 
SVM                 & \multicolumn{1}{l}{OD}              & \multicolumn{1}{l}{0.910}             & \multicolumn{1}{l}{-}                  & \multicolumn{1}{l}{-}                 & \multicolumn{1}{l}{0.894}        & -            & \multicolumn{1}{l}{AE-3L}           & \multicolumn{1}{l}{0.933$ \pm 0.002$} & \multicolumn{1}{l}{0.927$ \pm 0.002$}  & \multicolumn{1}{l}{0.935$ \pm 0.002$} & \multicolumn{1}{l}{0.942$ \pm 0.003$} & 0.927$ \pm 0.001$ \\ 
RF       & \multicolumn{1}{l}{OD}              & \multicolumn{1}{l}{\textbf{0.940}}             & \multicolumn{1}{l}{-}                  & \multicolumn{1}{l}{-}                 & \multicolumn{1}{l}{0.930}        & -            & \multicolumn{1}{l}{VT}              & \multicolumn{1}{l}{\textbf{0.949}$ \pm 0.003$} & \multicolumn{1}{l}{\textbf{0.948}$ \pm 0.002$}  & \multicolumn{1}{l}{\textbf{0.950}$ \pm 0.001$} & \multicolumn{1}{l}{\textbf{0.951}$ \pm 0.001$} & \textbf{0.949}$ \pm 0.002$ \\ 
AdaBoost            & \multicolumn{1}{l}{AE-1L}           & \multicolumn{1}{l}{0.911}             & \multicolumn{1}{l}{-}                  & \multicolumn{1}{l}{-}                 & \multicolumn{1}{l}{0.898}        & -            & \multicolumn{1}{l}{OD}              & \multicolumn{1}{l}{0.928$ \pm 0.003$} & \multicolumn{1}{l}{0.922$ \pm 0.001$}  & \multicolumn{1}{l}{0.927$ \pm 0.001$} & \multicolumn{1}{l}{0.934$ \pm 0.001$} & 0.921$ \pm 0.002$ \\ 
DNN-2L              & \multicolumn{1}{l}{AE-1L}           & \multicolumn{1}{l}{0.931}             & \multicolumn{1}{l}{-}                  & \multicolumn{1}{l}{-}                 & \multicolumn{1}{l}{0.914}        & -            & \multicolumn{1}{l}{MI}           & \multicolumn{1}{l}{0.930$ \pm 0.002$} & \multicolumn{1}{l}{0.922$ \pm 0.003$}  & \multicolumn{1}{l}{0.933$ \pm 0.002$} & \multicolumn{1}{l}{0.938$ \pm 0.003$} & 0.921$ \pm 0.001$ \\ 
DNN-4L              & \multicolumn{1}{l}{VT}              & \multicolumn{1}{l}{0.931}             & \multicolumn{1}{l}{-}                  & \multicolumn{1}{l}{-}                 & \multicolumn{1}{l}{0.920}        & -            & \multicolumn{1}{l}{PCA}             & \multicolumn{1}{l}{0.932$ \pm 0.001$} & \multicolumn{1}{l}{0.923$ \pm 0.002$}  & \multicolumn{1}{l}{0.931$ \pm 0.001$} & \multicolumn{1}{l}{0.941$ \pm 0.003$} & 0.922$ \pm 0.002$ \\ 
DNN-7L              & \multicolumn{1}{l}{OD}              & \multicolumn{1}{l}{0.930}             & \multicolumn{1}{l}{-}                  & \multicolumn{1}{l}{-}                 & \multicolumn{1}{l}{0.928}        & -            & \multicolumn{1}{l}{OD}              & \multicolumn{1}{l}{0.927$ \pm 0.002$} & \multicolumn{1}{l}{0.922$ \pm 0.001$}  & \multicolumn{1}{l}{0.925$ \pm 0.002$} & \multicolumn{1}{l}{0.933$ \pm 0.003$} & 0.921$ \pm 0.002$ \\ \bottomrule
\end{tabular}%
}
 \caption{Result comparison between Rathore et al.\ \cite{Rathore2021} and our reimplementation, \textcolor{black}{showing the effect of model choice for permissions-based models, and the most effective feature selection method for reducing the number of permissions for each ML model type}}
 \label{tab:RUS}
\end{table}

\begin{figure}[t]
\centering
\includegraphics[width=12cm]{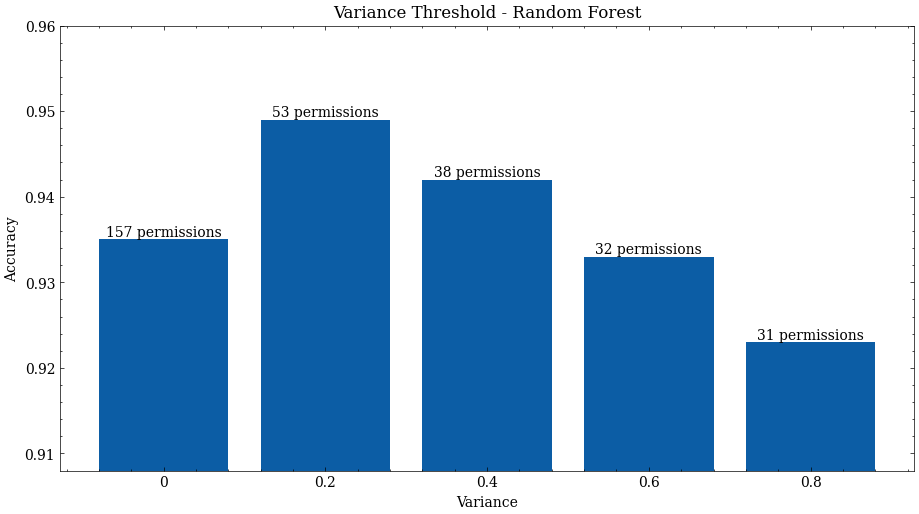}
\caption{\textcolor{black}{Effect of changing the variance threshold upon accuracy of random forest models. Also shows the number of permissions selected for different variance thresholds.}}
\label{fig:RUS}
\end{figure}

Table \ref{tab:RUS} summarises the performance metrics from our reimplementation, showing the figures for the most effective feature reduction method for each ML model. Our results are generally quite similar to those from the original study. Like Rathore et al., we found RFs to be the most effective model, with the highest mean values across all metrics. This is supported by a statistical analysis of the results in Table \ref{tab:RUS}, where Dunn's tests following a positive Kruskal-Wallis test (H = 54.58, p $<$ 0.05) indicated that RF accuracy was significantly higher than all other models except kNN. Our results also show that there is no particular value in reducing the set of permissions to those which are normal and dangerous prior to carrying out feature selection. 



We found that RFs work best when combined with variance threshold. Digging down a bit further, Fig. \ref{fig:RUS} shows the effect of changing the threshold, indicating that lower thresholds work better. For a threshold of 0.2, the number of permissions can be reduced by 64\%, whilst still producing better malware detection than when using all permissions.

\textcolor{black}{We reimplemented one more study that used only permissions. This study, by \textbf{Sahin et al.\ \cite{Sahin2023}}, is notable for using model-based feature selection. Unlike the more traditional forms of feature selection used in the previous studies, model-based feature selection uses an ML model to determine the importance of each feature. This can lead to more appropriate feature combinations; however, it does have a cost, and Sahin et al. attempt to mitigate against this by using a relatively inexpensive model, namely multiple regression, to assess features.}




\begin{table}[t]

\centering
\resizebox{\textwidth}{!}{%
\textcolor{black}{
\begin{tabular}{lllllllllllll}
\toprule
&\multicolumn{6}{c}{\textbf{Sahin et al. Results}}                                                                                                                                                                                                             & \multicolumn{6}{c}{\textbf{Our Results}}                                                                                                                                                                                          \\ \cmidrule(lr){2-7} \cmidrule(lr){8-13}
\multicolumn{1}{l}{\textbf{Classifier}} & \multicolumn{1}{l}{\textbf{Permissions}} & \multicolumn{1}{l}{\textbf{Accuracy}} & \multicolumn{1}{l}{\textbf{Precision}} & \multicolumn{1}{l}{\textbf{F1-Score}} & \multicolumn{1}{l}{\textbf{TPR}} & \textbf{TNR} & \multicolumn{1}{l}{\textbf{Permissions}} & \multicolumn{1}{l}{\textbf{Accuracy}} & \multicolumn{1}{l}{\textbf{Precision}} & \multicolumn{1}{l}{\textbf{F1-Score}} & \multicolumn{1}{l}{\textbf{TPR}}      & \textbf{TNR}      \\ \midrule
\multicolumn{1}{l}{SVM}            & \multicolumn{1}{l}{-}                    & \multicolumn{1}{l}{-}                 & \multicolumn{1}{l}{-}                  & \multicolumn{1}{l}{-}                 & \multicolumn{1}{l}{-}            & -            & \multicolumn{1}{l}{60}                   & \multicolumn{1}{l}{0.914$ \pm 0.001$} & \multicolumn{1}{l}{0.912$ \pm 0.001$}  & \multicolumn{1}{l}{0.913$ \pm 0.003$} & \multicolumn{1}{l}{0.921$ \pm 0.003$} & 0.911$ \pm 0.003$ \\ 
\multicolumn{1}{l}{RF}  & \multicolumn{1}{l}{27}                   & \multicolumn{1}{l}{-}                 & \multicolumn{1}{l}{-}                  & \multicolumn{1}{l}{0.930}             & \multicolumn{1}{l}{-}            & -            & \multicolumn{1}{l}{80}                   & \multicolumn{1}{l}{\textbf{0.924}$ \pm 0.003$} & \multicolumn{1}{l}{\textbf{0.920}$ \pm 0.003$}  & \multicolumn{1}{l}{\textbf{0.925}$ \pm 0.003$} & \multicolumn{1}{l}{\textbf{0.932}$ \pm 0.003$} & 0.921$ \pm 0.001$ \\ 
\multicolumn{1}{l}{DT} & \multicolumn{1}{l}{27}                   & \multicolumn{1}{l}{-}                 & \multicolumn{1}{l}{-}                  & \multicolumn{1}{l}{\textbf{0.956}}             & \multicolumn{1}{l}{-}            & -            & \multicolumn{1}{l}{80}                   & \multicolumn{1}{l}{0.909$ \pm 0.001$} & \multicolumn{1}{l}{0.914$ \pm 0.001$}  & \multicolumn{1}{l}{0.911$ \pm 0.001$} & \multicolumn{1}{l}{0.909$ \pm 0.003$} & 0.911$ \pm 0.001$ \\ 
\multicolumn{1}{l}{kNN}            & \multicolumn{1}{l}{43}                   & \multicolumn{1}{l}{-}                 & \multicolumn{1}{l}{-}                  & \multicolumn{1}{l}{0.954}             & \multicolumn{1}{l}{-}            & -            & \multicolumn{1}{l}{60}                   & \multicolumn{1}{l}{0.912$ \pm 0.003$} & \multicolumn{1}{l}{0.912$ \pm 0.003$}  & \multicolumn{1}{l}{0.907$ \pm 0.003$} & \multicolumn{1}{l}{0.911$ \pm 0.001$} & \textbf{0.922}$ \pm 0.003$ \\ 
\multicolumn{1}{l}{Naïve Bayes}    & \multicolumn{1}{l}{43}                   & \multicolumn{1}{l}{-}                 & \multicolumn{1}{l}{-}                  & \multicolumn{1}{l}{0.928}             & \multicolumn{1}{l}{-}            & -            & \multicolumn{1}{l}{20}                   & \multicolumn{1}{l}{0.817$ \pm 0.003$} & \multicolumn{1}{l}{0.824$ \pm 0.002$}  & \multicolumn{1}{l}{0.866$ \pm 0.002$} & \multicolumn{1}{l}{0.913$ \pm 0.001$} & 0.905$ \pm 0.003$ \\ \bottomrule
\end{tabular}%
}}
 \caption{\textcolor{black}{Result comparison between Sahin et al.\ \cite{Sahin2023} and our reimplementation, showing the effect of model choice when permissions-based features are selected using a model-based feature selection algorithm}}
 \label{tab:SUS}
\end{table}

\textcolor{black}{
Table \ref{tab:SUS} summarises the results of our reimplementation. These show that using a model-based feature selection approach based around multiple regression does not lead to an improvement in classification performance. This conclusion is backed by a statistical analysis (Kruskal-Wallis test (H = 31.22, p $<$ 0.05) of the results in Table \ref{tab:SUS} and the findings from the top-performing permission-based models listed in Table \ref{tab:PZUS}. It is possible that better performance could be obtained by matching the model used for classification with the model used for feature selection; however, this benefit was not observed in the original study, and the use of more expensive models for feature selection would create challenges for the larger feature sets considered in the remainder of this study.
}

\textcolor{black}{To conclude, reasonable levels of performance can be achieved with permissions-based approaches (see \cref{tab:WRES,tab:RUS,tab:SUS}), especially when the number of permissions is reduced using a feature selection algorithm, and RFs are used as the ML model. However, none of the permissions-based models we looked at were competitive against API call-based models (see table \ref{tab:per}, and also the results in Section \ref{representations}), with statistical analysis (Kruskal–Wallis H = 50.04, p $<$ 0.05) confirming that these differences are significant.} 


\subsection{\textcolor{black}{Comparison of API call representations}} \label{representations}
 
Next, we take a closer look at the best way of building models based on API calls, beginning with a reimplementation of the research carried out by \textbf{Ma et al.\ \cite{Ma2019}}, who considered three different ways of representing feature usage within an application: API usage, API frequency and API sequences.

For API usage, each API call is represented by a binary feature indicating whether the call is used within an application. We created our version of this dataset by using the complete set of 134,207 API calls; i.e., each application is represented by a feature vector $A=\{API_1, API_2, API_3, \dots API_{134207}\}$, making this a high-dimensional feature set. This is the same dataset used for our earlier reimplementation of Peiravian et al.\ \cite{Peiravian2013}. The API frequency data set is similar, except that each feature has an integer value, and indicates how many times each call is made within an application. For example, the feature vector for an application would be $M=\{5,4,0\}$ if $A=\{API_1, API_2, API_3 \}$ and the application used $API_1$ 5 times, $API_2$ 4 times and $API_3$ 0 times.

For API sequences, each application is represented as a variable-length sequence of API calls. In our reimplementation, a call graph is extracted using the FLOWDROID module of our static analysis tool. The tool stores the APIs in chronological order and hence allows us to gather information on the order in which the application calls the APIs. We developed an algorithm which uses depth-first search (DFS) to obtain a set of routes for an application. The last API of every route is usually a system API call as user methods and user-defined APIs will always call system APIs. Therefore, we were able to extract system APIs that the application called in chronological order making an API sequence dataset. The resulting API system calls sequence is then converted into a numerical representation. For example, if we have a set of APIs $A=\{a,b,c,d \}$ then the applications are subsets of $A$. If $M=[a,c,a,d]$ is a malware application and $B=[d,c,b,a]$ is a benign application, then the vector of $M$ will be $\{1,3,1,4 \}$ and the vector of $B$ will be $\{4,3,2,4 \}$, where 1 represents $a$, 2 represents $b$, etc.

\begin{table}[t]

\centering
\resizebox{\textwidth}{!}{%
\begin{tabular}{lllllllllllll}
\toprule
&\multicolumn{6}{c}{\textbf{Ma   et al.\ Results}}                                                                                                                                                                                                          & \multicolumn{6}{c}{\textbf{Our Results}}                                                                                                                                                                                                      \\  \cmidrule(lr){2-7} \cmidrule(lr){8-13}
\multicolumn{1}{l}{\textbf{Dataset}} & \multicolumn{1}{l}{\textbf{Classifier}} & \multicolumn{1}{l}{\textbf{Accuracy}} & \multicolumn{1}{l}{\textbf{Precision}} & \multicolumn{1}{l}{\textbf{F1-Score}} & \multicolumn{1}{l}{\textbf{TPR}} & \textbf{TNR} & \multicolumn{1}{l}{\textbf{Classifier}} & \multicolumn{1}{l}{\textbf{Accuracy}} & \multicolumn{1}{l}{\textbf{Precision}} & \multicolumn{1}{l}{\textbf{F1-Score}}                   & \multicolumn{1}{l}{\textbf{TPR}}      & \textbf{TNR}      \\ 
\midrule
\multicolumn{1}{l}{Usage}            & \multicolumn{1}{l}{DT}             & \multicolumn{1}{l}{-}                 & \multicolumn{1}{l}{0.968}              & \multicolumn{1}{l}{0.965}             & \multicolumn{1}{l}{0.962}        & -            & \multicolumn{1}{l}{RF}             & \multicolumn{1}{l}{0.966$ \pm 0.002$} & \multicolumn{1}{l}{0.968$ \pm 0.002$}  & \multicolumn{1}{l}{0.966$ \pm 0.002$} & \multicolumn{1}{l}{0.964$ \pm 0.003$} & 0.969$ \pm 0.001$ \\ 
\multicolumn{1}{l}{Frequency}        & \multicolumn{1}{l}{DNN(8)}         & \multicolumn{1}{l}{-}                 & \multicolumn{1}{l}{0.977}              & \multicolumn{1}{l}{0.974}            & \multicolumn{1}{l}{0.971}        & -            & \multicolumn{1}{l}{DNN(16)}       & \multicolumn{1}{l}{\textbf{0.970}$ \pm 0.001$} & \multicolumn{1}{l}{\textbf{0.971}$ \pm 0.001$}  & \multicolumn{1}{l}{\textbf{0.969}$ \pm 0.002$}                   & \multicolumn{1}{l}{\textbf{0.979}$ \pm 0.002$} & \textbf{0.971}$ \pm 0.002$ \\ 
\multicolumn{1}{l}{Sequence}         & \multicolumn{1}{l}{LSTM(8)}         & \multicolumn{1}{l}{-}                 & \multicolumn{1}{l}{0.985}              & \multicolumn{1}{l}{0.986}             & \multicolumn{1}{l}{\textbf{0.988}}        & -            & \multicolumn{1}{l}{LSTM(8)}        & \multicolumn{1}{l}{0.935$ \pm 0.003$} & \multicolumn{1}{l}{0.935$ \pm 0.002$}  & \multicolumn{1}{l}{0.936$ \pm 0.001$}                   & \multicolumn{1}{l}{0.936$ \pm 0.002$} & 0.935$ \pm 0.001$ \\ 
\multicolumn{1}{l}{Ensemble}         & \multicolumn{1}{l}{Voting}         & \multicolumn{1}{l}{-}                 & \multicolumn{1}{l}{\textbf{0.991}}              & \multicolumn{1}{l}{\textbf{0.990}}              & \multicolumn{1}{l}{\textbf{0.988}}        & -            & \multicolumn{1}{l}{Voting}         & \multicolumn{1}{l}{0.965$ \pm 0.002$} & \multicolumn{1}{l}{0.963$ \pm 0.003$}  & \multicolumn{1}{l}{0.962$ \pm 0.003$} & \multicolumn{1}{l}{0.967$ \pm 0.002$} & 0.963$ \pm 0.002$ \\ 
\bottomrule
\end{tabular}%
}
 \caption{Result comparison between Ma et al.\ \cite{Ma2019} and our reimplementation, \textcolor{black}{comparing the four different API call modelling approaches used in the original study}}
 \label{tab:APIUsage}
 
\end{table}

\begin{table}[t]

\centering
\resizebox{\columnwidth}{!}{%
\begin{tabular}{lllllll}
\toprule
\textbf{API Dataset} & \textbf{Layers} & \textbf{Accuracy}  & \textbf{Precision} & \textbf{F1-Score} & \textbf{TPR}      & \textbf{TNR}      \\  \midrule
Usage                                          & 4 (DNN)                                   & 0.959$ \pm 0.002$                                        & 0.958$ \pm 0.002$                                         & 0.958$ \pm 0.002$                                        & 0.957$ \pm 0.003$                                        & 0.954$ \pm 0.002$                                        \\ 
Usage                                          & 8 (DNN)                                   & 0.953$ \pm 0.002$                                        & 0.952$ \pm 0.003$                                         & 0.954$ \pm 0.002$                                        & 0.955$ \pm 0.003$                                        & 0.954$ \pm 0.002$                                        \\ 
Usage                                          & 16 (DNN)                                  & 0.961$ \pm 0.001$                           & 0.962$ \pm 0.003$                            & 0.963$ \pm 0.002$                           & 0.959$ \pm 0.001$                           & 0.962$ \pm 0.001$                           \\ 
Usage                                          & 32 (DNN)                                  & 0.960$ \pm 0.001$                                        & 0.959$ \pm 0.002$                                         & 0.961$ \pm 0.002$                                        & 0.958$ \pm 0.003$                                        & 0.961$ \pm 0.001$                                        \\ 
Frequency            & 4 (DNN)         & 0.961$ \pm 0.001$ & 0.955$ \pm 0.002$  & 0.958$ \pm 0.002$ & 0.963$ \pm 0.002$ & 0.956$ \pm 0.002$ \\ 
Frequency            & 8 (DNN)         & 0.967$ \pm 0.002$  & 0.968$ \pm 0.003$  & 0.966$ \pm 0.002$ & 0.966$ \pm 0.002$ & 0.968$ \pm 0.002$ \\ 
Frequency            & 16 (DNN)        & \textbf{0.970}$ \pm 0.001$  & \textbf{0.971}$ \pm 0.002$  & \textbf{0.969}$ \pm 0.002$ & \textbf{0.969}$ \pm 0.002$ & \textbf{0.971}$ \pm 0.002$ \\ 
Frequency            & 32 (DNN)        & 0.960$ \pm 0.002$  & 0.961$ \pm 0.002$  & 0.961$ \pm 0.002$ & 0.962$ \pm 0.002$ & 0.962$ \pm 0.002$ \\ 
Sequence             & 4 (LSTM)        & 0.888$ \pm 0.002$  & 0.890$ \pm 0.002$  & 0.884$ \pm 0.002$ & 0.881$ \pm 0.001$ & 0.891$ \pm 0.003$ \\ 
Sequence             & 8 (LSTM)        & 0.936$ \pm 0.003$ & 0.935$ \pm 0.004$  & 0.934$ \pm 0.001$ & 0.936$ \pm 0.002$ & 0.935$ \pm 0.001$ \\ 
Sequence             & 16 (LSTM)       & 0.929$ \pm 0.001$  & 0.930$ \pm 0.002$  & 0.926$ \pm 0.002$ & 0.927$ \pm 0.002$ & 0.931$ \pm 0.002$ \\ 
Sequence             & 32 (LSTM)       & 0.927$ \pm 0.002$  & 0.927$ \pm 0.001$  & 0.928$ \pm 0.002$ & 0.929$ \pm 0.002$ & 0.927$ \pm 0.002$ \\ 
\bottomrule
\end{tabular}%
}
 \caption{\textcolor{black}{The effect of varying the number of layers in the deep neural network for DNN models trained on API usage, API frequency and API sequence feature sets}}
 \label{tab:APIDL}
\end{table}

\begin{table}[t]

\centering
\resizebox{\columnwidth}{!}{%
\begin{tabular}{llllll}
\toprule
\textbf{Classifier} & \textbf{Accuracy} & \textbf{Precision} & \textbf{F1-Score} & \textbf{TPR}      & \textbf{TNR}      \\ 
\midrule

\multicolumn{6}{l}{\textit{\textcolor{black}{Features: API usage}}}\\
SVM & 0.953$ \pm 0.004$ & 0.957$ \pm 0.002$  & 0.954$ \pm 0.001$ & 0.956$ \pm 0.003$ & 0.958$ \pm 0.001$ \\ 
RF & 0.966$ \pm 0.002$ & 0.968$ \pm 0.002$  & 0.966$ \pm 0.002$ & 0.964$ \pm 0.003$ & 0.969$ \pm 0.001$ \\ 
DT & 0.945$ \pm 0.002$ & 0.950$ \pm 0.002$  & 0.954$ \pm 0.001$ & 0.947$ \pm 0.002$ & 0.950$ \pm 0.002$ \\ 
kNN & 0.942$ \pm 0.003$ & 0.942$ \pm 0.002$ & 0.944$ \pm 0.002$ & 0.941$ \pm 0.002$ & 0.944$ \pm 0.002$ \\ 
DNN & 0.961$ \pm 0.001$ & 0.962$ \pm 0.003$ & 0.963$ \pm 0.002$ & 0.959$ \pm 0.001$ & 0.962$ \pm 0.001$ \\ 
\midrule
\multicolumn{6}{l}{\textit{\textcolor{black}{Features: API frequency}}}\\
SVM                 & 0.951$ \pm 0.001$            & 0.952$ \pm 0.002$               & 0.953$ \pm 0.002$               & 0.949$ \pm 0.002$            & 0.951$ \pm 0.002$             \\ 
RF       & 0.962$ \pm 0.001$           & 0.963$ \pm 0.002$               & 0.961$ \pm 0.002$               & 0.958$ \pm 0.001$             & 0.962$ \pm 0.003$            \\ 
DT                  & 0.942$ \pm 0.001$           & 0.943$ \pm 0.002$                & 0.941$ \pm 0.002$               & 0.944$ \pm 0.001$             & 0.945$ \pm 0.001$             \\ 
kNN                 & 0.943$ \pm 0.001$            & 0.944$ \pm 0.002$               & 0.945$ \pm 0.003$      & 0.948$ \pm 0.001$            & 0.949$ \pm 0.003$            \\ 
DNN                 & \textbf{0.970}$ \pm 0.001$ & \textbf{0.971}$ \pm 0.001$ & \textbf{0.969}$ \pm 0.002$   & \textbf{0.979}$ \pm 0.002$ & \textbf{0.971}$ \pm 0.002$ \\
\bottomrule
\end{tabular}%
}\caption{\textcolor{black}{Comparison of different ML models using API usage and frequency feature sets}}
\label{tab:fullAPIFreq}
\end{table}

Ma et al.\ \cite{Ma2019} used a different model for the three datasets: a DT for the API usage dataset, a deep neural network (DNN) for the API frequency dataset, and a long short-term memory (LSTM) model for the API sequence dataset. They also built an ensemble from the three models. In Table \ref{tab:APIUsage}, we summarise our results from using this same approach. This indicates that the API frequency model performs the best, with both the usage and frequency models outperforming the API sequence models. We did not find any benefit to ensembling the models. This is in opposition to the findings of Ma et al.\ \cite{Ma2019}, for whom the best model was the ensemble, followed by the sequence model. Like Ma et al., we also looked more closely at the effect of network depth on the performance of the DNN and LSTM models. Table \ref{tab:APIDL} shows the results, showing that best performance is now achieved with a DNN of 16 layers, which is twice the depth of the best DNN found by Ma et al.

Kruskal-Wallis showed that the differences between the average accuracies in Table \ref{tab:APIUsage} are significant (H = 35.63, p $<$ 0.05). Dunn’s tests indicated a significant difference between the frequency model and the sequence model, and no significant difference between the ensemble model, usage model and the frequency model. This supports the conclusion that the simpler usage and frequency-based models are more discriminative, at least within a contemporary Android environment.

A limitation of the original study was that different models were used for different representations, meaning that the effect of changing the representation could not be isolated from the effect of changing the model. To give more insight into whether the additional information embedded in the frequency representation over the usage representation is useful, we trained each of our core models, plus DNN, on both the API usage and frequency datasets. The results are shown in \textcolor{black}{Table  \ref{tab:fullAPIFreq}}. Two things are notable: first, the differences in model accuracies are relatively small. Second, the ranking of the models changes depending on the representation, with DNNs doing best on the frequency dataset and RFs doing best on the usage dataset. Perhaps coincidentally, this justifies the model choices of Ma et al.

\subsection{\textcolor{black}{Selection of API call feature sets}}

Our results indicate that using API call features to build ML models leads to very good rates of malware detection. However, the large number of API calls available in the current Android SDK results in a high computational overhead when building models. In this section, we look at ways of reducing the size of the feature set.
 
\textbf{Jung et al.\ \cite{Jung2018}} approached this by analysing the top 50 API calls used in benign applications and malware applications, resulting in two feature sets, one from each of the applications type, which they used to train RF models. The approach was evaluated using a dataset of 30,159 benign and 30,084 malicious applications. The results of our reimplementation are shown in Table \ref{tab:JUS}. Of the two feature sets, the use of the top 50 malware API calls leads to better models, and this is supported by a Kruskal-Wallis test, which showed a significant difference (H = 14.23, p $<$ 0.05). However, the rates of accuracy achieved in our experiments are considerably less than those reported by Jung et al, and the resulting models performed relatively poorly. This difference may in part be due to the much larger size of the API call set in contemporary versions of Android.

\begin{table}[t]

        \centering
\resizebox{\textwidth}{!}{%
\begin{tabular}{lllllllllll}
\toprule
&\multicolumn{5}{c}{\textbf{Jung   et al.\ Results}}                                                                                                                                                                         & \multicolumn{5}{c}{\textbf{Our Results}}                                                                                                                                              \\  \cmidrule(lr){2-6} \cmidrule(lr){7-11}
\multicolumn{1}{l}{\textbf{Dataset}}        & \multicolumn{1}{l}{\textbf{Accuracy}} & \multicolumn{1}{l}{\textbf{Precision}} & \multicolumn{1}{l}{\textbf{F1-Score}} & \multicolumn{1}{l}{\textbf{TPR}} & \textbf{TNR} & \multicolumn{1}{l}{\textbf{Accuracy}} & \multicolumn{1}{l}{\textbf{Precision}} & \multicolumn{1}{l}{\textbf{F1-Score}} & \multicolumn{1}{l}{\textbf{TPR}}      & \textbf{TNR}      \\ \midrule
\multicolumn{1}{l}{Top 50 Benign}  & \multicolumn{1}{l}{\textbf{0.999}}             & \multicolumn{1}{l}{-}                  & \multicolumn{1}{l}{-}                 & \multicolumn{1}{l}{-}            & -            & \multicolumn{1}{l}{0.701$ \pm 0.002$} & \multicolumn{1}{l}{0.677$ \pm 0.003$}  & \multicolumn{1}{l}{0.721$ \pm 0.003$} & \multicolumn{1}{l}{0.632$ \pm 0.005$} & \textbf{0.771}$ \pm 0.004$ \\ 
\multicolumn{1}{l}{Top 50 Malware} & \multicolumn{1}{l}{0.978}             & \multicolumn{1}{l}{-}                  & \multicolumn{1}{l}{-}                 & \multicolumn{1}{l}{-}            & -            & \multicolumn{1}{l}{\textbf{0.762}$ \pm 0.002$} & \multicolumn{1}{l}{\textbf{0.691}$ \pm 0.002$}  & \multicolumn{1}{l}{\textbf{0.799}$ \pm 0.001$} & \multicolumn{1}{l}{\textbf{0.948}$ \pm 0.002$} & 0.576$ \pm 0.002$ \\ 
\bottomrule
\end{tabular}%
    }
    \caption{Result comparison between Jung et al.\ \cite{Jung2018} and our reimplementation, \textcolor{black}{showing the performance of models trained just using the top 50 API calls found in malware and the top 50 API calls found in benign software}}
    \label{tab:JUS}
\end{table}

\begin{table}[t]

\centering
\resizebox{\textwidth}{!}{%
\begin{tabular}{lllllllllll}
\toprule
&\multicolumn{5}{c}{\textbf{Muzaffar   et al.\ Results}} & \multicolumn{5}{c}{\textbf{Our Results}}                                                                                                                                              \\ 
\cmidrule(lr){2-6} \cmidrule(lr){7-11}
\multicolumn{1}{l}{\textbf{Classifier}} & \multicolumn{1}{l}{\textbf{Accuracy}} & \multicolumn{1}{l}{\textbf{Precision}} & \multicolumn{1}{l}{\textbf{F1-Score}} & \multicolumn{1}{l}{\textbf{TPR}}      & \textbf{TNR}      & \multicolumn{1}{l}{\textbf{Accuracy}} & \multicolumn{1}{l}{\textbf{Precision}} & \multicolumn{1}{l}{\textbf{F1-Score}} & \multicolumn{1}{l}{\textbf{TPR}}      & \textbf{TNR}      \\ 
\midrule
\multicolumn{1}{l}{SVM}                 & \multicolumn{1}{l}{0.955$ \pm 0.002$} & \multicolumn{1}{l}{0.957$ \pm 0.004$}  & \multicolumn{1}{l}{0.955$ \pm 0.002$} & \multicolumn{1}{l}{0.953$ \pm 0.004$} & 0.957$ \pm 0.004$ & \multicolumn{1}{l}{0.953$ \pm 0.004$} & \multicolumn{1}{l}{0.957$ \pm 0.002$}  & \multicolumn{1}{l}{0.954$ \pm 0.001$} & \multicolumn{1}{l}{0.956$ \pm 0.003$} & 0.958$ \pm 0.001$ \\ 
\multicolumn{1}{l}{RF}       & \multicolumn{1}{l}{\textbf{0.959}$ \pm 0.001$} & \multicolumn{1}{l}{\textbf{0.960}$ \pm 0.001$}  & \multicolumn{1}{l}{\textbf{0.959}$ \pm 0.002$} & \multicolumn{1}{l}{\textbf{0.957}$ \pm 0.001$} & \textbf{0.960}$ \pm 0.002$ & \multicolumn{1}{l}{\textbf{0.966}$ \pm 0.002$} & \multicolumn{1}{l}{\textbf{0.968}$ \pm 0.002$}  & \multicolumn{1}{l}{\textbf{0.966}$ \pm 0.002$} & \multicolumn{1}{l}{\textbf{0.964}$ \pm 0.003$} & \textbf{0.969}$ \pm 0.001$ \\ 
\multicolumn{1}{l}{DT}                  & \multicolumn{1}{l}{0.940$ \pm 0.000$} & \multicolumn{1}{l}{0.938$ \pm 0.003$}  & \multicolumn{1}{l}{0.941$ \pm 0.002$} & \multicolumn{1}{l}{0.943$ \pm 0.002$} & 0.938$ \pm 0.002$ & \multicolumn{1}{l}{0.945$ \pm 0.002$} & \multicolumn{1}{l}{0.950$ \pm 0.002$}  & \multicolumn{1}{l}{0.954$ \pm 0.001$} & \multicolumn{1}{l}{0.947$ \pm 0.002$} & 0.950$ \pm 0.002$ \\ 
\multicolumn{1}{l}{Naïve Bayes}         & \multicolumn{1}{l}{0.744$ \pm 0.002$} & \multicolumn{1}{l}{0.673$ \pm 0.003$}  & \multicolumn{1}{l}{0.789$ \pm 0.002$} & \multicolumn{1}{l}{0.957$ \pm 0.002$} & 0.531$ \pm 0.005$ & \multicolumn{1}{l}{0.750$ \pm 0.001$} & \multicolumn{1}{l}{0.675$ \pm 0.001$}  & \multicolumn{1}{l}{0.791$ \pm 0.002$} & \multicolumn{1}{l}{0.957$ \pm 0.002$} & 0.541$ \pm 0.003$ \\ 
\multicolumn{1}{l}{AdaBoost}            & \multicolumn{1}{l}{0.943$ \pm 0.002$} & \multicolumn{1}{l}{0.943$ \pm 0.001$}  & \multicolumn{1}{l}{0.943$ \pm 0.002$} & \multicolumn{1}{l}{0.944$ \pm 0.004$} & 0.942$ \pm 0.001$ & \multicolumn{1}{l}{0.943$ \pm 0.002$} & \multicolumn{1}{l}{0.943$ \pm 0.003$}  & \multicolumn{1}{l}{0.943$ \pm 0.002$} & \multicolumn{1}{l}{0.944$ \pm 0.004$} & 0.943$ \pm 0.001$ \\ 
\multicolumn{1}{l}{kNN}           & \multicolumn{1}{l}{-}                 & \multicolumn{1}{l}{-}                 & \multicolumn{1}{l}{-}                 & \multicolumn{1}{l}{-}                 & -                                           & \multicolumn{1}{l}{0.942$ \pm 0.003$} & \multicolumn{1}{l}{0.942$ \pm 0.002$} & \multicolumn{1}{l}{0.944$ \pm 0.002$} & \multicolumn{1}{l}{0.941$ \pm 0.002$} & 0.944$ \pm 0.002$                           \\ 
\multicolumn{1}{l}{DNN (16)}           & \multicolumn{1}{l}{-}                 & \multicolumn{1}{l}{-}                 & \multicolumn{1}{l}{-}                 & \multicolumn{1}{l}{-}                 & -                                           & \multicolumn{1}{l}{0.961$ \pm 0.001$} & \multicolumn{1}{l}{0.962$ \pm 0.003$} & \multicolumn{1}{l}{0.963$ \pm 0.002$} & \multicolumn{1}{l}{0.959$ \pm 0.001$} & 0.962$ \pm 0.001$                           \\ 

\bottomrule

\end{tabular}%
}\caption{Result comparison between Muzaffar et al.\ \cite{Muzaffar2021} and our reimplementation, \textcolor{black}{showing the relative performance of different ML models trained using the full API usage feature set}}
\label{tab:fullAPICalls}
\end{table}

\begin{table}[t]

\centering
\resizebox{\textwidth}{!}{%
\begin{tabular}{lllllllllllll}

\toprule
&\multicolumn{6}{c}{\textbf{Muzaffar   et al.\ Results}}                                                                                                                                                                                                                           & \multicolumn{6}{c}{\textbf{Our Results}}                                                                                                                                                                                       \\  \cmidrule(lr){2-7} \cmidrule(lr){8-13}
\multicolumn{1}{l}{\textbf{Selection}} & \multicolumn{1}{l}{\textbf{Features}} & \multicolumn{1}{l}{\textbf{Accuracy}} & \multicolumn{1}{l}{\textbf{Precision}} & \multicolumn{1}{l}{\textbf{F1-Score}} & \multicolumn{1}{l}{\textbf{TPR}}      & \textbf{TNR}      & \multicolumn{1}{l}{\textbf{Features}} & \multicolumn{1}{l}{\textbf{Accuracy}} & \multicolumn{1}{l}{\textbf{Precision}} & \multicolumn{1}{l}{\textbf{F1-Score}} & \multicolumn{1}{l}{\textbf{TPR}}      & \textbf{TNR}      \\ 
\midrule
\multicolumn{1}{l}{MI}         & \multicolumn{1}{l}{10,000}            & \multicolumn{1}{l}{\textbf{0.962}$ \pm 0.001$} & \multicolumn{1}{l}{0.960$ \pm 0.001$}  & \multicolumn{1}{l}{\textbf{0.961}$ \pm 0.002$} & \multicolumn{1}{l}{\textbf{0.959}$ \pm 0.002$} & 0.963$ \pm 0.002$ & \multicolumn{1}{l}{10,000}            & \multicolumn{1}{l}{0.967$ \pm 0.001$} & \multicolumn{1}{l}{0.969$ \pm 0.001$}  & \multicolumn{1}{l}{0.967$ \pm 0.002$} & \multicolumn{1}{l}{0.963$ \pm 0.002$} & 0.970$ \pm 0.002$ \\ 
\multicolumn{1}{l}{VT}         & \multicolumn{1}{l}{6,443}             & \multicolumn{1}{l}{0.961$ \pm 0.002$} & \multicolumn{1}{l}{\textbf{0.964}$ \pm 0.001$}  & \multicolumn{1}{l}{\textbf{0.961}$ \pm 0.002$} & \multicolumn{1}{l}{0.958$ \pm 0.002$} & 0.964$ \pm 0.001$ & \multicolumn{1}{l}{6,443}             & \multicolumn{1}{l}{\textbf{0.968}$ \pm 0.002$} & \multicolumn{1}{l}{\textbf{0.971}$ \pm 0.002$}  & \multicolumn{1}{l}{\textbf{0.968}$ \pm 0.002$} & \multicolumn{1}{l}{\textbf{0.965}$ \pm 0.002$} & \textbf{0.971}$ \pm 0.001$ \\ 
\multicolumn{1}{l}{PCC}                        & \multicolumn{1}{l}{15,000}            & \multicolumn{1}{l}{0.951$ \pm 0.001$} & \multicolumn{1}{l}{0.952$ \pm 0.002$}  & \multicolumn{1}{l}{0.951$ \pm 0.002$} & \multicolumn{1}{l}{0.938$ \pm 0.001$} & \textbf{0.965}$ \pm 0.002$ & \multicolumn{1}{l}{15,000}            & \multicolumn{1}{l}{0.961$ \pm 0.001$} & \multicolumn{1}{l}{0.970$ \pm 0.003$}  & \multicolumn{1}{l}{0.961$ \pm 0.002$} & \multicolumn{1}{l}{0.950$ \pm 0.001$} & \textbf{0.971}$ \pm 0.002$ \\ 
\multicolumn{1}{l}{CS}         & \multicolumn{1}{l}{-}        & \multicolumn{1}{l}{-}                 & \multicolumn{1}{l}{-}                 & \multicolumn{1}{l}{-}                 & \multicolumn{1}{l}{-}                 & -                                           & \multicolumn{1}{l}{10,000}   & \multicolumn{1}{l}{0.961$ \pm 0.003$} & \multicolumn{1}{l}{0.961$ \pm 0.004$} & \multicolumn{1}{l}{0.954$ \pm 0.002$} & \multicolumn{1}{l}{0.967$ \pm 0.001$} & 0.953$ \pm 0.001$                           \\ 
\multicolumn{1}{l}{\textcolor{black}{None}}  & \multicolumn{1}{l}{134,207}     & \multicolumn{1}{l}{0.959$ \pm 0.001$} & \multicolumn{1}{l}{0.960$ \pm 0.001$}  & \multicolumn{1}{l}{0.959$ \pm 0.002$} & \multicolumn{1}{l}{0.957$ \pm 0.001$} & 0.960$ \pm 0.002$ & \multicolumn{1}{l}{134,207}  & \multicolumn{1}{l}{0.966$ \pm 0.002$} & \multicolumn{1}{l}{0.968$ \pm 0.002$}  & \multicolumn{1}{l}{0.966$ \pm 0.002$} & \multicolumn{1}{l}{0.964$ \pm 0.003$} & 0.969$ \pm 0.001$ \\

\bottomrule
\end{tabular}%
}\caption{Result comparison between Muzaffar et al.\ \cite{Muzaffar2021} and our reimplementation, showing the relative performance of random forest models trained using API usage feature sets reduced using different feature selection algorithms. \textcolor{black}{For comparison, the corresponding results without feature selection (from Table \ref{tab:fullAPICalls}) are also shown.}}
\label{tab:fstable}
\end{table}

\begin{figure}[t]
\includegraphics[width=\columnwidth]{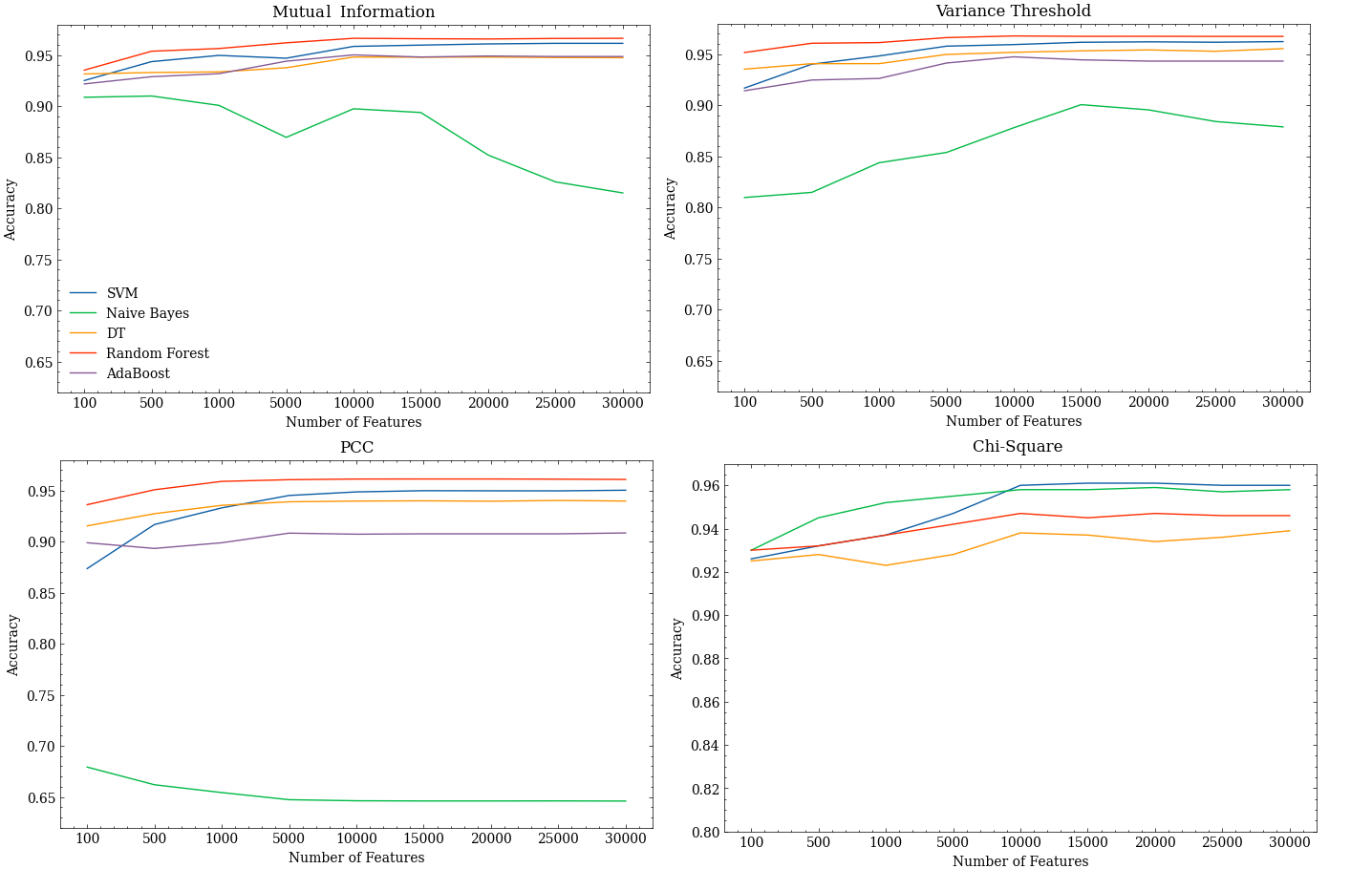}
\centering
\caption{\textcolor{black}{Relationship between number of API calls used and model accuracy for API usage features ranked using various methods}}
\label{fig:combinep}
\end{figure}

\begin{figure}[t]
\includegraphics[width=9cm]{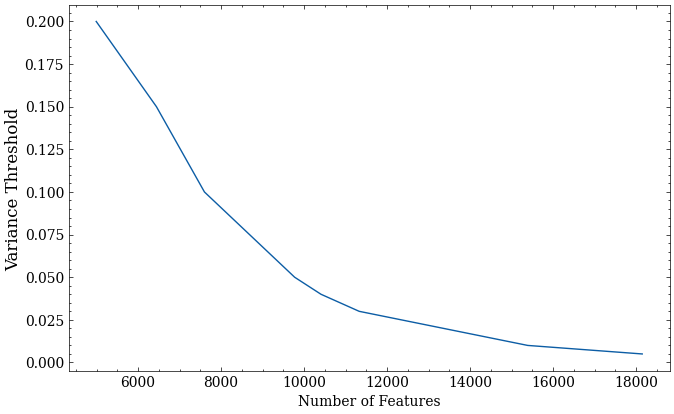}
\centering
\caption{Number of \textcolor{black}{API usage} features selected according to variance threshold}
\label{fig:vtthresholds}
\end{figure}

 
Although these results suggest that 50 API calls are insufficient to train accurate models, from a practical perspective, it remains desirable to work with smaller feature sets. Consequently, in \textbf{Muzaffar et al.\ \cite{Muzaffar2021}}, we investigated the use of dimensionality reduction in API feature sets more broadly. We used a dataset of 20,000 benign applications and 20,000 malicious applications, and our results showed that the accuracy of the ML models could be improved by reducing the number of API calls by approximately 95\% from the original set of 134,207. We trained SVM, RF, DT, AdaBoost and Naïve Bayes models, and used MI, VT and PCC for feature selection.

We reimplemented the study by using the current, considerably larger, dataset of 62,000 benign and 62,000 malicious applications. Table \ref{tab:fullAPICalls} shows the results for the full API calls dataset. We also added kNN and chi-square and compared the results. The RF models achieved the best results, followed by SVM. Naïve Bayes was included because of its training efficiency with large feature sets, but it performed poorly in terms of discrimination.
 

The feature selection algorithms were then used to reduce the feature set sizes to between 100 and 30,000 API calls. Fig. \ref{fig:combinep} shows the results for model accuracy, indicating that both the feature set size and the feature selection algorithm used have an impact on performance. In each case, it can be seen that accuracy plateaus beyond a certain number of API calls, supporting the idea that we only need to use a subset of the full Android API when training malware detection models.

Regardless of feature set size, RFs were the best models. Table \ref{tab:fstable} shows the RF performance metrics for each of the feature selection algorithms. A  Kruskal-Wallis test \textcolor{black}{across these models and the best performing RF model trained on the full API set} indicated significant differences (H = 32.59, p $<$ 0.05). Post hoc Dunn’s tests then showed that feature sets derived using variance threshold led to significantly better models. Notably, these models could outperform the full API set models across all metrics while using only 5\% (6,443) of the total API calls.



Fig. \ref{fig:vtthresholds} shows variance values for feature set sizes of 100 to 30,000. This shows that about half the features have a variance below 0.025, which also signifies that many of the API calls are unlikely to play a useful role in discrimination. Overall, these results show that the number of API calls can be significantly reduced without loss of accuracy.



\subsection{\textcolor{black}{Use of the Drebin feature set}}
 
The Drebin dataset of \textbf{Arp et al.\ \cite{Arp2014}} is widely used in Android malware detection, both due to its availability and the large number of extracted features. They provided the following features for each application in the dataset:
\begin{itemize}
\item Hardware components: The set of hardware components that an application requests.
\item Requested permissions: The set of permissions an application requests in their manifest file.
\item App components: The different components the application requests in the manifest.
\item Filtered intents: Intents used by the application; these could include intents like “BOOT\_COMPLETED”. Applications use intents for inter-process communication.
\item Restricted API calls: Some critical calls are restricted by Android, these are looked for in the Dex files.
\item Used permissions: Felt et al.\ \cite{Felt} introduced a method which was used to match API calls to permissions, hence obtaining the permissions which are actually used by the application.
\item Suspicious API calls: API calls that can access sensitive data.
\item Network addresses: These include IP addresses, URLs, and hostnames.
\end{itemize}
 
To get an indication of whether this feature set remains useful within a current Android environment, we extracted the same features from our set of contemporary applications. Table \ref{tab:DrebinUS} outlines the differences in the datasets, showing a growth from 535,000 to three million distinct features.
Table \ref{tab:DrebinUS} also shows the accuracy of SVM models trained on these two datasets. This shows that, although not as effective as the API call models we looked at earlier, the Drebin feature set does remain competitive.
 
\begin{table}[t]

    \centering
\resizebox{\columnwidth}{!}{%
    \begin{tabular}{lrrlll }
  \toprule
    \bf & \bf Benign  & \bf Malware & \bf Number of features & \bf Accuracy \\ 
   \midrule
 
        \bf Drebin &  123,453 & 5,560 & 535,000 & 0.94 \\ 
        \bf Our Dataset & 62,000  & 62,000 & 3 million reduced to 600,000 & 0.956  \\ 
 \bottomrule
 
    \end{tabular}%
    }    \caption{\textcolor{black}{Comparison of samples and Drebin feature counts in Arp et al's original dataset and our dataset, also showing the accuracy of SVM models trained on these two datasets}}
    \label{tab:DrebinUS}
 
\end{table}

\begin{table}[t]

\centering
\resizebox{\columnwidth}{!}{%
\begin{tabular}{llllllll}
\toprule
\textbf{Feature Selection} & \textbf{Classifier} & \textbf{Number of Features} & \textbf{Accuracy} & \textbf{F1-Score} & \textbf{Precision} & \textbf{TPR} & \textbf{TNR} \\ 
\midrule
Features used twice & SVM & 600,000  & 0.956$ \pm 0.002$     &  0.952$ \pm 0.003$ & 0.949$ \pm 0.002$     &  0.961$ \pm 0.002$                  & 0.948$ \pm 0.001$  \\  
MI         & SVM                 & 30000                       & \textbf{0.897}$ \pm 0.002$              & \textbf{0.895}$ \pm 0.003$               & 0.887$ \pm 0.003$                & \textbf{0.897}$ \pm 0.003$          & \textbf{0.891}$ \pm 0.001$          \\ 
VT         & SVM                 & 35000                       & 0.893$ \pm 0.002$              & 0.893$ \pm 0.002$              & 0.886$ \pm 0.004$                & 0.894$ \pm 0.002$         & 0.886$ \pm 0.002$         \\ 
CS                 & RF       & 25000                       & 0.895$ \pm 0.002$              & 0.892$ \pm 0.004$              & \textbf{0.889}$ \pm 0.003$               & 0.896$ \pm 0.002$         & 0.889$ \pm 0.002$         \\ 
PCC                        & RF       & 35000                       & 0.891$ \pm 0.002$              & 0.890$ \pm 0.002$              & 0.885$ \pm 0.002$               & 0.894$ \pm 0.002$         & 0.884$ \pm 0.003$         \\ 
\bottomrule
\end{tabular}%
}\caption{\textcolor{black}{Comparison of models trained on reduced Drebin feature sets, using different feature selection algorithms, showing results for the most effective ML model in each case}}
\label{tab:drebinfs}
\end{table}
 
However, the size of the feature set means that it is computationally very challenging to train models. To address this, we only used the features that were used by at least two applications in the dataset, thereby reducing the feature set to 600,000. This allowed us to construct a dataset of a similar dimensionality to the original one. We then retrained the SVM models on this reduced dataset, and Table \ref{tab:drebinfs} shows that this led to a small reduction in accuracy.
 

 
 
 
 
 
We then looked at whether feature selection algorithms could be used to reduce the Drebin feature set to an even smaller set of useful features. Table \ref{tab:drebinfs} shows the results of applying the core set of feature selection algorithms and ML models. Although this significantly decreases the feature set size, it also leads to a significant decrease in accuracy. To conclude, the Drebin feature set does perform well; however, it may be difficult to use it in a production environment and reducing the number of features using feature selection algorithms decreases the performance of the models significantly.

\subsection{\textcolor{black}{Use and representation of opcode features}}
 
 
 

Rather than a sequence of API calls, a program can instead be viewed, at a lower level, as a sequence of machine language instructions, or \textit{opcodes}. A relatively small number of studies have considered opcode analysis as a basis for Android malware detection, and most have involved extracting features in the form of n-grams of opcodes, referred to as \textit{n-opcodes}. For example, lets say that an application $A$ is the sequence $op_1$, $op_2$, $op_3$, $op_4$. A 2-gram, also known as bigram, representation of $A$ would be three bigrams, namely [$op_1$, $op_2$], [$op_2$, $op_3$] and [$op_3$, $op_4$]. To measure the benefits of this approach, and compare it against more traditional static features like permissions and API calls, we reimplement a study by \textbf{Kang et al.\ \cite{Kang2016}}, which considered both usage and frequency representations (see Section \ref{representations}) of n-opcodes.

Kang et al.\ used a dataset comprising 1,260 benign and 1,260 malware applications. They extracted n-opcodes of sizes one up to 10, and then used MI to reduce the number of features, before training NB, RF, SVM and partial decision tree (PART) models. In our reimplementation, we replaced PART with our standard DT model and also trained kNNs. We used our four core feature selectors. Table \ref{tab:nop} shows the number of n-opcodes for different values of $n$ up to 10 both for our dataset and the dataset used by Kang et al.\ This indicates a significant increase in the number of unique n-opcodes, particularly for larger values of $n$, making feature selection a necessary step. Table \ref{tab:infonop} shows the number of features for each value of $n$ after carrying out feature selection using MI, which we found to be the most effective feature selector in our experiments.
 
\begin{table}[t]

\centering
 
\begin{tabular}{lrr}
\toprule
\textbf{n} & \textbf{Kang   et al.}          & \textbf{Our Dataset}          \\ 
\midrule
\multicolumn{1}{l}{1}          & 214                & 218                \\ 
\multicolumn{1}{l}{2}          & 22,371             & 44,411              \\ 
\multicolumn{1}{l}{3}          & 399,598            & 1,049,741            \\ 
\multicolumn{1}{l}{4}          & 2,201,377          & 8,474,254                   \\ 
\multicolumn{1}{l}{5}          & 6,458,246          & 35,033,022                   \\ 
\multicolumn{1}{l}{6}          & 12,969,857         & 95,082,086               \\ 
\multicolumn{1}{l}{7}          & 20,404,473         & 195,415,174                    \\ 
\multicolumn{1}{l}{8}          & 27,366,890         & 330,418,164                  \\ 
\multicolumn{1}{l}{9}          & 33,024,116         &   484,711,644                 \\ 
\bottomrule
\end{tabular}
\caption{Number of unique n-opcodes for different values of $n$}
\label{tab:nop}
\end{table}

\begin{table}[t]

\centering
\begin{tabular}{lrr}
\toprule
\textbf{n} & \textbf{Usage} & \textbf{Frequency} \\ 
\midrule
1          & 99              & 159                \\ 
2          & 3,991           & 5,387              \\ 
3          & 14,321          & 20,219             \\ 
4          & 38,456          & 45,201             \\ 
5          & 46,781          & 49,246             \\ 
6          & 53,578          & 56,983             \\ 
7          & 60,210          & 64,392             \\ 
8          & 58,219          & 62,239             \\ 
9          & 57,985          & 59,238             \\ 
10         & 58,456          & 59,249             \\ 
\bottomrule
\end{tabular}
\caption{Number of selected n-opcodes using mutual information}
\label{tab:infonop}
\end{table}
 
 
\begin{figure}[t]
\includegraphics[width=\columnwidth]{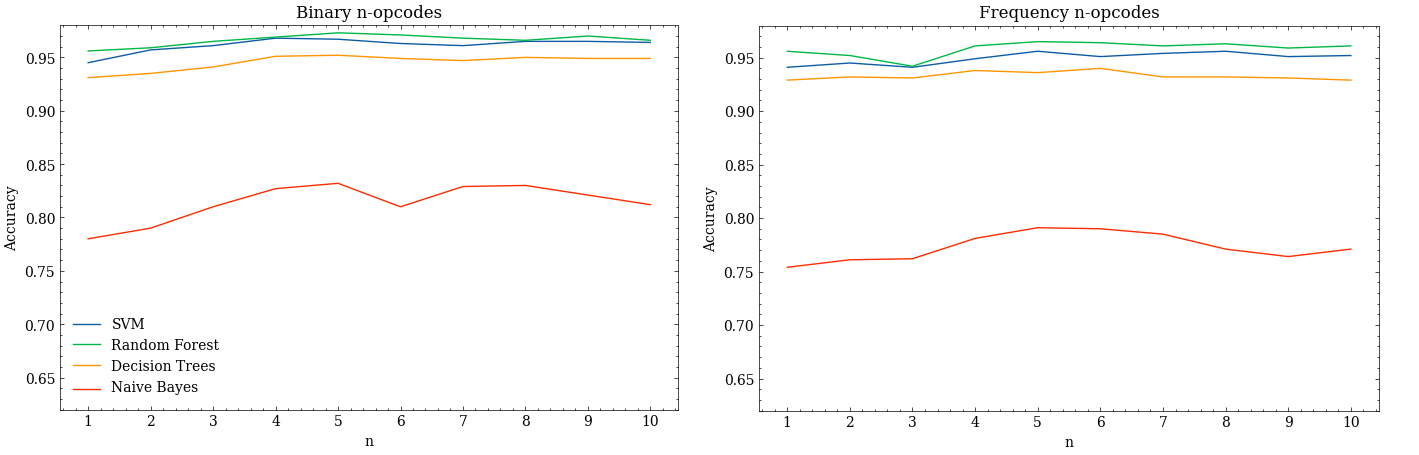}
\centering
\caption{Accuracy of ML models trained on usage and frequency-based $n$-opcode features selected using mutual information, \textcolor{black}{showing the effect of changing the value of $n$}}
\label{fig:nop}
\end{figure}

Fig. \ref{fig:nop} shows the accuracy of the models trained on the features selected by MI, and also shows the effect of varying $n$. In general, there appears to be little benefit to increasing $n$ much beyond 4, especially given the lower computational effort associated with smaller n-opcodes. Among the four ML models, it can be seen that RF models performed best for all values of $n$ and both feature representations. The average accuracies for models that use usage features were slightly higher than those which used frequency features, but the difference was small in comparison to the effect of ML model choice.

Table \ref{tab:evalnop} summarises the performance of the best models. \textcolor{black}{The best value of $n$ across both representations and across most of the ML models is $5$. This is a more consistent observation than in Kang et al's original study, where the best value of $n$ varied between 3 and 8, with a preference towards 3 and 4. Given that the F1 scores from our reimplementation are broadly similar to those originally published, this may indicate a change in the underlying pattern of opcode usage within Android malware. However, the original dataset was relatively small, so there may also have been a degree of sample bias.}

A Kruskal-Wallis test applied to our results indicates a significant difference (H = 36.44, p $<$ 0.05) between models, with Dunn's tests showing that RF and SVM models are significantly better than the others. Notably, these models have similar performance to models based on API calls. Opcodes take less time to extract compared to API calls and are much easier to train computationally on the base feature sets. However, it is much easier to interpret the API call results as APIs correspond to specific features in the Android SDK.

\begin{table}[t]
\centering
\resizebox{\textwidth}{!}{%
\begin{tabular}{lllllllllllll}
\toprule
& \multicolumn{6}{c}{\textbf{Kang et al.}} & \multicolumn{6}{c}{\textbf{Our Results}} \\ 
\cmidrule(lr){2-7} \cmidrule(lr){8-13}
\textbf{Classifier} & \textbf{n} & \textbf{Accuracy} & \textbf{Precision} & \textbf{F1-Score} & \textbf{TPR} & \textbf{TNR} & \textbf{n} & \textbf{Accuracy} & \textbf{Precision} & \textbf{F1-Score} & \textbf{TPR} & \textbf{TNR} \\ 
\midrule
\multicolumn{13}{l}{\textit{Features: Opcode usage}} \\ 
SVM & 3 & - & - & 0.98 & - & - & 4 & 0.968 $\pm$ 0.002 & 0.971 $\pm$ 0.001 & 0.971 $\pm$ 0.002 & 0.961 $\pm$ 0.001 & 0.971 $\pm$ 0.002 \\ 
RF & 4 & - & - & 0.98 & - & - & 5 & \textbf{0.973} $\pm$ 0.002 & \textbf{0.979} $\pm$ 0.002 & \textbf{0.975} $\pm$ 0.001 & \textbf{0.965} $\pm$ 0.001 & \textbf{0.979} $\pm$ 0.002 \\ 
DT & 5 & - & - & 0.98 & - & - & 5 & 0.952 $\pm$ 0.001 & 0.952 $\pm$ 0.001 & 0.951 $\pm$ 0.003 & 0.949 $\pm$ 0.002 & 0.952 $\pm$ 0.001 \\ 
Naïve Bayes & 3 & - & - & 0.84 & - & - & 5 & 0.832 $\pm$ 0.002 & 0.834 $\pm$ 0.002 & 0.828 $\pm$ 0.001 & 0.821 $\pm$ 0.003 & 0.838 $\pm$ 0.002 \\ 
\midrule
\multicolumn{13}{l}{\textit{Features: Opcode frequency}} \\ 
SVM & 4 & - & - & 0.96 & - & - & 5 & 0.956 $\pm$ 0.001 & 0.954 $\pm$ 0.002 & 0.956 $\pm$ 0.002 & 0.958 $\pm$ 0.003 & 0.954 $\pm$ 0.003 \\ 
RF & 8 & - & - & 0.97 & - & - & 5 & \textbf{0.965} $\pm$ 0.002 & \textbf{0.969} $\pm$ 0.003 & \textbf{0.967} $\pm$ 0.002 & \textbf{0.961} $\pm$ 0.002 & \textbf{0.970} $\pm$ 0.002 \\ 
DT & 3 & - & - & 0.97 & - & - & 6 & 0.940 $\pm$ 0.001 & 0.942 $\pm$ 0.001 & 0.937 $\pm$ 0.002 & 0.935 $\pm$ 0.003 & 0.941 $\pm$ 0.001 \\ 
Naïve Bayes & 4 & - & - & 0.85 & - & - & 5 & 0.791 $\pm$ 0.002 & 0.794 $\pm$ 0.001 & 0.784 $\pm$ 0.003 & 0.781 $\pm$ 0.002 & 0.798 $\pm$ 0.001 \\ 
\bottomrule
\end{tabular}%
}
\caption{Result comparison between Kang et al.\ \cite{Kang2016} and our reimplementation, \textcolor{black}{showing the effect of ML model choice when using usage and frequency-based $n$-opcode features, and also showing the \textcolor{black}{best experimentally-derived} value of $n$ \textcolor{black}{for each model}}. F1-scores for the original study are approximate values as they were read from the authors' plots.}
\label{tab:evalnop}
\end{table}
 
 
A common approach within deep learning is to convert numerical data into pseudoimages and use these to train a convolutional neural network (CNN). An example of this approach within the Android malware detection literature can be found in \textbf{Xiao and Yang \cite{Xiao2019a}}, who converted disassembled opcode files into RGB images. To determine whether this has any advantage over n-opcode encodings, we reimplemented the same approach and wrote an algorithm that converts opcode sequence files into RGB images. We then used these to train a CNN model, using the same CNN topology as Xiao and Yang, which comprises standard convolutional, pooling and fully-connected layers.
Xiao and Yang trained and evaluated their model using a dataset of 4,406 benign and 6,134 malicious applications. Table \ref{tab:CNNOP} compares their results against those of our reimplementation, which reported slightly lower metric scores. \textcolor{black}{This deviation may be due to the restricted focus of the original study, which only considered three malware families due to data sparsity.}

\textcolor{black}{Compared to Table \ref{tab:evalnop}, the results of our reimplementation are lower across the board than when using Kang et al's n-opcode representation.} This shows that there is generally no benefit to using this more complex modelling approach over simpler models. \textcolor{black}{Rather, it appears that either the direct sequential encoding of opcodes, the mapping to a pseudoimage, or the particular nature of the CNN model, makes the underlying discriminative pattern more difficult to detect.} This observation is supported by a statistical analysis, with Kruskal-Wallis (H = 56.44, p $<$ 0.05) and Dunn's test showing Xiao and Yang's approach to be significantly worse on our dataset than the best models in Table \ref{tab:evalnop}.

\begin{table}[t]

\centering
\resizebox{\textwidth}{!}{%
\begin{tabular}{llllllllll}
\toprule
\multicolumn{5}{c}{\textbf{Xiao and Yang’s   Results}}                                                                            & \multicolumn{5}{c}{\textbf{Our Results}}                                                                                                                                              \\ \cmidrule(lr){1-5} \cmidrule(lr){6-10}
\multicolumn{1}{l}{\textbf{Accuracy}} & \multicolumn{1}{l}{\textbf{Precision}} & \multicolumn{1}{l}{\textbf{F1-Score}} & \multicolumn{1}{l}{\textbf{TPR}} & \textbf{TNR} & \multicolumn{1}{l}{\textbf{Accuracy}} & \multicolumn{1}{l}{\textbf{Precision}} & \multicolumn{1}{l}{\textbf{F1-Score}} & \multicolumn{1}{l}{\textbf{TPR}}      & \textbf{TNR}      \\ 
\midrule
\multicolumn{1}{l}{0.93}              & \multicolumn{1}{l}{0.936/0.921}              & \multicolumn{1}{l}{0.94}              & \multicolumn{1}{l}{0.944/0.910}        & 0.903        & \multicolumn{1}{l}{0.913$ \pm 0.003$} & \multicolumn{1}{l}{0.912$ \pm 0.001$}  & \multicolumn{1}{l}{0.916$ \pm 0.003$} & \multicolumn{1}{l}{0.921$ \pm 0.001$} & 0.911$ \pm 0.001$ \\ 
\bottomrule
\end{tabular}%
}\caption{Result comparison between Xiao and Yang's  \cite{Xiao2019a} \textcolor{black}{CNN model trained on image-based opcode features} and our reimplementation. \textcolor{black}{Note that the authors  of the original study reported separate precision and TPR figures when predicting malware or benign software; we have reported both for completeness.}}
\label{tab:CNNOP}
\end{table}
 

\textcolor{black}{Another example of using deep learning for opcode analysis can be found in a study by \textbf{Yeboah and Baz Musah \cite{yeboah2022}}. In this case, they used a 1D CNN; that is, a one-dimensional CNN which can be applied to sequential numeric data. A notable element of their approach was the use of a word2vec embedding model to represent each opcode as a single numeric value, allowing opcode sequences to be represented as dense real-valued vectors (as opposed to using one-hot encodings). The embedding model was trained separately on unigram and bigram opcode sequences, resulting in matrices of size M×N, where M is the number of unique opcodes and N is the vector dimension. N was set to 64 for both unigram and bigram embeddings. The 1D CNN comprised convolutional layers, pooling layers, and fully connected layers. Table \ref{tab:1d} shows the results of our reimplementation, and again shows that there is no benefit to using a more complex modelling approach over the simpler models reported in Table \ref{tab:evalnop}. Our results were significantly lower than those reported by Yeboah and Baz Musah, who used a small (4948 malware, 2477 benign) though relatively recent dataset.}

\begin{table}[t]

\centering
\resizebox{\textwidth}{!}{%
\begin{tabular}{llllllllll}
\toprule
\multicolumn{5}{c}{{\color[HTML]{000000} \textbf{Yeboah and Baz Musah's Results}}}                                                                                                                                                                                                               & \multicolumn{5}{c}{{\color[HTML]{000000} \textbf{Our Results}}}                                                                                                                                                                                                                                          \\ \cmidrule(lr){1-5} \cmidrule(lr){6-10}
\multicolumn{1}{l}{{\color[HTML]{000000} \textbf{Accuracy}}} & \multicolumn{1}{l}{{\color[HTML]{000000} \textbf{Precision}}} & \multicolumn{1}{l}{{\color[HTML]{000000} \textbf{F1-Score}}} & \multicolumn{1}{l}{{\color[HTML]{000000} \textbf{TPR}}} & {\color[HTML]{000000} \textbf{TNR}} & \multicolumn{1}{l}{{\color[HTML]{000000} \textbf{Accuracy}}} & \multicolumn{1}{l}{{\color[HTML]{000000} \textbf{Precision}}} & \multicolumn{1}{l}{{\color[HTML]{000000} \textbf{F1-Score}}} & \multicolumn{1}{l}{{\color[HTML]{000000} \textbf{TPR}}}      & {\color[HTML]{000000} \textbf{TNR}}      \\ 
\midrule
\multicolumn{1}{l}{{\color[HTML]{000000} -}}                 & \multicolumn{1}{l}{{\color[HTML]{000000} 0.98}}               & \multicolumn{1}{l}{{\color[HTML]{000000} 0.97}}              & \multicolumn{1}{l}{{\color[HTML]{000000} 0.97}}         & {\color[HTML]{000000} -}            & \multicolumn{1}{l}{{\color[HTML]{000000} 0.936$ \pm 0.001$}} & \multicolumn{1}{l}{{\color[HTML]{000000} 0.933$ \pm 0.002$}}  & \multicolumn{1}{l}{{\color[HTML]{000000} 0.935$ \pm 0.002$}} & \multicolumn{1}{l}{{\color[HTML]{000000} 0.923$ \pm 0.002$}} & {\color[HTML]{000000} 0.929$ \pm 0.001$} \\ 
\bottomrule
\end{tabular}%
}\caption{\textcolor{black}{Result comparison between Yeboah and Baz Musah's \cite{yeboah2022} 1D CNN model trained on sequence-based opcode features and our reimplementation}}
\label{tab:1d}
\end{table}

\subsection{Comparison of static modelling approaches}

To identify the best combination of features, ML models and feature selection algorithms for static analysis, we carried out a Kruskal-Wallis test followed by a pairwise comparison between the  modelling approaches from this section. The results indicate that five of the models have statistically better accuracies than the others. Three of these are API call models: SVMs trained using 10,000 features selected using chi-square, RF models trained using 10,000 features selected using MI, and RF models trained using 6,443 features selected using variance threshold (Table \ref{tab:fstable}). The others are the RF usage 5-opcode model, trained using 46,781 features selected using mutual information (Table \ref{tab:evalnop}) and DNN trained using API frequency (Table \ref{tab:APIUsage}). Neither of these five modelling approaches appears to be significantly better than the others. \textcolor{black}{Within the scope of our evaluation, }this suggests that the best classes of features to use for static analysis are opcodes and API calls, and the best ML model to use is RF. However, the best choice of feature selection algorithm depends on the feature type and model type.

\section{\textcolor{black}{Reimplementation and extension of dynamic modelling approaches}} \label{sec:dynamic}

In this section, we focus on the use of dynamic features in building Android malware detection models. We reimplement five previous studies, and also report the results of new experiments that are intended to fill knowledge gaps found in these studies. 

Dynamic analysis is not as common as static analysis. This is due to several factors, including the complex set up, relatively high computational costs, and the longer time required to run the analysis.
A particular issue when running dynamic analysis is that each application needs to be executed on an emulator in order to record its dynamic features. During this process, some applications terminate unexpectedly. In some cases, the emulator itself crashes. Taking this into account, the set of applications for which we were able to complete dynamic analysis comprises 53,960 benign and 53,202 malware. However, this remains the largest and most up-to-date dataset used for dynamic analysis, to the best of our knowledge.

\subsection{\textcolor{black}{Use of system call features}}
 
System calls are the most commonly used dynamic features in the literature. Android applications use system calls to communicate with the kernel of the OS, and the \textit{strace} module can be used to trace system calls that are made by an application during its runtime.
 
\textbf{Ananya et al.\ \cite{Ananya2020}} used sequences of system calls represented as unigrams, bigrams, and trigrams as the features of their models. They also used feature selection to select relevant features, and trained four ML models: linear regression, decision tree, RF and XGBoost. They used a dataset of 2,475 benign applications and 2,474 malware.    
The authors proposed two novel approaches to feature selection. The first, called SAILS, builds on the conventional feature selection algorithms mutual information, chi-square and DFS. These are used to score the benign and malware features. SAILS then sorts the resulting scores in descending order, to create two sorted lists of benign and malware features. It then creates a final list that is the union of these two lists. 
The second feature selection algorithm proposed was WFS, which assigns  weights to system calls. This is done by calculating the ratio of the number of occurrences of the system call in malware to the number of occurrences in the sum of benign and malware. In all the experiments, the authors used SAILS, WFS, mutual information, chi-square and DFS to select relevant features. In our reimplementation, we also add PCC. 
 
\begin{table}[t]

\centering
\resizebox{\textwidth}{!}{%
\begin{tabular}{lllllllllllll}
\toprule
&\multicolumn{6}{c}{\textbf{Ananya et al.\   Results}}                                                                                                                                                                                            & \multicolumn{6}{c}{\textbf{Our Results}}                                                                                                                                                                                                \\                                                                                                                                                                                                                                                                                                                                                                                                                           \cmidrule(lr){2-7} \cmidrule(lr){8-13}
 
\multicolumn{1}{l}{\textbf{Classifier}}        & \multicolumn{1}{l}{\textbf{Selector}}  & \multicolumn{1}{l}{\textbf{Accuracy}} & \multicolumn{1}{l}{\textbf{Precision}} & \multicolumn{1}{l}{\textbf{F1-Score}} & \multicolumn{1}{l}{\textbf{TPR}} & \multicolumn{1}{l}{\textbf{TNR}} & \multicolumn{1}{l}{\textbf{Selector}}  & \multicolumn{1}{l}{\textbf{Accuracy}}          & \multicolumn{1}{l}{\textbf{Precision}}         & \multicolumn{1}{l}{\textbf{F1-Score}}          & \multicolumn{1}{l}{\textbf{TPR}}               & \textbf{TNR}               \\ 
\midrule
\multicolumn{13}{l}{\textit{Features: Unigrams}}      \\
\multicolumn{1}{l}{SVM} & \multicolumn{1}{l}{-} & \multicolumn{1}{l}{\textbf{-}}    & \multicolumn{1}{l}{-}         & \multicolumn{1}{l}{-}        & \multicolumn{1}{l}{-}   & \multicolumn{1}{l}{-}   & \multicolumn{1}{l}{MI} & \multicolumn{1}{l}{0.928$ \pm 0.001$} & \multicolumn{1}{l}{0.926$ \pm 0.001$} & \multicolumn{1}{l}{0.924$ \pm 0.003$} & \multicolumn{1}{l}{0.922$ \pm 0.002$} & 0.929$ \pm 0.001$ \\ 
\multicolumn{1}{l}{Linear Regression} & \multicolumn{1}{l}{MI} & \multicolumn{1}{l}{\textbf{0.977}}    & \multicolumn{1}{l}{-}         & \multicolumn{1}{l}{-}        & \multicolumn{1}{l}{-}   & \multicolumn{1}{l}{-}   & \multicolumn{1}{l}{MI} & \multicolumn{1}{l}{0.921$ \pm 0.002$} & \multicolumn{1}{l}{0.922$ \pm 0.001$} & \multicolumn{1}{l}{0.919$ \pm 0.002$} & \multicolumn{1}{l}{0.915$ \pm 0.001$} & 0.923$ \pm 0.002$ \\ 
\multicolumn{1}{l}{RF}     & \multicolumn{1}{l}{WFS}                & \multicolumn{1}{l}{0.972}    & \multicolumn{1}{l}{-}         & \multicolumn{1}{l}{-}        & \multicolumn{1}{l}{-}   & \multicolumn{1}{l}{-}   & \multicolumn{1}{l}{MI} & \multicolumn{1}{l}{\textbf{0.934}$ \pm 0.001$} & \multicolumn{1}{l}{\textbf{0.935}$ \pm 0.001$} & \multicolumn{1}{l}{\textbf{0.933}$ \pm 0.003$} & \multicolumn{1}{l}{\textbf{0.931}$ \pm 0.003$} & \textbf{0.936}$ \pm 0.001$ \\ 
\multicolumn{1}{l}{DT}     & \multicolumn{1}{l}{MI} & \multicolumn{1}{l}{0.965}    & \multicolumn{1}{l}{-}         & \multicolumn{1}{l}{-}        & \multicolumn{1}{l}{-}   & \multicolumn{1}{l}{-}   & \multicolumn{1}{l}{MI} & \multicolumn{1}{l}{0.925$ \pm 0.002$} & \multicolumn{1}{l}{0.926$ \pm 0.002$} & \multicolumn{1}{l}{0.921$ \pm 0.002$} & \multicolumn{1}{l}{0.922$ \pm 0.001$} & 0.926$ \pm 0.001$ \\ 
\multicolumn{1}{l}{XGBoost}           & \multicolumn{1}{l}{WFS}                & \multicolumn{1}{l}{0.966}    & \multicolumn{1}{l}{-}         & \multicolumn{1}{l}{-}        & \multicolumn{1}{l}{-}   & \multicolumn{1}{l}{-}   & \multicolumn{1}{l}{MI} & \multicolumn{1}{l}{0.933$ \pm 0.001$} & \multicolumn{1}{l}{0.931$ \pm 0.001$} & \multicolumn{1}{l}{0.930$ \pm 0.002$} & \multicolumn{1}{l}{\textbf{0.931}$ \pm 0.002$} & 0.932$ \pm 0.001$ \\ 
\multicolumn{1}{l}{kNN} & \multicolumn{1}{l}{-} & \multicolumn{1}{l}{\textbf{-}}    & \multicolumn{1}{l}{-}         & \multicolumn{1}{l}{-}        & \multicolumn{1}{l}{-}   & \multicolumn{1}{l}{-}   & \multicolumn{1}{l}{MI} & \multicolumn{1}{l}{0.922$ \pm 0.002$} & \multicolumn{1}{l}{0.918$ \pm 0.002$} & \multicolumn{1}{l}{0.921$ \pm 0.002$} & \multicolumn{1}{l}{0.917$ \pm 0.003$} & 0.923$ \pm 0.002$ \\ 
\midrule
\multicolumn{13}{l}{\textit{Features: Bigrams}} \\  
\multicolumn{1}{l}{SVM} & \multicolumn{1}{l}{-} & \multicolumn{1}{l}{\textbf{-}}    & \multicolumn{1}{l}{-}         & \multicolumn{1}{l}{-}        & \multicolumn{1}{l}{-}   & \multicolumn{1}{l}{-}   & \multicolumn{1}{l}{MI} & \multicolumn{1}{l}{0.941$ \pm 0.002$} & \multicolumn{1}{l}{0.939$ \pm 0.002$} & \multicolumn{1}{l}{0.942$ \pm 0.003$} & \multicolumn{1}{l}{0.938$ \pm 0.002$} & 0.943$ \pm 0.001$ \\ 
\multicolumn{1}{l}{Linear Regression} & \multicolumn{1}{l}{CS}         & \multicolumn{1}{l}{0.991}    & \multicolumn{1}{l}{-}         & \multicolumn{1}{l}{-}        & \multicolumn{1}{l}{-}   & \multicolumn{1}{l}{-}   & \multicolumn{1}{l}{MI} & \multicolumn{1}{l}{0.936$ \pm 0.001$} & \multicolumn{1}{l}{0.930$ \pm 0.001$} & \multicolumn{1}{l}{0.934$ \pm 0.002$} & \multicolumn{1}{l}{0.938$ \pm 0.002$} & 0.931$ \pm 0.001$ \\ 
\multicolumn{1}{l}{RF}     & \multicolumn{1}{l}{CS}         & \multicolumn{1}{l}{0.988}    & \multicolumn{1}{l}{-}         & \multicolumn{1}{l}{-}        & \multicolumn{1}{l}{-}   & \multicolumn{1}{l}{-}   & \multicolumn{1}{l}{CS}         & \multicolumn{1}{l}{\textbf{0.945}$ \pm 0.002$} & \multicolumn{1}{l}{0.941$ \pm 0.001$} & \multicolumn{1}{l}{\textbf{0.943}$ \pm 0.001$} & \multicolumn{1}{l}{\textbf{0.945}$ \pm 0.002$} & 0.941$ \pm 0.002$ \\ 
\multicolumn{1}{l}{DT}     & \multicolumn{1}{l}{CS}         & \multicolumn{1}{l}{0.977}    & \multicolumn{1}{l}{-}         & \multicolumn{1}{l}{-}        & \multicolumn{1}{l}{-}   & \multicolumn{1}{l}{-}   & \multicolumn{1}{l}{MI} & \multicolumn{1}{l}{0.931$ \pm 0.003$} & \multicolumn{1}{l}{0.929$ \pm 0.001$} & \multicolumn{1}{l}{0.929$ \pm 0.001$} & \multicolumn{1}{l}{0.931$ \pm 0.002$} & 0.929$ \pm 0.003$ \\ 
\multicolumn{1}{l}{XGBoost}           & \multicolumn{1}{l}{DFS}                & \multicolumn{1}{l}{\textbf{0.994}}    & \multicolumn{1}{l}{-}         & \multicolumn{1}{l}{-}        & \multicolumn{1}{l}{-}   & \multicolumn{1}{l}{-}   & \multicolumn{1}{l}{CS}         & \multicolumn{1}{l}{0.943$ \pm 0.002$} & \multicolumn{1}{l}{\textbf{0.943}$ \pm 0.001$} & \multicolumn{1}{l}{0.942$ \pm 0.003$} & \multicolumn{1}{l}{0.939$ \pm 0.002$} & \textbf{0.944}$ \pm 0.002$ \\ 
\multicolumn{1}{l}{kNN} & \multicolumn{1}{l}{-} & \multicolumn{1}{l}{\textbf{-}}    & \multicolumn{1}{l}{-}         & \multicolumn{1}{l}{-}        & \multicolumn{1}{l}{-}   & \multicolumn{1}{l}{-}   & \multicolumn{1}{l}{CS} & \multicolumn{1}{l}{0.932$ \pm 0.001$} & \multicolumn{1}{l}{0.929$ \pm 0.003$} & \multicolumn{1}{l}{0.931$ \pm 0.001$} & \multicolumn{1}{l}{0.926$ \pm 0.002$} & 0.931$ \pm 0.002$ \\ 
\midrule
\multicolumn{13}{l}{\textit{Features: Trigrams}} \\
\multicolumn{1}{l}{SVM} & \multicolumn{1}{l}{-} & \multicolumn{1}{l}{\textbf{-}}    & \multicolumn{1}{l}{-}         & \multicolumn{1}{l}{-}        & \multicolumn{1}{l}{-}   & \multicolumn{1}{l}{-}   & \multicolumn{1}{l}{CS} & \multicolumn{1}{l}{0.947$ \pm 0.001$} & \multicolumn{1}{l}{0.945$ \pm 0.002$} & \multicolumn{1}{l}{0.946$ \pm 0.002$} & \multicolumn{1}{l}{0.947$ \pm 0.002$} & 0.951$ \pm 0.003$ \\ 
\multicolumn{1}{l}{Linear Regression} & \multicolumn{1}{l}{MI} & \multicolumn{1}{l}{\textbf{0.994}}    & \multicolumn{1}{l}{-}         & \multicolumn{1}{l}{-}        & \multicolumn{1}{l}{-}   & \multicolumn{1}{l}{-}   & \multicolumn{1}{l}{MI} & \multicolumn{1}{l}{0.932$ \pm 0.003$} & \multicolumn{1}{l}{0.932$ \pm 0.001$} & \multicolumn{1}{l}{0.934$ \pm 0.003$} & \multicolumn{1}{l}{0.937$ \pm 0.002$} & 0.932$ \pm 0.002$ \\ 
\multicolumn{1}{l}{RF}     & \multicolumn{1}{l}{MI} & \multicolumn{1}{l}{0.982}    & \multicolumn{1}{l}{-}         & \multicolumn{1}{l}{-}        & \multicolumn{1}{l}{-}   & \multicolumn{1}{l}{-}   & \multicolumn{1}{l}{CS}         & \multicolumn{1}{l}{\textbf{0.951}$ \pm 0.002$} & \multicolumn{1}{l}{\textbf{0.951}$ \pm 0.001$} & \multicolumn{1}{l}{\textbf{0.950}$ \pm 0.002$} & \multicolumn{1}{l}{\textbf{0.949}$ \pm 0.003$} & \textbf{0.952}$ \pm 0.001$ \\ 
\multicolumn{1}{l}{DT}     & \multicolumn{1}{l}{WFS}                & \multicolumn{1}{l}{0.987}    & \multicolumn{1}{l}{-}         & \multicolumn{1}{l}{-}        & \multicolumn{1}{l}{-}   & \multicolumn{1}{l}{-}   & \multicolumn{1}{l}{MI} & \multicolumn{1}{l}{0.945$ \pm 0.003$} & \multicolumn{1}{l}{0.943$ \pm 0.001$} & \multicolumn{1}{l}{0.942$ \pm 0.002$} & \multicolumn{1}{l}{0.943$ \pm 0.002$} & 0.943$ \pm 0.001$ \\ 
\multicolumn{1}{l}{XGBoost}           & \multicolumn{1}{l}{MI} & \multicolumn{1}{l}{0.992}    & \multicolumn{1}{l}{-}         & \multicolumn{1}{l}{-}        & \multicolumn{1}{l}{-}   & \multicolumn{1}{l}{-}   & \multicolumn{1}{l}{CS}         & \multicolumn{1}{l}{0.950$ \pm 0.002$} & \multicolumn{1}{l}{0.948$ \pm 0.001$} & \multicolumn{1}{l}{0.947$ \pm 0.002$} & \multicolumn{1}{l}{0.947$ \pm 0.002$} & 0.948$ \pm 0.002$ \\ 
\multicolumn{1}{l}{kNN} & \multicolumn{1}{l}{-} & \multicolumn{1}{l}{\textbf{-}}    & \multicolumn{1}{l}{-}         & \multicolumn{1}{l}{-}        & \multicolumn{1}{l}{-}   & \multicolumn{1}{l}{-}   & \multicolumn{1}{l}{CS} & \multicolumn{1}{l}{0.942$ \pm 0.002$} & \multicolumn{1}{l}{0.941$ \pm 0.001$} & \multicolumn{1}{l}{0.943$ \pm 0.002$} & \multicolumn{1}{l}{0.938$ \pm 0.001$} & 0.941$ \pm 0.002$ \\ 
\bottomrule
\end{tabular}%
}\caption{Result comparison between Ananya et al.\ \cite{Ananya2020} and our reimplementation, \textcolor{black}{showing the effect of ML model choice when trained using system call features represented as unigrams, bigrams and trigrams. The most effective feature selection method is shown in each case.}}
\label{tab:sys1}
\end{table}
 
Table \ref{tab:sys1} shows the best results for each ML model using unigrams, bigrams, and trigrams, and also indicates the most effective feature selection approach in each case. Notably, the performance metrics achieved in our reimplementation are substantially lower than those  published in the original study. Ananya et al.\ reported accuracies of above 99\%, whereas our best models achieved an accuracy of around 95\%. This shows the importance of reimplementing studies using larger contemporary datasets, since it leads to the conclusion that dynamic analysis using system calls is \textit{less} effective than the less expensive static analysis approaches considered earlier.
 
\begin{figure}
     \centering
     \begin{subfigure}[b]{\columnwidth}
         \centering
         \includegraphics[width=\columnwidth]{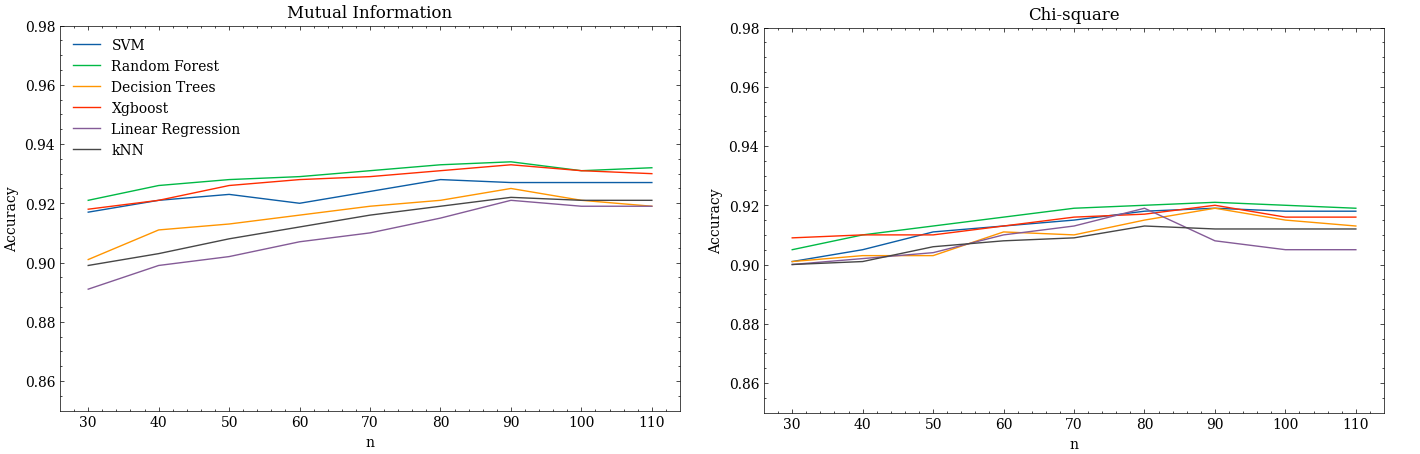}
         \caption{unigram}
         \label{fig:y equals x}
     \end{subfigure}
     \hfill
     \begin{subfigure}[b]{\columnwidth}
         \centering
         \includegraphics[width=\columnwidth]{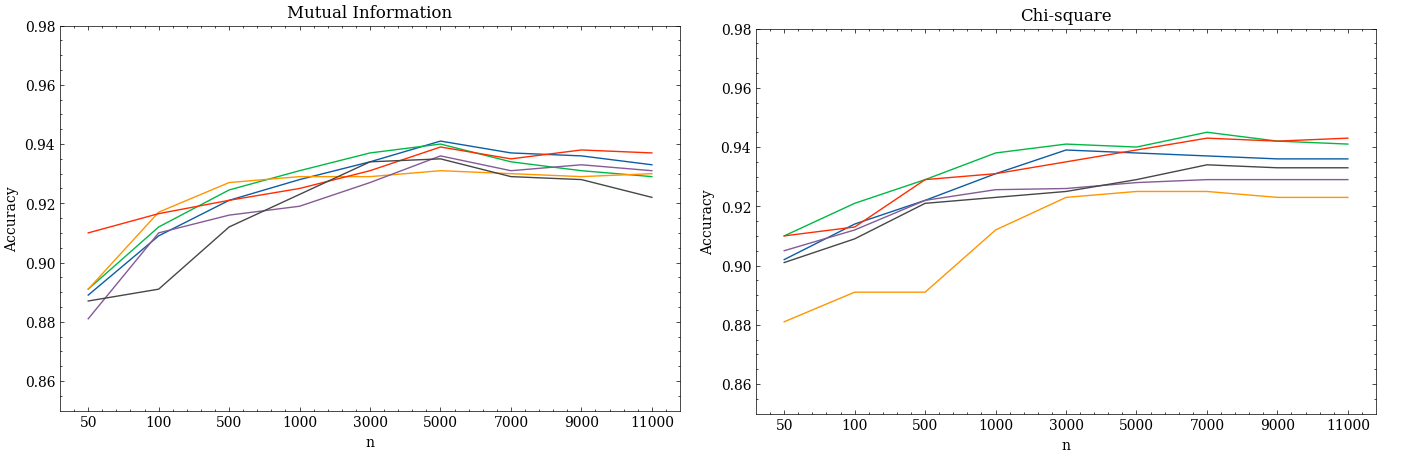}
         \caption{bigram}
         \label{fig:three sin x}
     \end{subfigure}
     \hfill
     \begin{subfigure}[b]{\columnwidth}
         \centering
         \includegraphics[width=\columnwidth]{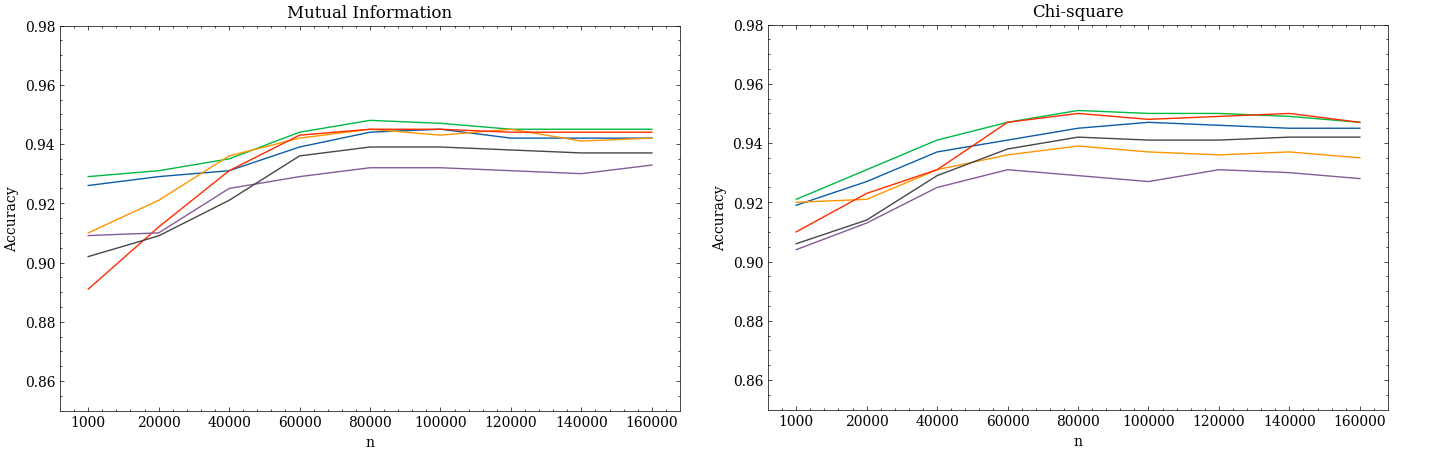}
         \caption{trigram}
         \label{fig:five over x}
     \end{subfigure}
        \caption{\textcolor{black}{Mean accuracy of ML models trained on system call features represented as unigrams, bigrams and trigrams when using mutual information and chi-square to select features, showing the effect of varying the feature count n.}}
        \label{fig:sys1}
\end{figure}

Neverthelesss, our results suggest that larger n-grams are more effective, and that RFs are again the best performing classifiers. A Kruskal-Wallis test between the models in Table \ref{tab:sys1} showed that there is a statistically significant difference in accuracy (H = 46.64, p $<$ 0.05). Post hoc Dunn’s tests confirmed that RFs outperformed the decision trees and linear regression models. Mutual information and chi-square selected the best features for malware detection. Fig. \ref{fig:sys1} shows the accuracy rates of the ML models with different number of features selected by mutual information and chi-square. The accuracy does not increase significantly, or in some cases dropped, after 90 features for unigrams, 5,000 features for bigrams using mutual information, 7,000 features for bigrams using chi-square, and 80,000 features for trigrams.
 
\begin{table}[t]

\centering
\resizebox{\textwidth}{!}{%
\begin{tabular}{llllllllllll}
\toprule
\multicolumn{6}{c}{\textbf{Malik   et al.\ Results}}                                                                                                                & \multicolumn{6}{c}{\textbf{Our Results}}                                                                                                                                                                                \\ \cmidrule(lr){1-6} \cmidrule(lr){7-12}
\multicolumn{1}{l}{\textbf{Classifier}} & \multicolumn{1}{l}{\textbf{Accuracy}} & \multicolumn{1}{l}{\textbf{Precision}} & \multicolumn{1}{l}{\textbf{F1-Score}} & \multicolumn{1}{l}{\textbf{TPR}}   & \textbf{TNR} & \multicolumn{1}{l}{\textbf{Classifier}} & \multicolumn{1}{l}{\textbf{Accuracy}}          & \multicolumn{1}{l}{\textbf{Precision}}         & \multicolumn{1}{l}{\textbf{F1-Score}}          & \multicolumn{1}{l}{\textbf{TPR}}               & \textbf{TNR}                \\ 
\midrule
\multicolumn{1}{l}{kNN (3)}    & \multicolumn{1}{l}{-}        & \multicolumn{1}{l}{\textbf{0.852}}     & \multicolumn{1}{l}{0.846}    & \multicolumn{1}{l}{0.839} & -   & \multicolumn{1}{l}{kNN (10)}   & \multicolumn{1}{l}{\textbf{0.820}$ \pm 0.003$ } & \multicolumn{1}{l}{\textbf{0.869}$ \pm 0.002$} & \multicolumn{1}{l}{\textbf{0.806}$ \pm 0.002$} & \multicolumn{1}{l}{\textbf{0.752}$ \pm 0.002$} & \textbf{0.887}$ \pm 0.004$  \\ 
\multicolumn{1}{l}{LSTM (128)} & \multicolumn{1}{l}{-}        & \multicolumn{1}{l}{0.786}     & \multicolumn{1}{l}{\textbf{0.859}}    & \multicolumn{1}{l}{\textbf{0.946}} & -   & \multicolumn{1}{l}{LSTM (128)} & \multicolumn{1}{l}{0.7521$ \pm 0.001$} & \multicolumn{1}{l}{0.753$ \pm 0.003$} & \multicolumn{1}{l}{0.749$ \pm 0.002$} & \multicolumn{1}{l}{0.732$ \pm 0.002$} & 0.758 $ \pm 0.001$ \\ 
\bottomrule
\end{tabular}%
}\caption{Result comparison between Malik et al.\ \cite{Malik2019} and our reimplementation, \textcolor{black}{showing the performance of kNN models trained on system call frequency features and LSTM models trained on system call sequences}}
\label{tab:sys3}
\end{table}

\begin{table}[t]

\centering
\resizebox{\textwidth}{!}{%
\begin{tabular}{lllllllllll}
\toprule
&\multicolumn{5}{c}{\textbf{Usage}}                                                                                                                                                                                                                                          & \multicolumn{5}{c}{\textbf{Frequency}}                                                                                                                                      \\ 
\cmidrule(lr){2-6} \cmidrule(lr){7-11}
\multicolumn{1}{l}{\textbf{Classifier}} & \multicolumn{1}{l}{\textbf{Accuracy}}          & \multicolumn{1}{l}{\textbf{Precision}}         & \multicolumn{1}{l}{\textbf{F1-Score}}          & \multicolumn{1}{l}{\textbf{TPR}}               & \textbf{TNR}               & \multicolumn{1}{l}{\textbf{Accuracy}} & \multicolumn{1}{l}{\textbf{Precision}} & \multicolumn{1}{l}{\textbf{F1-Score}} & \multicolumn{1}{l}{\textbf{TPR}} & \textbf{TNR} \\ 
\midrule
\multicolumn{1}{l}{SVM}                 & \multicolumn{1}{l}{0.830$ \pm 0.002$}          & \multicolumn{1}{l}{0.843$ \pm 0.002$}          & \multicolumn{1}{l}{0.827$ \pm 0.002$}          & \multicolumn{1}{l}{0.811$ \pm 0.003$}          & 0.849$ \pm 0.003$          & \multicolumn{1}{l}{0.821$ \pm 0.002$}             & \multicolumn{1}{l}{0.823$ \pm 0.004$}              & \multicolumn{1}{l}{0.824$ \pm 0.002$}             & \multicolumn{1}{l}{0.810$ \pm 0.003$}        & \textbf{0.832}$ \pm 0.003$        \\ 
\multicolumn{1}{l}{RF}       & \multicolumn{1}{l}{\textbf{0.845$ \pm 0.002$}} & \multicolumn{1}{l}{0.858$ \pm 0.001$}          & \multicolumn{1}{l}{\textbf{0.842$ \pm 0.002$}} & \multicolumn{1}{l}{\textbf{0.826$ \pm 0.002$}} & 0.863$ \pm 0.001$          & \multicolumn{1}{l}{\textbf{0.839}$ \pm 0.002$}             & \multicolumn{1}{l}{\textbf{0.835}$ \pm 0.002$}              & \multicolumn{1}{l}{\textbf{0.833}$ \pm 0.002$}             & \multicolumn{1}{l}{\textbf{0.856}$ \pm 0.002$}        & 0.821$ \pm 0.002$        \\ 
\multicolumn{1}{l}{DT}                  & \multicolumn{1}{l}{0.812$ \pm 0.003$}          & \multicolumn{1}{l}{0.818$ \pm 0.002$}          & \multicolumn{1}{l}{0.810$ \pm 0.002$}          & \multicolumn{1}{l}{0.803$ \pm 0.002$}          & 0.821$ \pm 0.001$          & \multicolumn{1}{l}{0.802$ \pm 0.002$}             & \multicolumn{1}{l}{0.809$ \pm 0.003$}              & \multicolumn{1}{l}{0.806$ \pm 0.003$}             & \multicolumn{1}{l}{0.802$ \pm 0.001$}        & 0.815$ \pm 0.003$        \\ 
\multicolumn{1}{l}{kNN}                 & \multicolumn{1}{l}{0.820$ \pm 0.003$}          & \multicolumn{1}{l}{\textbf{0.869$ \pm 0.002$}} & \multicolumn{1}{l}{0.806$ \pm 0.002$}          & \multicolumn{1}{l}{0.752$ \pm 0.002$}          & \textbf{0.887$ \pm 0.004$} & \multicolumn{1}{l}{0.823$ \pm 0.002$}             & \multicolumn{1}{l}{0.826$ \pm 0.003$}              & \multicolumn{1}{l}{0.828$ \pm 0.003$}             & \multicolumn{1}{l}{0.822$ \pm 0.001$}        & 0.828$ \pm 0.001$        \\ 
\bottomrule
\end{tabular}%
}\caption{\textcolor{black}{Comparison of core ML models trained on usage and frequency-based system call features}}
\label{tab:sys5}
\end{table}

We reimplemented one more study that built models solely from system call features, carried out by \textbf{Malik et al.\ \cite{Malik2019}}. This study is notable for using a different group of ML models and a different representation of the features. In particular, they used a 3-layer LSTM to classify malware based on the complete sequence of system calls used by an application, comparing this against a kNN model trained on a system call usage feature set. Both $k$ and the number of neurons per LSTM layer were optimised. Table \ref{tab:sys3} shows the results. Like Malik et al., we found that there was no benefit to using a more complex, slower to train, sequence-based model, since the best kNN model (with $k=3$) substantially out-performed the best LSTM (with 128 neurons per layer). However, our LSTM model performed much better than the model reported by Malik et al.
 
 
 
 

To get a slightly broader perspective, we also trained our standard set of ML models using both the system call usage set and the corresponding system call frequency feature set. The results are shown in Table \ref{tab:sys5}. Reflecting our previous findings, the effect of moving between a usage-based and a frequency-based representation is relatively small --- although, generally, the best models were found when using the usage-based feature set. Increasing the model set shows that kNN is not the best choice, with RF models leading to the best accuracy and F1-scores for both feature representations. This was confirmed by Kruskal-Wallis (H = 22.41, p $<$ 0.05) and Dunn's tests, which showed RF to be statistically better than the other models, with no significant difference between RF usage and frequency models.


\subsection{\textcolor{black}{Use of dynamic API call features}}

Perhaps an obvious question at this point is whether API calls, which we found to be the most effective class of features in models resulting from static analysis, would also be beneficial for dynamic analysis. \textbf{Afonso et al.\ \cite{Afonso2015}} addressed this question to a certain extent by using information about both system calls and API calls in their models. Specifically, they used call frequencies as features, and a dataset of 2,968 benign and 4,552 malicious applications. We reimplemented their approach using SVM, RF, decision trees, kNN and Naïve Bayes. Afonso et al.\ also considered several other models, but those performed poorly in their study, so we did not reimplement them.



\begin{table}[b]

\centering
\resizebox{\textwidth}{!}{%
\begin{tabular}{lllllllllll}
\toprule
&\multicolumn{5}{c}{\textbf{Afonso   et al.\ Results}}                                                                                                                                                                   & \multicolumn{5}{c}{\textbf{Our Results}}                                                                                                                                                                                         \\ \cmidrule(lr){2-6} \cmidrule(lr){7-11}
 \multicolumn{1}{l}{\textbf{Classifier}} & 
 \multicolumn{1}{l}{\textbf{Accuracy}} & \multicolumn{1}{l}{\textbf{Precision}} & \multicolumn{1}{l}{\textbf{F1-Score}} & \multicolumn{1}{l}{\textbf{TPR}} & \textbf{TNR} &\multicolumn{1}{l}{\textbf{Accuracy}} & \multicolumn{1}{l}{\textbf{Precision}} & \multicolumn{1}{l}{\textbf{F1-Score}} & \multicolumn{1}{l}{\textbf{TPR}}      & \textbf{TNR}      \\ \midrule
\multicolumn{1}{l}{RF}       & \multicolumn{1}{l}{0.968}             & \multicolumn{1}{l}{0.975}              & \multicolumn{1}{l}{0.968}             & \multicolumn{1}{l}{0.961}        & 0.976        &  \multicolumn{1}{l}{\textbf{0.962}$ \pm 0.002$} & \multicolumn{1}{l}{\textbf{0.968}$ \pm 0.002$}  & \multicolumn{1}{l}{\textbf{0.96}4$ \pm 0.001$} & \multicolumn{1}{l}{\textbf{0.959}$ \pm 0.003$} & \textbf{0.969}$ \pm 0.002$ \\ 
\multicolumn{1}{l}{SVM}&-&-&-&-&-&          \multicolumn{1}{l}{0.951$ \pm 0.001$}             & \multicolumn{1}{l}{0.953$ \pm 0.001$}              & \multicolumn{1}{l}{0.952$ \pm 0.001$}             & \multicolumn{1}{l}{0.949$ \pm 0.002$}             & 0.959$ \pm 0.002$             \\ 
\multicolumn{1}{l}{kNN}&-&-&-&-&-&           \multicolumn{1}{l}{0.946$ \pm 0.003$}             & \multicolumn{1}{l}{0.943$ \pm 0.003$}              & \multicolumn{1}{l}{0.946$ \pm 0.003$}             & \multicolumn{1}{l}{0.943$ \pm 0.003$}             & 0.948$ \pm 0.002$             \\ 
\multicolumn{1}{l}{DT}&-&-&-&-&-&           \multicolumn{1}{l}{0.936$ \pm 0.001$}             & \multicolumn{1}{l}{0.938$ \pm 0.001$}              & \multicolumn{1}{l}{0.935$ \pm 0.001$}             & \multicolumn{1}{l}{0.933$ \pm 0.003$}             & 0.938$ \pm 0.002$             \\ 
\multicolumn{1}{l}{Naïve Bayes}&-&-&-&-&-&           \multicolumn{1}{l}{0.857$ \pm 0.002$}             & \multicolumn{1}{l}{0.860$ \pm 0.002$}              & \multicolumn{1}{l}{0.856$ \pm 0.003$}             & \multicolumn{1}{l}{0.851$ \pm 0.002$}             & 0.862$ \pm 0.002$             \\ 
\bottomrule
\end{tabular}%
}\caption{Result comparison between Afonso et al.\ \cite{Afonso2015} and our reimplementation, \textcolor{black}{showing performance of ML models trained on a combined system and API call feature set}}
\label{tab:sys4}
\end{table}

\begin{table}[b]

\centering
\resizebox{\textwidth}{!}{%
\begin{tabular}{lllllllllll}
\toprule
& \multicolumn{5}{c}{\textbf{Usage}}                                                                                                                                                                                                                                                & \multicolumn{5}{c}{\textbf{Frequency}}                                                                                                                                                                                           \\ 
\cmidrule(lr){2-6} \cmidrule(lr){7-11}
\multicolumn{1}{l}{\textbf{Features}} & \multicolumn{1}{l}{\textbf{Accuracy}} & \multicolumn{1}{l}{\textbf{Precision}} & \multicolumn{1}{l}{\textbf{F1-Score}} & \multicolumn{1}{l}{\textbf{TPR}}      & \textbf{TNR}     & \multicolumn{1}{l}{\textbf{Accuracy}} & \multicolumn{1}{l}{\textbf{Precision}} & \multicolumn{1}{l}{\textbf{F1-Score}} & \multicolumn{1}{l}{\textbf{TPR}}      & \textbf{TNR}      \\ 
\midrule
\multicolumn{1}{l}{Full Set}                        & \multicolumn{1}{l}{0.829$ \pm 0.002$} & \multicolumn{1}{l}{0.839$ \pm 0.003$}  & \multicolumn{1}{l}{0.826$ \pm 0.003$} & \multicolumn{1}{l}{0.813$ \pm 0.003$} & 0.844$ \pm 0.001$      & \multicolumn{1}{l}{0.812$ \pm 0.002$} & \multicolumn{1}{l}{0.815$ \pm 0.001$}  & \multicolumn{1}{l}{0.812$ \pm 0.002$} & \multicolumn{1}{l}{0.802$ \pm 0.002$} & 0.831$ \pm 0.001$ \\ 
\multicolumn{1}{l}{MI}              & \multicolumn{1}{l}{\textbf{0.855}$ \pm 0.003$} & \multicolumn{1}{l}{\textbf{0.858}$ \pm 0.002$}  & \multicolumn{1}{l}{\textbf{0.851}$ \pm 0.003$} & \multicolumn{1}{l}{\textbf{0.855}$ \pm 0.001$} & \textbf{0.845}$ \pm 0.001$      & \multicolumn{1}{l}{0.842$ \pm 0.003$} & \multicolumn{1}{l}{0.841$ \pm 0.001$}  & \multicolumn{1}{l}{0.833$ \pm 0.003$} & \multicolumn{1}{l}{0.816$ \pm 0.003$} & \textbf{0.846}$ \pm 0.003$ \\ 
\multicolumn{1}{l}{PCC}                             & \multicolumn{1}{l}{0.842$ \pm 0.002$} & \multicolumn{1}{l}{0.843$ \pm 0.002$}  & \multicolumn{1}{l}{0.848$ \pm 0.002$} & \multicolumn{1}{l}{0.834$ \pm 0.002$} & 0.844$ \pm 0.003$      & \multicolumn{1}{l}{\textbf{0.846}$ \pm 0.002$} & \multicolumn{1}{l}{\textbf{0.851}$ \pm 0.002$}  & \multicolumn{1}{l}{\textbf{0.855}$ \pm 0.003$} & \multicolumn{1}{l}{\textbf{0.842}$ \pm 0.002$} & 0.833$ \pm 0.001$ \\ 
\multicolumn{1}{l}{VT}              & \multicolumn{1}{l}{0.833$ \pm 0.002$} & \multicolumn{1}{l}{0.845$ \pm 0.002$}  & \multicolumn{1}{l}{0.840$ \pm 0.002$} & \multicolumn{1}{l}{0.845$ \pm 0.003$} & 0.831$ \pm 0.002$      & \multicolumn{1}{l}{0.845$ \pm 0.001$} & \multicolumn{1}{l}{0.843$ \pm 0.001$}  & \multicolumn{1}{l}{0.843$ \pm 0.003$} & \multicolumn{1}{l}{0.831$ \pm 0.003$} & 0.829$ \pm 0.001$ \\ 
\multicolumn{1}{l}{CS}                      & \multicolumn{1}{l}{0.829$ \pm 0.002$} & \multicolumn{1}{l}{0.829$ \pm 0.002$}  & \multicolumn{1}{l}{0.806$ \pm 0.002$} & \multicolumn{1}{l}{0.814$ \pm 0.002$} & 0.834$ \pm 0.001$      & \multicolumn{1}{l}{0.826$ \pm 0.003$} & \multicolumn{1}{l}{0.821$ \pm 0.002$}  & \multicolumn{1}{l}{0.822$ \pm 0.002$} & \multicolumn{1}{l}{0.804$ \pm 0.002$} & 0.834$ \pm 0.002$ \\ 
\bottomrule
\end{tabular}%
}\caption{\textcolor{black}{Performance of RF models trained only on API calls, also comparing the benefit of usage and frequency-based features}}
\label{tab:api}
\end{table}

Table \ref{tab:sys4} shows the results of our reimplementation. The RF model does particularly well, and has comparable performance to the best static approaches. Notably, its performance is significantly higher than the system call-only models reported in the previous section (Kruksal-Wallis, H = 24.57, p $<$ 0.05, supported by Dunn’s tests). However, it should be noted those models used n-gram and sequence-based representations, rather than API call frequencies.

This still leaves the question of whether dynamic API calls alone can be used to build effective models. To address this, Table \ref{tab:api} shows the performance of RF models trained solely on API call frequencies. The results are shown for both the full feature set and for feature sets reduced using the core feature selection algorithms. It can be seen that the performance of these models is a lot lower than the models that used both system and API calls. This seems to suggest that whilst the combination of dynamic system calls and dynamic API calls is more effective than either of these alone, information about dynamic API calls is not sufficient to train good models. This is an interesting finding, given that information about \textit{static} API calls was sufficient to train good models. The poor performance of using the dynamic API calls alone could also be explained by the fact that while all system calls are monitored, only a subset of the API calls can be monitored due to their large number (134,207). 

We also show, in Table \ref{tab:api}, the performance of models based upon usage (rather than frequency) of dynamic system and API calls. In common with our results from earlier studies, this shows that the choice of usage-based or frequency-based features has only a minor impact upon model performance. However, it is perhaps more notable within a dynamic context, where frequency-based features might be expected to contain information that were not available through static analysis alone, and highlights the fact that this additional information is not necessarily required in order to train good malware detection models.

\subsection{\textcolor{black}{Use of network traffic features}} \label{sec:network}

\begin{table}[b]

\centering
\resizebox{\textwidth}{!}{%
\begin{tabular}{llllllllllll}
\toprule
\multicolumn{6}{c}{\textbf{Zulkifli   et al.\ Results}}                                                                                                                    & \multicolumn{6}{c}{\textbf{Our Results}}                                                                                                                                                                                         \\ \cmidrule(lr){1-6} \cmidrule(lr){7-12}
\multicolumn{1}{l}{\textbf{Dataset}} & \multicolumn{1}{l}{\textbf{Classifier}} & \multicolumn{1}{l}{\textbf{Accuracy}} & \multicolumn{1}{l}{\textbf{F1-Score}} & \multicolumn{1}{l}{\textbf{TPR}} & \textbf{TNR} & \multicolumn{1}{l}{\textbf{Classifier}} & \multicolumn{1}{l}{\textbf{Accuracy}} & \multicolumn{1}{l}{\textbf{Precision}} & \multicolumn{1}{l}{\textbf{F1-Score}} & \multicolumn{1}{l}{\textbf{TPR}}      & \textbf{TNR}      \\ 
\midrule
\multicolumn{1}{l}{drebin}           & \multicolumn{1}{l}{DT}           &\multicolumn{1}{l}{\textbf{0.984}}             & \multicolumn{1}{l}{-}                 & \multicolumn{1}{l}{0.920}        & -            & \multicolumn{1}{l}{SVM}                 & \multicolumn{1}{l}{0.961$ \pm 0.001$} & \multicolumn{1}{l}{0.958$ \pm 0.002$}  & \multicolumn{1}{l}{0.957$ \pm 0.002$} & \multicolumn{1}{l}{0.952$ \pm 0.002$} & 0.961$ \pm 0.002$ \\ 
\multicolumn{1}{l}{contagiodumpset}  & \multicolumn{1}{l}{DT}           &\multicolumn{1}{l}{0.976}             & \multicolumn{1}{l}{-}                 & \multicolumn{1}{l}{0.920}        & -            & \multicolumn{1}{l}{RF}       & \multicolumn{1}{l}{\textbf{0.977}$ \pm 0.003$} & \multicolumn{1}{l}{\textbf{0.969}$ \pm 0.003$}  & \multicolumn{1}{l}{\textbf{0.981}$ \pm 0.003$} & \multicolumn{1}{l}{\textbf{0.978}$ \pm 0.002$} & \textbf{0.971}$ \pm 0.002$ \\ 
\multicolumn{6}{l}{\multirow{2}{*}{}}                                                                                                                                     & \multicolumn{1}{l}{DT}                  & \multicolumn{1}{l}{0.973$ \pm 0.002$} & \multicolumn{1}{l}{0.967$ \pm 0.003$}  & \multicolumn{1}{l}{0.977$ \pm 0.001$} & \multicolumn{1}{l}{0.973$ \pm 0.002$} & 0.969$ \pm 0.001$ \\ 
\multicolumn{6}{l}{}                                                                                                                                                      & \multicolumn{1}{l}{kNN}                 & \multicolumn{1}{l}{0.961$ \pm 0.001$} & \multicolumn{1}{l}{0.962$ \pm 0.003$}  & \multicolumn{1}{l}{0.959$ \pm 0.002$} & \multicolumn{1}{l}{0.968$ \pm 0.001$} & 0.962$ \pm 0.001$ \\ 
\bottomrule
\end{tabular}%
}\caption{Result comparison between Zulkifli et al.\ \cite{Zulkifli2018} and our reimplementation, \textcolor{black}{showing the performance of ML models trained using their TCP-based network traffic feature set}}
\label{tab:net1}
\end{table}

\begin{table}[b]

\centering
\resizebox{\textwidth}{!}{%
\begin{tabular}{lllllllllllll}
\toprule
&\multicolumn{6}{c}{\textbf{Wang   et al.\ Results}}                                                                                                                                                                                                             &                     \multicolumn{6}{c}{\textbf{Our Results}}                                                                                                                                              \\ 
\cmidrule(lr){2-7} \cmidrule(lr){8-13}
\multicolumn{1}{l}{\textbf{Dataset}} & \multicolumn{1}{l}{\textbf{Classifier}} & \multicolumn{1}{l}{\textbf{Accuracy}} & \multicolumn{1}{l}{\textbf{Precision}} & \multicolumn{1}{l}{\textbf{F1-Score}} & \multicolumn{1}{l}{\textbf{TPR}} & \textbf{TNR} & \textbf{Classifier} & \multicolumn{1}{l}{\textbf{Accuracy}} & \multicolumn{1}{l}{\textbf{Precision}} & \multicolumn{1}{l}{\textbf{F1-Score}} & \multicolumn{1}{l}{\textbf{TPR}}      & \textbf{TNR}      \\
\midrule
\multicolumn{1}{l}{TCP}              & \multicolumn{1}{l}{DT}                  & \multicolumn{1}{l}{0.982}             & \multicolumn{1}{l}{-}                  & \multicolumn{1}{l}{-}                 & \multicolumn{1}{l}{-}            & -            & RF       & \multicolumn{1}{l}{\textbf{0.973}$ \pm 0.002$} & \multicolumn{1}{l}{\textbf{0.967}$ \pm 0.003$}  & \multicolumn{1}{l}{\textbf{0.977}$ \pm 0.001$} & \multicolumn{1}{l}{\textbf{0.973}$ \pm 0.002$} & \textbf{0.969}$ \pm 0.001$ \\ 
\multicolumn{1}{l}{HTTP}             & \multicolumn{1}{l}{DT}                  & \multicolumn{1}{l}{\textbf{0.997}}             & \multicolumn{1}{l}{-}                  & \multicolumn{1}{l}{-}                 & \multicolumn{1}{l}{-}            & -            & RF       & \multicolumn{1}{l}{0.963$ \pm 0.002$} & \multicolumn{1}{l}{0.961$ \pm 0.002$}  & \multicolumn{1}{l}{0.965$ \pm 0.002$} & \multicolumn{1}{l}{0.965$ \pm 0.002$} & 0.961$ \pm 0.003$ \\ 
\bottomrule
\end{tabular}%
}\caption{Result comparison between Wang et al.\ \cite{Wang2016} and our reimplementation, \textcolor{black}{showing the performance of ML models trained using their TCP and HTTP-based network traffic feature sets. For our reimplementation, we only show results of the best ML model for each feature set.}}
\label{tab:net2}
\end{table}

\begin{table}[!b]
\centering
\resizebox{\columnwidth}{!}{%
\begin{tabular}{llllll}
\toprule
\textbf{Classifier} & \textbf{Accuracy} & \textbf{Precision} & \textbf{F1-Score} & \textbf{TPR} & \textbf{TNR} \\ 
\midrule
\multicolumn{6}{l}{\textit{\textcolor{black}{Features:} HTTP}}\\
SVM                 & 0.941$ \pm 0.003$               & 0.942$ \pm 0.001$                & 0.941$ \pm 0.002$               & 0.943$ \pm 0.002$          & 0.946$ \pm 0.003$          \\ 
RF       &  0.963$ \pm 0.002$ & 0.961$ \pm 0.002$  & 0.965$ \pm 0.002$ & 0.965$ \pm 0.002$ & 0.961$ \pm 0.003$             \\ 
DT                  & 0.932$ \pm 0.002$              & 0.935$ \pm 0.003$                & 0.937$ \pm 0.004$             & 0.933$ \pm 0.001$          & 0.932$ \pm 0.002$          \\ 
kNN                 & 0.951$ \pm 0.002$              & 0.956$ \pm 0.003$               & 0.955$ \pm 0.004$              & 0.949$ \pm 0.003$          & 0.959$ \pm 0.002$          \\ 
\midrule
\multicolumn{6}{l}{\textit{\textcolor{black}{Features:} TCP + HTTP}}\\
SVM                 & 0.938$ \pm 0.002$              & 0.933$ \pm 0.003$               & 0.934$ \pm 0.004$              & 0.933$ \pm 0.001$          & 0.932$ \pm 0.003$          \\ 
RF       &  0.952$ \pm 0.002$               & 0.953$ \pm 0.002$               & 0.956$ \pm 0.002$              & 0.955$ \pm 0.004$         & 0.951$ \pm 0.002$         \\ 
DT                  & 0.929$ \pm 0.002$               & 0.926$ \pm 0.002$               & 0.925$ \pm 0.003$               & 0.922$ \pm 0.002$         & 0.925$ \pm 0.003$          \\ 
kNN                 & \textbf{0.967}$ \pm 0.004$                   &  \textbf{0.964}$ \pm 0.002$                   & \textbf{0.967}$ \pm 0.001$                    & \textbf{0.969}$ \pm 0.002$              &  \textbf{0.964}$ \pm 0.003$             \\ 
\bottomrule
\end{tabular}%
}
\caption{\textcolor{black}{Performance of all core ML models trained on HTTP features or a combination of both TCP and HTTP features. Note that models trained only on TCP features are shown in Table \ref{tab:net1}.}}
\label{tab:net3}
\end{table}
 
Dynamic analysis tools can also monitor network traffic during the execution of an application, and this provides another potentially useful source of information about the application's behaviour. Previous works have focused on using network traffic features extracted from TCP and HTTP protocols. 
 
\textbf{Zulkifli et al.\ \cite{Zulkifli2018}} used DT with TCP-based features in their study. They used a small dataset of 500 benign and 200 malicious applications, the latter assembled from two existing malware collections (Drebin and contagiodumpset). Notably, all the applications were executed on a real Android device.
They extracted the following features:
 
\begin{itemize}
    \item The average packet size during the runtime of the dynamic analysis
    \item The average of the number of packets sent per flow between host and client during the runtime of the dynamic analysis
    \item The average of the number of packets sent per flow between client and host during the runtime of the dynamic analysis
    \item The average packet size in terms of bytes sent per flow during the runtime of the dynamic analysis
    \item The average packet size in terms of bytes received per flow during the runtime of the dynamic analysis
    \item Ratio of the number of incoming bytes to outgoing bytes of every packet during the runtime of the dynamic analysis
    \item The average number of packets received per second during the course of the dynamic analysis
\end{itemize}

We extracted the same seven features for each of the applications in our dataset and trained the four core ML models. Table \ref{tab:net1} shows the results, comparing our reimplementation with Zulkifli et al., who reported results separately for each of their two malware datasets. The best performing model was RF. Significantly, its accuracy and F1-score is the highest we have seen so far, suggesting that network traffic is an important source of information within a malware detection context. 

Zulkifli et al.\ used a  small dataset, presumably because they used a real Android device to perform their dynamic analysis. In this respect, it is interesting to see that our results show similar accuracy to the original paper. This supports the case for using an emulator to perform dynamic analysis, especially since an emulator is both a lot faster to configure and allows the OS image to be reloaded each time a dynamic analysis of an application is carried out, meaning that the analysis can be carried out in a controlled environment.
 
To further investigate network traffic features, we reimplemented another study by \textbf{Wang et al.\ \cite{Wang2016}} involving network data. Wang et al.\ extracted both TCP and HTTP features from network traffic. Wang et al.\ used the same set of TCP features as Zulkifli et al., with the exception of average packet size. The features they extracted from HTTP headers were:
\begin{itemize}
    \item Host: This field specifies the host and port number that the request is sent to.
    \item Request-URI: This is the uniform resource identifier from the requested source.
    \item Request-Method: The action to be performed for a given resource. This includes GET and POST requests.
    \item User-Agent: This holds information about the application, OS, vendor and version of the requesting user agent.
\end{itemize}
 
Wang et al.\ used a malware dataset consisting of 5,560 applications from the Drebin dataset and 8,312 benign applications downloaded from several application markets. The authors separately trained decision trees on both the HTTP and TCP datasets. In our reimplementation, we expanded on the original study by training our core set of ML models on the HTTP and TCP features both individually and in combination.

Table \ref{tab:net2} shows the results, and compares them against those from the original study. For our reimplementation, Table \ref{tab:net2} shows the results of the best ML models, which were RFs for both the HTTP and TCP feature sets. We observed no benefit to combining the two feature sets, but, for completeness, we also show results for the other models for both the HTTP and combined feature sets in Table \ref{tab:net3}. Whilst the results show good discrimination between benign applications and malware, we did not see the exceptionally high level of accuracy reported by Wang et al.\ for HTTP features, and in fact found these to be less discriminative than TCP features. These conclusions are supported by a positive Kruskal-Wallis test (H = 14.29, p $<$ 0.05) and Dunn's tests.
 



Although network traffic can be used to train accurate models, it is worth noting that consistently generating network traffic in an application is a major challenge. For instance, a network connection issue may result in the application not generating any network traffic. Furthermore, depending on the coverage of the dynamic analysis, i.e. the different activities invoked during the analysis, this may impact the network traffic captured. 

\subsection{Comparison of dynamic modelling approaches}

To identify the best combination of ML models and feature selection algorithm on dynamic features, we carried out a Kruskal-Wallis test followed by a pairwise comparison between the models from section \ref{sec:dynamic}. This showed that the network traffic models from Table \ref{tab:net3} were significantly better than the others. However, as it can be a challenge to generate network traffic in dynamic analysis, it is useful to identify other well-performing dynamic approaches. Statistically, the best of the rest were the RF and XGBoost models trained on system call trigrams, selected using mutual information and chi-square, respectively.
 
RF, as was the case for static features, was the best performing model for most of the dynamic features. Mutual information and chi-square selected the most relevant features when used with both system calls and API calls. However, the results indicate that using static features generally produced the best results, with the notable exception of network traffic. System calls did lead to results comparable to the best performing static feature, API calls; however, the computational effort and time required to carry out a system call analysis probably outweighs the benefits.
 
\section{\textcolor{black}{Reimplementation and extension of hybrid modelling approaches}} \label{sec:hybrid}

\begin{table}[t]

\centering
\resizebox{\textwidth}{!}{%
\begin{tabular}{lllllllllll}
\toprule
& \multicolumn{5}{c}{\textbf{Kandukuru and Sharma Results}}                                                                                                                                                              & \multicolumn{5}{c}{\textbf{Our Results}}                                                                                                                                                                                         \\ 
\cmidrule(lr){2-6} \cmidrule(lr){7-11}
\multicolumn{1}{l}{\textbf{Classifier}} & \multicolumn{1}{l}{\textbf{Accuracy}} & \multicolumn{1}{l}{\textbf{Precision}} & \multicolumn{1}{l}{\textbf{F1-Score}} & \multicolumn{1}{l}{\textbf{TPR}} & \textbf{TNR} & \multicolumn{1}{l}{\textbf{Accuracy}} & \multicolumn{1}{l}{\textbf{Precision}} & \multicolumn{1}{l}{\textbf{F1-Score}} & \multicolumn{1}{l}{\textbf{TPR}}      & \textbf{TNR}      \\ 
\midrule
\multicolumn{1}{l}{DT}                  & \multicolumn{1}{l}{0.956}             & \multicolumn{1}{l}{-}                  & \multicolumn{1}{l}{-}                 & \multicolumn{1}{l}{0.95}         & 0.96         & \multicolumn{1}{l}{0.921$ \pm 0.003$}             & \multicolumn{1}{l}{0.923$ \pm 0.003$}              & \multicolumn{1}{l}{0.923$ \pm 0.002$}             & \multicolumn{1}{l}{0.926$ \pm 0.003$}             & 0.925$ \pm 0.001$             \\ 
\multicolumn{1}{l}{RF}&-&-&-&-&-& \multicolumn{1}{l}{\textbf{0.943}$ \pm 0.001$} & \multicolumn{1}{l}{\textbf{0.941}$ \pm 0.002$}  & \multicolumn{1}{l}{\textbf{0.945}$ \pm 0.002$} & \multicolumn{1}{l}{\textbf{0.948}$ \pm 0.002$} & \textbf{0.940}$ \pm 0.002$ \\ 
\multicolumn{1}{l}{SVM}&-&-&-&-&-&  \multicolumn{1}{l}{0.933$ \pm 0.002$}             & \multicolumn{1}{l}{0.936$ \pm 0.001$}              & \multicolumn{1}{l}{0.935$ \pm 0.002$}             & \multicolumn{1}{l}{0.939$ \pm 0.001$}             & 0.937$ \pm 0.003$             \\ 
\multicolumn{1}{l}{kNN}&-&-&-&-&-&  \multicolumn{1}{l}{0.936$ \pm 0.002$}             & \multicolumn{1}{l}{0.938$ \pm 0.002$}              & \multicolumn{1}{l}{0.939$ \pm 0.002$}             & \multicolumn{1}{l}{0.941$ \pm 0.004$}             & 0.937$ \pm 0.002$             \\ 
\bottomrule
\end{tabular}%
}\caption{Result comparison between Kandukuru and Sharma \cite{Kandukuru2017} and our reimplementation, \textcolor{black}{showing the performance of ML models trained on dynamic network traffic features and static permissions}}
\label{tab:hyb1}
\end{table}

\begin{table}[b]

\centering
\resizebox{\columnwidth}{!}{%
 
\begin{tabular}{llllll}
\toprule
\textbf{Classifier} & \textbf{Accuracy} & \textbf{Precision} & \textbf{F1-Score} & \textbf{TPR} & \textbf{TNR} \\ 
\midrule
SVM                 & 0.937$ \pm 0.01$             & 0.937$ \pm 0.02$               & 0.937$ \pm 0.02$               & 0.923$ \pm 0.01$         & 0.951$ \pm 0.02$          \\ 
RF       & \textbf{0.948}$ \pm 0.02$               & \textbf{0.949}$ \pm 0.01$               & \textbf{0.946}$ \pm 0.01$              & \textbf{0.935}$ \pm 0.01$         & \textbf{0.959}$ \pm 0.02$         \\ 
Naïve Bayes         & 0.860$ \pm 0.02$              & 0.861$ \pm 0.02$                & 0.860$ \pm 0.02$               & 0.831$ \pm 0.01$         & 0.889$ \pm 0.01$         \\ 
Logistic Regression & 0.937$ \pm 0.02$              & 0.938$ \pm 0.01$               & 0.936$ \pm 0.02$               & 0.925$ \pm 0.01$         & 0.949$ \pm 0.02$         \\ 
kNN                 & 0.921$ \pm 0.02$               & 0.924$ \pm 0.02$                & 0.922$ \pm 0.03$               & 0.886$ \pm 0.01$         & 0.958$ \pm 0.01$         \\ 
DT      & 0.919$ \pm 0.02$               & 0.919$ \pm 0.02$                & 0.917$ \pm 0.01$              & 0.923$ \pm 0.02$          & 0.917$ \pm 0.02$          \\ 
\bottomrule
\end{tabular}%
}\caption{Reimplementation of Kapratwar et al.’s \cite{Kapratwar2017} approach, \textcolor{black}{in which models are trained on static permissions and dynamic system call features}}
\label{tab:hyb2a}
\end{table}

\begin{table}[!b]

\centering
\resizebox{\columnwidth}{!}{%
\begin{tabular}{lllllll}
\toprule
\textbf{Classifier} & \textbf{Features} & \textbf{Accuracy} & \textbf{Precision} & \textbf{F1-Score} & \textbf{TPR} & \textbf{TNR} \\ 
\midrule
SVM                 & 300                      & 0.936$ \pm 0.02$  & 0.936$ \pm 0.01$              & 0.935$ \pm 0.02$             & 0.921$ \pm 0.01$        & 0.951$ \pm 0.02$        \\ 
RF       & 250                      & \textbf{0.948}$ \pm 0.02$             & \textbf{0.948}$ \pm 0.03$             & \textbf{0.949}$ \pm 0.03$             & \textbf{0.935}$ \pm 0.01$        & 0.961$ \pm 0.02$        \\ 
Naïve Bayes         & 100                      & 0.865$ \pm 0.03$             & 0.868$ \pm 0.02$              & 0.864$ \pm 0.01$             & 0.825$ \pm 0.01$        & 0.907$ \pm 0.02$        \\ 
Logistic Regression & 300                      & 0.936$ \pm 0.02$             & 0.937$ \pm 0.02$              & 0.935$ \pm 0.02$             & 0.923$ \pm 0.02$        & 0.949$ \pm 0.02$        \\ 
kNN                 & 150                      & 0.922$ \pm 0.02$             & 0.925$ \pm 0.01$              & 0.922$ \pm 0.01$             & 0.883$ \pm 0.03$        & \textbf{0.962}$ \pm 0.01$        \\ 
DT                  & 200                      & 0.918$ \pm 0.02$             & 0.917$ \pm 0.02$              & 0.916$ \pm 0.02$             & 0.920$ \pm 0.03$        & 0.915$ \pm 0.01$        \\ 
\bottomrule
\end{tabular}%
}\caption{Reimplementation of Kapratwar et al.’s \cite{Kapratwar2017} approach \textcolor{black}{using reduced feature sets selected using mutual information}}
\label{tab:hyb2}
\end{table}
 
Hybrid analysis in theory uses the best of both worlds, static and dynamic features. As with dynamic analysis, hybrid studies are not as common as static analysis. In this section, we report and compare models that use hybrid features by reimplementing and extending past studies. As hybrid analysis uses dynamic features, in this section we use the slightly reduced dataset introduced in section \ref{sec:dynamic}, which consists of 53,960 benign and 53,202 malware.
 
We begin by reimplementing a study by \textbf{Kandukuru and Sharma \cite{Kandukuru2017}} that combined dynamic network traffic features with static permission features. The network traffic features included all seven of the TCP features listed in Section \ref{sec:network} and the authors used a malware dataset of 1,200 applications collected from 2010 to 2011 to train a decision tree model. In our reimplementation, we trained all four core ML models, \textcolor{black}{using a dataset of feature vectors comprising 166 permissions and the seven TCP features}. Table \ref{tab:hyb1} compares their performance with Kandukuru and Sharma's results. Since their performance is worse than models trained only on network features (see Tables \cref{tab:net1,tab:net2}), this suggests that there is no benefit to including information about static permissions.

To further investigate the use of hybrid features, we reimplemented a study based around two frequently used groups of static and dynamic features,  permissions and system calls. \textbf{Kapratwar et al.\ \cite{Kapratwar2017}} carried out experiments using these features both separately and in combination. In the case of system calls, a frequency-based representation was used. For each experiment, the authors trained RF, decision trees, Naïve Bayes, logistic regression, kNN and SVM. However, as with most past hybrid works, Kapratwar et al.\ used a small dataset of just 103 malware and 97 benign applications. Therefore, a reimplementation of this approach using a larger contemporary dataset is essential to understand the effect of combining two of the most prevalent features in Android malware detection. \textcolor{black}{The dataset used for this reimplementation included 166 permissions and 599 system calls.}

\begin{figure}[t]
\includegraphics[width=9cm]{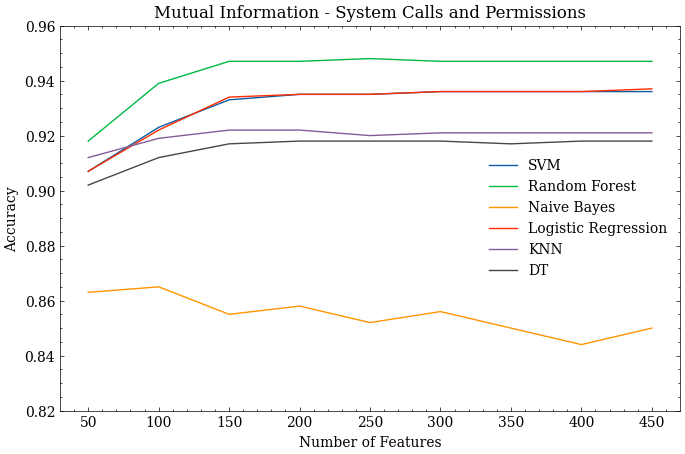}
\centering
\caption{\textcolor{black}{Effect on ML model accuracies of reducing the number of system calls and permissions features using mutual information}}
\label{fig:hyb1}
\end{figure}

The results of our reimplementation are shown in Table \ref{tab:hyb2a}, though note that we do not report the results of the original study in this case, since only AUC scores were provided. The performance of our models is comparable to the performance of models discussed earlier that used system calls or permissions separately, so again there does not appear to be any benefit to using a hybrid approach. We also used feature selection to select the most relevant features; mutual information was most useful for this, and Table \ref{tab:hyb2} summarises the performance of the best models trained on the feature sets reduced according to mutual information. This suggests that there is a only a small benefit to doing this. However, as Fig. \ref{fig:hyb1} shows, the number of features can be reduced by about a third without degrading performance. A Kruskal-Wallis test between the models in Table \ref{tab:hyb2} showed that there was a statistically significant difference in accuracy (H = 25.62, p $<$ 0.05), and post hoc Dunn’s tests showed that the best performing ML model was RF.


\section{Best performing models} \label{sec:bestmodels}

\subsection{\textcolor{black}{Comparison of existing models}}

\textcolor{black}{Table \ref{tab:ensemblemodels} lists the best classification models seen during this study. Two modelling approaches stand out in terms of the metric scores achieved when discriminating malware from benign applications. TCP network traffic models using random forests ($C_9$) achieved the highest scores in terms of accuracy, precision and TNR. Five-gram opcode models, again using random forests ($C_6$), achieved the highest scores in terms of F1 and TPR. As noted in Section \ref{sec:methodology:metrics}, TPR is particularly important for malware detection, since a benign application labelled as malware is less likely to cause as many security problems as malware labelled as benign. It is also notable that 11 of the 12 classifiers listed in Table \ref{tab:ensemblemodels} use random forests, suggesting that this is a good default choice within malware detection. This reflects the wider success of tree-based ensemble models, especially when working with pre-extracted features \cite{grinsztajn2022tree}.}

\textcolor{black}{There is also one kNN instance in the table ($C_{10}$), suggesting that there may sometimes be a benefit to using other models. However, we also observed that kNN is challenging in terms of scalability, since its inference time grows in line with the product of the number of samples and the number of features in the training data. This is an issue when using large feature sets such as API calls.}

\textcolor{black}{A notable omission within this table is the presence of any deep learning models. The deep learning approaches that we reimplemented in this study generally performed significantly less well than simpler approaches. At least within the scope of this study, this suggests they are not a good choice of model to use with pre-extracted features, especially given that} deep learning models also bring risks of overfitting, responding to spurious patterns within data, and limited transparency \cite{Lones2021}, not to mention higher training and inference times --- so in general are not likely to be a productive approach for malware detection unless they lead to a significant improvement in performance. \textcolor{black}{This reflects observations made elsewhere about the limitations of deep learning within a feature-based modelling context \cite{grinsztajn2022tree}.}



\begin{table}[h]

\centering
\resizebox{\textwidth}{!}{%
\begin{tabular}{llllllllll}
\toprule
\textbf{ID} & \textbf{Feature Set}        &\textbf{Type}            & \textbf{Model} & \textbf{Selection} & \textbf{Accuracy} & \textbf{F1-Score} & \textbf{Precision} & \textbf{TPR} & \textbf{TNR} \\ 
\midrule
$C_1$        & API & Static                        & RF       & MI & 0.967$ \pm 0.001$ & 0.969$ \pm 0.001$  & 0.967$ \pm 0.002$ & 0.963$ \pm 0.002$ & 0.970$ \pm 0.002$        \\ 
$C_2$        & API & Static                        & RF       & VT & 0.968$ \pm 0.002$ & 0.971$ \pm 0.002$  & 0.968$ \pm 0.002$ & 0.965$ \pm 0.002$ & 0.971$ \pm 0.001$        \\ 
$C_3$        & API & Static                         & RF                 & CS  &  0.961$ \pm 0.003$ & 0.915$ \pm 0.004$  & 0.954$ \pm 0.002$ & 0.967$ \pm 0.001$ & 0.953$ \pm 0.001$                \\ 
$C_4$        & Permissions   & Static                        & RF       & MI & 0.930$ \pm 0.001$ & 0.932$ \pm 0.001$  & 0.929$ \pm 0.002$ & 0.931$ \pm 0.002$ & 0.931$ \pm 0.003$         \\ 
$C_5$        & Permissions     & Static                      & RF       & VT     & 0.949$ \pm 0.003$ & 0.948$ \pm 0.002$  & 0.950$ \pm 0.001$ & 0.951$ \pm 0.001$ & 0.949$ \pm 0.002$       \\ 
$C_6$        & Opcodes (5-gram)    & Static                  & RF       & MI        & 0.973$ \pm 0.002$ & \textbf{0.979}$ \pm 0.002$  & 0.975$ \pm 0.001$ & 0.965$ \pm 0.001$ & \textbf{0.979}$ \pm 0.002$        \\ 
$C_7$        & System Calls (3-gram) & Dynamic                & RF       & CS  & 0.951$ \pm 0.002$ & 0.951$ \pm 0.001$ & 0.950$ \pm 0.002$ & 0.949$ \pm 0.003$ & 0.952$ \pm 0.001$               \\ 
$C_8$        & System Calls + API Call Usage  & Dynamic     & RF       & None & 0.961$ \pm 0.002$ & 0.964$ \pm 0.002$  & 0.963$ \pm 0.001$ & 0.957$ \pm 0.003$ & 0.965$ \pm 0.002$                       \\ 
$C_9$        & TCP   & Dynamic                & RF       & None  & \textbf{0.977}$ \pm 0.003$ & 0.969$ \pm 0.003$  & \textbf{0.981}$ \pm 0.003$ & \textbf{0.978}$ \pm 0.002$ & 0.971$ \pm 0.002$                       \\ 
$C_{10}$       & TCP + HTTP  & Dynamic           & kNN                 & None  & 0.967$ \pm 0.004$                   &  0.964$ \pm 0.002$                   & 0.967$ \pm 0.001$                    & 0.969$ \pm 0.002$              &  0.964$ \pm 0.003$                        \\ 
$C_{11}$       & TCP + Permissions & Hybrid    & RF       & None    & 0.943$ \pm 0.001$ & 0.941$ \pm 0.002$  & 0.945$ \pm 0.002$ & 0.948$ \pm 0.002$ & 0.940$ \pm 0.002$                  \\ 
$C_{12}$       & Permissions + System Calls    & Hybrid          & RF       & MI  & 0.948$ \pm 0.02$             & 0.948$ \pm 0.03$             & 0.949$ \pm 0.03$             & 0.935$ \pm 0.01$        & 0.961$ \pm 0.02$           \\ 
\bottomrule
\end{tabular}%
}\caption{\textcolor{black}{Overall best-performing models, in each case showing the features, classifier model, and feature selection algorithm used}}
\label{tab:ensemblemodels}
\end{table}

\subsection{\textcolor{black}{Comparison of ensemble models}} \label{sec:ensemble}

In the hybrid approaches discussed in the previous section, a single ML model was built from a feature set containing more than one type of feature. However, a limitation of this approach is that all the features must be used with the same model, and must be represented in a way that is compatible with the same model. In this section, we explore a more flexible approach: ensemble modelling. Ensembles are ML models that combine the outputs of other, potentially diverse, ML models. A general benefit of ensembles is that they are often able to combine the strengths of their component models in order to provide a more robust overall prediction.


\textcolor{black}{In this section, we explore whether better models can be created by combining the best standalone malware classification models. In particular, we construct voting ensembles from the best-performing classifiers listed in Table \ref{tab:ensemblemodels}. Voting ensembles are constructed by selecting an odd number of these so-called \textit{base} models}, and then outputting the majority decision, so they do not require further training. We considered each odd combination of base classifiers, and then sorted them by accuracy. Table \ref{tab:ensemble} lists the top five ensemble models according to this criterion.





\begin{table}[t]

\centering
\resizebox{\columnwidth}{!}{%
\begin{tabular}{lllllll}
\toprule
\textbf{ID} & \textbf{Classifiers} & \textbf{Accuracy} & \textbf{F1-Score} & \textbf{Precision} & \textbf{TPR} & \textbf{TNR} \\ 
\midrule
$E_1$  & $C_1$ + $C_3$ + $C_7$       & \textbf{0.978}$ \pm 0.02$  & \textbf{0.976}$ \pm 0.03$  & \textbf{0.975}$ \pm 0.01$   & \textbf{0.979}$ \pm 0.01$ & \textbf{0.977}$ \pm 0.03$ \\ 
$E_2$  & $C_2$ + $C_3$ + $C_7$       & 0.971$ \pm 0.02$  & 0.971$ \pm 0.02$  & 0.970$ \pm 0.02$   & 0.971$ \pm 0.02$ & 0.967$ \pm 0.02$ \\ 
$E_3$  & $C_1$ + $C_4$ + $C_6$        & 0.959$ \pm 0.03$  & 0.955$ \pm 0.03$  & 0.953$ \pm 0.02$   & 0.956$ \pm 0.03$ & 0.952$ \pm 0.04$ \\ 
$E_4$  & $C_3$ + $C_4$ + $C_9$ + $C_{10}$ + $C_{12}$ & 0.969$ \pm 0.01$  & 0.970$ \pm 0.04$  & 0.965$ \pm 0.04$   & 0.966$ \pm 0.01$ & 0.969$ \pm 0.03$ \\ 
$E_5$  & $C_4$ + $C_5$ + $C_8$ + $C_9$ + $C_{10}$ & 0.954$ \pm 0.02$  & 0.953$ \pm 0.01$  & 0.952$ \pm 0.02$   & 0.955$ \pm 0.01$ & 0.958$ \pm 0.03$ \\ 
\bottomrule
\end{tabular}%
}\caption{\textcolor{black}{Performance of ensemble models, showing the top five voting ensemble models formed from the overall best-performing base classifier models ($C_n$) listed in Table \ref{tab:ensemblemodels}.}}
\label{tab:ensemble}
\end{table}

Whilst all of these models perform better than the hybrid models presented in the previous section, Kruskal-Wallis and Dunn's tests showed that none of them performed significantly better than the best standalone network traffic model. However, the $E_1$ ensemble does perform better than the best API call model, and, importantly, it achieves this by combining non-network features. This is significant because it avoids the challenges (discussed in Section \ref{sec:network}) of recording network traffic during the dynamic analysis process, resulting in a model that may be more practical for real world deployment.

\section{Discussion} \label{sec:discussion}

\begin{table}[ptb]
    \centering
\resizebox{\columnwidth}{!}{%
\begin{tabular}{lrlrlrl}
\toprule
\textbf{Study} & \multicolumn{2}{c}{\textbf{Accuracy}} & \multicolumn{2}{c}{\textbf{F1-Score}} & \multicolumn{2}{c}{\textbf{TPR}} \\ \cmidrule(lr){2-3} \cmidrule(lr){4-5} \cmidrule(lr){6-7}
 & \textbf{Original} & \textbf{Ours} & \textbf{Original} & \textbf{Ours} & \textbf{Original} & \textbf{Ours} \\ \midrule
 \textit{Static analysis models} & & & & \\
Permissions and API calls \cite{Peiravian2013} & \textbf{0.969} & 0.961 & --- & 0.959 & 0.948 & \textbf{0.961} \\
Normal/dangerous permissions \cite{Wang2014} & \textbf{0.996} & 0.930 & 0.895 & \textbf{0.946} & 0.926 & \textbf{0.931} \\
Selected permissions \cite{Rathore2021} & 0.940 & \textbf{0.949} & --- & 0.959 & 0.930 & \textbf{0.951} \\
Model-selected permissions \cite{Sahin2023} & --- & 0.924 & \textbf{0.956} & 0.925 & --- & 0.932 \\
API calls with deep learning \cite{Ma2019} & --- & 0.970 & \underline{\textbf{0.990}} & 0.969 & \textbf{0.988} & \underline{0.979} \\
Reduced API calls \cite{Jung2018} & \underline{\textbf{0.999}} & 0.762 & --- & 0.799 & --- & 0.948 \\
Selected API calls \cite{Muzaffar2021} & 0.962 & \textbf{0.968} & 0.961 & \textbf{0.968} & 0.959 & \textbf{0.965} \\
Drebin feature set \cite{Arp2014} & 0.940 & \textbf{0.956} & --- & 0.895 & --- & 0.961 \\
Opcodes \cite{Kang2016} & --- & 0.973 & \textbf{0.980} & 0.975 & --- & 0.965 \\
Opcodes with 2D deep learning \cite{Xiao2019a} & \textbf{0.930} & 0.913 & \textbf{0.940} & 0.916 & \textbf{0.944} & 0.921 \\
Opcodes with 1D deep learning \cite{yeboah2022} & --- & 0.936 & \textbf{0.970} & 0.935 & \textbf{0.970} & 0.923 \\
\midrule
\textit{Dynamic analysis models} & & & & \\
Selected system calls \cite{Ananya2020} & \textbf{0.994} & 0.951 & --- & 0.950 & --- & 0.952 \\
System calls with deep learning \cite{Malik2019} & \textbf{0.852} & 0.820 & \textbf{0.859} & 0.806 & \textbf{0.839} & 0.752 \\
System and API calls \cite{Afonso2015} & \textbf{0.968} & 0.962 & \textbf{0.968} & 0.964 & \textbf{0.961} & 0.959 \\
TCP network features \cite{Zulkifli2018} & \textbf{0.984} & 0.977 & --- & \underline{0.981} & 0.920 & \textbf{0.978} \\
TCP and HTTP features \cite{Wang2016} & \textbf{0.997} & 0.973 & --- & 0.977 & --- & 0.973 \\
\midrule
\textit{Hybrid analysis models} & & & & \\
TCP features and permissions \cite{Kandukuru2017} & \textbf{0.956} & 0.943 & --- & 0.945 & \textbf{0.950} & 0.948 \\
Permissions and system calls \cite{Kapratwar2017} & --- & 0.968 & --- & 0.946 & --- & 0.969 \\
\midrule
\textit{Ensemble models} & & & & \\
\multicolumn{2}{l}{$E_1$ Dynamic system + static API calls (ours)}  & \underline{0.978} & & 0.976 & & \underline{0.979} \\
\multicolumn{2}{l}{$E_2$ Dynamic system + static API calls (ours)}  & 0.971 & & 0.971 & & 0.971 \\
\multicolumn{2}{l}{$E_4$ System/API calls + permissions + network (ours)}  & 0.969 & & 0.970 & & 0.966 \\ \bottomrule
\end{tabular}%
}
    \caption{\textcolor{black}{Summary of results, comparing originally published accuracies, F1 scores and TPRs (Original) against those of our reimplementations (Ours) and ensemble models. Where multiple models were evaluated in a study, only the best result is shown for each metric. Where a metric was not reported in the original study, we have indicated ---. Bold highlighting is used to show whether the original or reimplemented study produced a better result for each metric. The overall best values for each metric are underlined.}}
    \label{tab:comparison}
\end{table}

This study allowed us to take a broader perspective on the feature-based Android malware-detection literature and---arguably for the first time---see how previously published approaches compare and perform within a contemporary Android environment. Table \ref{tab:comparison} summarises the results. \textcolor{black}{Perhaps most importantly, these show that feature-based approaches can still perform very well within a contemporary context, with the best models achieving F1 scores of around 98\%. This exceeds many of the figures published in recent studies that use transformers and end-to-end learning, but without the added risks around overfitting, transparency, execution speed and computational costs that these approaches bring \cite{Lones2021}. Whilst this does not imply that more modern approaches are unable to achieve better performance than traditional machine learning, it does question the sometimes implicit assumption that more recent models are necessarily better. This suggests that the Android malware detection community should approach developments in AI in a more nuanced manner, building on the strengths of feature-based approaches.}

\textcolor{black}{Although the best feature-based malware detection models still achieve levels of performance inline with those originally published,} we also observed that a number of approaches performed significantly less well when reimplemented and evaluated on a large contemporary dataset. This is despite us performing hyperparameter optimisation and often broadening the scope of studies, both of which might be expected to produce better models. Some of this difference may be due to distribution shifts in malware and benign software since the original studies were published, plus the increasing complexity of the Android environment. However, poor experimental practice is also likely to be a contributing factor. Whilst the aim here is not to diminish the value of the  individual studies, we have observed many instances of common errors of practice \cite{Lones2021} which may have led to over-confident measures of performance. A typical example of this is performing feature selection prior to splitting, which potentially leads to information leakage from the test data. Within this context, results of several of the studies dropped from near-perfect classification scores to relatively low scores.

\textcolor{black}{The falls in accuracy are particularly pronounced. Table \ref{tab:comparison} shows that for those studies which did originally report accuracy figures, 77\% (i.e.~10 studies) of them performed less well in our reimplementation, with the mean dropping from 0.961 to 0.928. This is a considerable drop, and the fact that this particularly affects accuracy may suggest that the figures reported in the original studies were unreliable due to their use of imbalanced datasets. For F1 scores, although 78\% (7) performed less well after reimplementation, the difference in means is less pronounced, dropping from 0.947 to 0.934. However, fewer studies published F1 figures, so this distribution may be skewed towards those studies which followed better practice generally, including the reporting of a suitable selection of metrics.} Nevertheless, regardless of the underlying cause, the reduction in performance shows the importance of periodically reimplementing and reevaluating past studies.

Another important observation is the challenge caused by the growing complexity of Android. For instance, the number of API calls has grown by two orders of magnitude since early work on ML-based Android malware detection. This means the use of feature selection is now very important in order to reduce feature numbers down to manageable levels. Within this context, an important observation is that feature selection does not just reduce the computational resources needed for training; it also, in most cases, increases the accuracy of the resulting models. Most notably, we have shown that reducing the number of API calls by 95\% leads to higher levels of malware discrimination. In general, we found mutual information and chi-square to be the most effective feature selectors.
\textcolor{black}{
This is a useful observation, since these model-agnostic approaches are relatively cheap, which is important when working with large feature sets. Model-based selection, which attempts to fit features to a particular downstream model, is more expensive, but was not found to yield better results.
}

Some of the observations around feature utility are surprising. For instance, our results suggest that, \textcolor{black}{within the scope of our evaluation,} there is generally little advantage to using dynamic analysis techniques, and that models with similar accuracy can be built using relatively simple static features. The only scenario where dynamic analysis seemed to bring a marginal advantage was when we built models based around network activity. However, this is expensive to implement in practice, since collection of network activity requires dynamic analysis at both training and inference time, and relies on features which are both difficult to collect and potentially brittle.

This is not the only instance where we observed a trade-off between the difficulty of extracting features, and the accuracy of the resulting models. Another notable example is the use of API calls versus permissions. Generating features based upon API calls requires the use of a static analysis tool. By comparison, permissions can be read directly from an application's manifest file, a trivial operation. However, we found models trained on API call features to be significantly more accurate than models based upon permissions, showing that it is sometimes worth expending more effort on feature extraction.

We also found that hybrid models, built from a combination of static and dynamic features, were no better than those built from static or dynamic features alone. However, importantly, we showed that more accurate models can be constructed by ensembling the best static and dynamic models. Most significantly, we found that ensembling can be used to build models with accuracies comparable to the best network traffic models, but using less brittle features.

The Drebin feature set has historically played an important role in ML-based malware detection studies. Although this feature set still performs reasonably well, our results suggest that it is no longer the best feature set to use, and is difficult to deploy due to the large number of features it entails when used within a contemporary Android environment. Unlike other feature sets we studied, it does not benefit from feature selection.

For features such as API calls, where it is possible to measure frequency of use in addition to whether they are used, this extra information generally does not improve a model's ability to discriminate malware. This is a somewhat unexpected finding, but was observed consistently across a number of feature sets, both static and dynamic.

Beyond features, we also found that model choice is important, and random forests were surprisingly consistent in being the best performing models across the majority of experiments and feature sets. SVMs also performed well in general, but were more expensive to train than random forests. Naïve Bayes models generally performed poorly, suggesting that, for malware detection, interactions between features are important. A somewhat surprising observation, given their increasing prevalence within ML more generally, is that deep neural network models did not perform particularly well. However, this may be due to the use of pre-extracted features in the majority of malware detection studies.

\section{Limitations and Future Work} \label{sec:limitations}


Our aim in this study was to sample and reimplement representative feature-based ML studies from the literature. It is not feasible to reimplement all existing studies, and there is inevitably some sampling bias within our approach. In particular, we focus on studies which used similar underlying features, since this enables a more meaningful basis for comparison across a standardised data set. This approach was motivated by our earlier review~\cite{Muzaffar2022} of the most common ML-based modelling approaches in Android malware detection. However, it should be noted that there are other studies which have used broader information sources to construct features. For example, Hou et al.~\cite{Hou2017}, in their HinDroid framework, consider the relationships between applications and APIs. 
Similarly, Ye et al.~\cite{ye2019out} utilized features that incorporated various forms of metadata from apps and devices, including IMEI numbers. In future work, it would be interesting to implement and compare studies that incorporate these more heterogeneous features.

\textcolor{black}{Although beyond the scope of this study, there are other characteristics of malware which can be important in practice. This includes the use of code obfuscation and packing methods, which are both sometimes used to evade detection. By assessing the detection of malware in bulk, it is unclear from our results how well different modelling approaches deal with this, and it would also be an interesting topic for a future study. Another important dimension within antimalware research is how to deal with the changing nature of malware over time. Whilst our study gives some insight into the robustness of particular modelling approaches in this regard, this was not a focus of this work, and we point interested readers to our recent work on using active learning as a means for responding to changes in the distribution of Android malware over time \cite{actdroid}.}

\textcolor{black}{Finally, it is likely that recent end-to-end learning and transformer-based approaches might suffer from the same issues around ML practice observed in feature-based studies. For this reason, it would also be useful to reimplement some of these more recent studies within a level playing field. However, the reimplementation of these models is beyond the scope of the framework developed for this current study.}

\section{Conclusions} \label{sec:conclusions}


\textcolor{black}{This work revisited and reimplemented previous studies in feature-based ML for Android malware detection. The 18 foundational studies that we considered were originally published over a 10 year time period, from 2013--2023, during which both the Android operating system and the malware ecosystem evolved significantly. One of our aims aim was to determine how well these approaches work within a contemporary Android environment. Our results suggest they still hold up well, with the most effective models achieving accuracies and F1 scores of around 98\%. This is more or less in line with the published performance of more recent approaches based around transformers and end-to-end learning.}

However, overfitting is easily achieved  when good practice is not followed. In this respect, it is concerning that the performance of most of the anti-malware models following reimplementation was less---and sometimes considerably less---than the figures originally published. This is likely to be at least partially attributable to issues of ML practice. We noted poor reporting standards (including the inappropriate use of metrics and the lack of statistical rigour) as a particularly common problem, but there are many other aspects of practice that can lead to over-confident results. We believe there is a need for better awareness of good ML practice in order to improve the replicability and robustness of research in this area.


A key observation was that features extracted using static analysis of Android APKs were often more effective than those extracted using dynamic analysis of running applications. This is important to know, since static analysis is much cheaper than dynamic analysis. Within this context, we found that simple representations of API usage led to very effective models, with accuracies of up to 96.8\%. The exception to this is when features extracted from network traffic are used to build models. In our study, it is these models that provided the highest levels of malware discrimination, achieving accuracies of up to 97.7\%. However, they are expensive to train and use, and rely on features that are difficult to collect and potentially brittle. To address this, we also showed that ensemble models built from non-network features can achieve a similar level of discrimination (97.8\%), offering a more practical alternative for real world deployment.

Other notable findings include the importance of using feature selection algorithms when dealing with the complexity of modern Android systems, the relatively poor performance of the historically-important Drebin feature set, and the high accuracy of random forest models in comparison to other ML models, including various deep learning models. \textcolor{black}{This latter point is especially relevant in the face of current trends in anti-malware research, which are exploring increasingly complex ML models, including LLMs. Given that relatively simple models can reach very high accuracies, this questions the need to use slower more resource-intensive models that are in general more susceptible to overfitting and to adversarial attacks.}


\appendix
\section{DroidDissector feature extraction tool} \label{appendix:droiddissector}

This appendix provides full details of the automated analysis tool, DroidDissector~\cite{muzaffar2023droiddissector}, which we used to extract static and dynamic features from Android applications. It comprises two sub-systems: one for extracting static features, and another for extracting dynamic features.

\subsection{Static Analysis \textcolor{black}{Sub-System}} \label{sec:methodology:statictool}
 
\textcolor{black}{The static analysis sub-system} takes in a directory of APK (the application file format used by Android) files as its parameter and produces static analysis reports for each of these applications.
 
APK files are first reverse-engineered and decompressed using APKtool \cite{APKTool} to decode the manifest file and smali files from the APK package. Permissions used by the application, the application package name and the different application components \textcolor{black}{--- activities, services, broadcast receivers, and content providers ---} are stated in the Android manifest file. APKtool decodes the dex (binary code) files into readable Java-like smali files.  APKTool creates a directory for each APK file. The static analysis tool then analyzes the files produced by the APKTool to retrieve features. For each application, the features extracted by our static analysis tool include: 
 
\begin{description}
 
    \item[Hardware Components:] The hardware components that the application needs to use are defined in the manifest file of the application. 
 
    \item[Requested Permissions:] The permissions required by the application as defined in the manifest file. 
 
    \item[App Components:] The application components required by the application as defined in the manifest file. These are activities, services, broadcast receivers, and content providers.
 
    \item[Filtered Intents:] The filtered intents used by the application defined in the manifest file.
 
    \item[Used Permissions:] The permissions actually used by the application from manifest and smali files.
 
    \item[Network Addresses:] The network addresses present in the source code of the application from the smali files. 
    
    \item[API Calls v30:] The API calls in the latest SDK present in the source code of the application in the smali files. 
    
    \item[Restricted API Calls:] List of restricted API calls defined by Drebin \cite{Arp2014}. We built this feature set by analyzing the smali files extracted from the APK.
    
    \item[Suspicious API Calls:] List of suspicious API calls defined by Drebin \cite{Arp2014}. We built this feature set by analyzing the smali files extracted from the APK.
   
    \item[API Call Graphs:] FLOWDROID is used to extract call graphs produced by the application from smali files.
 
    \item[Opcodes:] The operation codes, or opcodes, present in the application. Opcodes are machine language instructions.
 
\end{description}   
 
We built the static analysis tool using Python 3.8 and Java. Except \textcolor{black}{for} API call graphs, we performed the static analysis of each APK file using Python. We integrated FLOWDROID, a Java library for taint analysis, into our static analysis tool to extract API call graphs. A static analysis of an APK file can take anywhere from five seconds to ten minutes depending on the size of the application. We saved a separate file containing all the features extracted from an application for every APK file. The resulting feature files follow a standard format, comprising the feature type followed by ``::" and then \textcolor{black}{the name of the feature extracted. For example}, ``RequiredPermission::android.permission.ACCESS\_NETWORK\_STATE". The tool stores API call graphs extracted using FLOWDROID in a separate JSON file. It also stores opcodes from an application's bytecode in a separate file.

\subsection{Dynamic Analysis \textcolor{black}{Sub-System}} \label{sec:methodology:dynamictool}

\textcolor{black}{The dynamic analysis sub-system runs on the readily-available} Android emulators provided by Android SDK \cite{android_sdk}. The virtual images in the emulators also allow root access, which is essential for dynamic analysis. Here we describe the environment required to run the tool and the features extracted by the tool. 
 
\subsection{Environment}
 
We developed and tested the tool on Ubuntu, CentOS and Windows 10. The tool requires the following to run: 
 
\begin{itemize}
 
    \item Python 3: All the dynamic modules require Python to run.
 
    \item Frida-Android: The tool requires Frida’s Python library to hook to API calls during execution of an application.
 
    \item Android Emulator: Our tool uses the emulator provided by Android SDK as the virtual environment to run the analysis. The emulator should be on the system path. We have tested the tool on Android versions up to and including the current release (version 15).
 
    \item The tool requires \textit{adb} from Android SDK and it should be on the system path. The dynamic analysis tool connects to the virtual device using adb and executes commands on the emulator using adb.
 
    \item Frida-server needs to be running on the virtual device to capture the hooked API calls while running the application.
 
    \item The tool requires a static analysis report \textcolor{black}{in order to analyse API calls}. If a static analysis report is not \textcolor{black}{provided, the dynamic analysis sub-system also} requires APKTool to perform static analysis on the application.
 
\end{itemize}
 
\subsection{Feature Extraction}
 
The dynamic analysis \textcolor{black}{sub-system} can extract the key dynamic features used in Android malware detection. These include system calls, API calls, logcat files and network traffic. The tool uses Android's \textit{monkey} tool to input virtual events that imitate an application's user input. It allows the user to set the number of virtual events and/or the maximum time the application should run for. \textcolor{black}{The application is run in the controlled environment of the Android emulator provided by the Android SDK. There is} an option to hide the emulator to save memory, but before running the analysis the Android emulator needs to be set up. 

 
\textcolor{black}{The dynamic analysis sub-system comprises the following feature extraction modules:}
 
\begin{description}
\item[System Calls:] The system calls module extracts and saves all the system calls made by the application during its \textcolor{black}{runtime} using \textit{strace}. \textcolor{black}{This} is a Linux utility used to monitor the interaction between the process (application) and the OS. While performing system calls analysis, it is essential to collect all the system calls that occur during \textcolor{black}{runtime}. From past works, we see researchers run the application and then retrieve the application process ID (pid) to monitor system calls associated with this. This approach might cause some system calls to be missed in the time between starting the application and running the strace tool. To mitigate against this, we run strace on the zygote process of the OS. The zygote process \textcolor{black}{is the first process to run on an Android system}. By monitoring the zygote process, we monitor all the processes in the OS. When the analysis ends, we filter out the system calls used by the application we are monitoring from the complete system call analysis extracted by monitoring the zygote process. The tool then saves the output in a separate file for every application.

\item[Network Traffic:] Our tool uses Android emulator’s \textit{tcpdump} to collect the complete network traffic of the application. It stores the resulting data in ``pcap" files, later used to analyze and extract features. \textit{Wireshark} \cite{Wireshark} can process this file to analyze the application's network traffic. It also saves a separate network traffic file for every application.
 
\item[Android Logcat:] The tool uses \textit{logcat}, a command line tool provided by Android SDK to dump all the messages and errors the device throws during the execution of the application. The \textit{adb} utility runs \textit{logcat} and saves the output in a file for further analysis.
 
\item[API Calls:] The tool requires hooking of API calls beforehand to monitor execution of API calls during the run-time of an application. However, it is not computationally feasible to monitor the 130,000+ API calls present in the Android SDK. Therefore, the tool carries out the following steps: before starting the dynamic analysis, it checks if a corresponding static analysis report is present for the application. In addition to hooking the classic calls used in the literature of dynamic analysis tools \cite{mobsf}, we also hooked application-specific API calls. If present, it hooks the API calls extracted in the static analysis report, or else, it performs a static analysis on the application using the static analysis tool. The dynamic analysis tool then hooks the API calls to monitor during the execution of the application using Frida. The tool saves a separate file with all the APIs called during execution for further analysis. As the process may involve static analysis, the API call module is optional in the dynamic analysis tool. By default, the tool monitors a pre-defined list of API calls; however to monitor the API calls mined from an application's static analysis, the static analysis module is required. 
 
\end{description}

\bibliography{library.bib}

\end{document}